\documentclass[Afour,sageh,times]{sagej}

\usepackage{moreverb,url}
\usepackage{xcolor}
\usepackage{color, soul}
\usepackage{caption}
\usepackage{subcaption}
\usepackage{amsmath}
\usepackage{booktabs}
\usepackage{xstring}
\usepackage{hhline}
\usepackage{siunitx}
\usepackage{physics}
\usepackage{tikz}
\usepackage{amssymb,amsfonts}
\usepackage{algorithmic}
\usepackage{tcolorbox}
\usepackage{textcomp}
\usepackage{array,multirow,graphicx}

\definecolor{lightgray}{RGB}{200, 200, 200}
\sethlcolor{lightgray}

\usepackage[colorlinks,bookmarksopen,bookmarksnumbered,citecolor=black,urlcolor=red]{hyperref}

\newcommand\BibTeX{{\rmfamily B\kern-.05em \textsc{i\kern-.025em b}\kern-.08em
T\kern-.1667em\lower.7ex\hbox{E}\kern-.125emX}}

\definecolor{rvc}{RGB}{0, 0, 0}
\definecolor{suyong}{RGB}{0, 255, 0}
\definecolor{cv2}{RGB}{0, 0, 0}
\definecolor{qw}{RGB}{0, 0, 0}
\definecolor{qwr}{RGB}{0, 0, 0}
\definecolor{qwe}{RGB}{0, 0, 0}
\newcommand{\argmin}{\mathop{\mathrm{argmin}}}  
 
\newcommand{\rom}[1]{\uppercase\expandafter{\romannumeral #1\relax}}

\newcommand{\numconvention}{5}
\newcommand{\numdeep}{3}

\begin{document}

\runninghead{Lim et al.}

\title{Quatro++: Robust Global Registration Exploiting Ground Segmentation \\for Loop Closing in LiDAR SLAM}

\author{Hyungtae Lim\affilnum{1}, Beomsoo Kim\affilnum{2}, Daebeom Kim\affilnum{3}, Eungchang Mason Lee\affilnum{1}, and Hyun Myung\affilnum{1*}}

\affiliation{\noindent \affilnum{1}School of Electrical Engineering at KAIST (Korea Advanced Institute of Science and Technology), Daejeon, 34141, Republic of Korea \\
\affilnum{2}Department of Artificial Intelligence, Hanyang University, Seoul, 04763, Republic of Korea \\
\affilnum{3}Robotics Program at KAIST, Daejeon, 34141, Republic of Korea}
\corrauth{$^{*}$Prof. Hyun Myung is with School of Electrical Engineering and KI-R at KAIST, Daejeon, 34141, Republic of Korea.}
\email{hmyung@kaist.ac.kr}

\begin{abstract}

Global registration is a fundamental task that estimates the relative pose between two viewpoints of 3D point clouds. 
However, there are two issues that degrade the performance of global registration in LiDAR SLAM: 
one is the sparsity issue and the other is degeneracy.
The sparsity issue is caused by the sparse characteristics of the 3D point cloud measurements in a mechanically spinning LiDAR sensor.
The degeneracy issue sometimes occurs because the outlier-rejection methods reject too many correspondences, leaving less than three inliers.
These two issues have become more severe as the pose discrepancy between the two viewpoints of 3D point clouds becomes greater.
To tackle these problems, we propose a robust global registration framework, called \textit{Quatro++}. 
Extending our previous work that solely focused on the global registration itself, we address the robust global registration in terms of the loop closing in LiDAR SLAM. 
To this end, ground segmentation is exploited to achieve robust global registration.
Through the experiments, we demonstrate that our proposed method shows a higher success rate than the state-of-the-art global registration methods, overcoming the sparsity and degeneracy issues.
In addition,  we show that ground segmentation significantly helps to increase the success rate for the ground vehicles.
Finally, we apply our proposed method to the loop closing module in LiDAR SLAM and confirm that the quality of the loop constraints is improved, showing more precise mapping results.
Therefore, the experimental evidence corroborated the suitability of our method as an initial alignment in the loop closing.
Our code is available at \href{https://quatro-plusplus.github.io}{\texttt{https://quatro-plusplus.github.io}}.

\end{abstract}


\maketitle

\section{Introduction} \label{sec:intro}

3D point cloud registration is a fundamental task that estimates relative pose between two viewpoints of 3D point clouds, namely source and target, in robotics and computer vision fields. 
Using the characteristic of being able to estimate the relative pose between two point clouds, 3D point cloud registration is widely exploited in various applications such as relocalization~\citep{du2020dh3d}, ego-motion estimation~\citep{koide2020vgicp}, object recognition~\citep{chua19963d, ashbrook1998finding,belongie2002shape,cho2014projection}.

Furthermore, these point cloud registration methods are key components in light detection and ranging~(LiDAR) sensor-based simultaneous localization and mapping~(SLAM), which is the integral part of building a map while localizing a robot or an autonomous vehicle itself.
In particular, point cloud registration is not only used in odometry, but also exploited in the loop closing procedure to obtain constraints for pose graph optimization~(PGO) in graph-based SLAM.

In general, the graph-based SLAM mainly consists of three parts~\citep{thrun2006graph, grisetti2010tutorial, qin2018vins, shan2020lio}: a)~odometry, which estimates relative pose between the consecutive frames, b)~loop detection, which achieves data association between two non-consecutive frames, and c)~loop closing, which estimates relative pose between the non-consecutive frames searched by the loop detection module.
The results through these three parts are represented in a graph structure, i.e. nodes and vertices. Then, the results are taken as inputs of PGO.
The graph-based SLAM reduces global trajectory errors through this procedure, building a precise and globally consistent map.

In recent years, as the demand for LiDAR-based SLAM has increased, novel and accurate LiDAR-based SLAM frameworks have been proposed. 
However, these novel methods usually focus on improving the odometry or loop detection modules. 
Once the performance of loop detection is improved, the quality of constraints from the loop closing is also increased because the relation between loop detection and loop closing is not completely independent~\citep{chen2022overlapnet}. 
However, the methods for increasing the quality of loop constraints in the loop closing step is still less actively studied. 

Furthermore, many LiDAR-based SLAM methods only utilize iterative closest point~(ICP) or its variants, which are referred to as local registration, on the loop closing step. 
However, the local registration searches the point pairs by using nearest neighbor search, so the local registration-based pose estimation is only valid if two point clouds are mostly overlapped and the pose discrepancy between the two point clouds is small (empirically, within 2 m and 10$^\circ$~\citep{kim2019gp,lim2020normal}). 
Otherwise, the estimates of the local registration are highly likely get stuck in the local minima.
For this reason, even though loop detection finds the correct loop candidate, the loop candidate with a large pose discrepancy can be rejected owing to the narrow convergence region.
Consequently, these false negative loops potentially hinder more accurate PGO results.

To tackle this problem, global registration methods, whose performance is less affected by the initial pose difference between the two point clouds, can be utilized as a coarse alignment.
That is, the initial guess using the global registration reduces the pose difference in a coarse manner. 
This reduction of the pose discrepancy helps local registration methods converge into a global optimum.
As a result, the loop candidates with large pose discrepancies can also be exploited as precise loop constraints.

However, there are still two issues that degrade the performance of the global registration in 3D LiDAR SLAM: one is the sparsity issue~\citep{lim2021patchwork} and the other is degeneracy~\citep{lim2022single}.
First, the sparsity issue occurs in 3D point cloud measurements when using a mechanically spinning LiDAR sensor: the farther the range from the origin goes, the dramatically more sparse the density of the 3D point cloud becomes.
This decreasing density degrades the expressibility of the feature descriptors.
That is, even if the same place is observed, the descriptor has a different value due to the density difference between the two point clouds.
Consequently, the sparsity issue induces lots of false matching, increasing the ratio of the outliers within the estimated correspondences.

Second, degeneracy sometimes occurs owing to the effect of outlier rejection algorithms. Here, degeneracy refers to the case when fewer inliers remain than the degree of freedom~(DoF). 
The degeneracy is occasionally observed when the well-known outlier rejection methods, such as maximum clique inlier selection~(MCIS)~\citep{shi2020robin,sun2021ransic} or weight update in graduated non-convexity~(GNC)~\citep{zhou2016fast,yang2020teaser}, are employed.
These methods unintentionally prune too many estimated correspondences because these methods usually aim to achieve high recall of outlier rejection; thus, sometimes less than three inliers are left after the pruning step.
As a result, the degeneracy triggers tilted or flipped results~(see Section~\ref{sec:exp_sota}).
These two issues become more severe as the pose difference between the two viewpoints of 3D point clouds becomes greater. 

\begin{figure*}[t!]
\centering
\captionsetup{font=footnotesize}
\begin{subfigure}[b]{0.9\textwidth}
	\centering
	\includegraphics[width=1\textwidth]{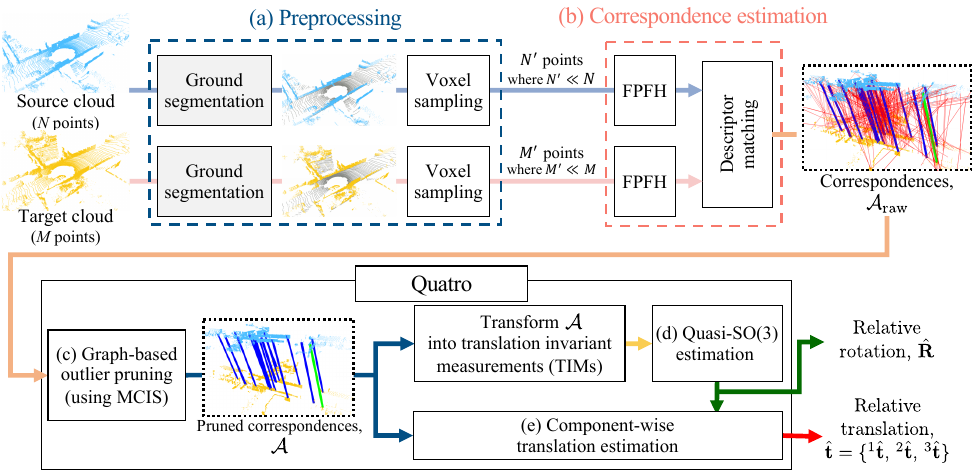}
\end{subfigure}
\caption{Overview of our global registration method, called \textit{Quatro++}. As an extension of our previous work~\citep{lim2022single}, Quatro++ consists of five parts. The red, green, and blue lines of the raw and pruned correspondences~(shown inside dotted boxes), i.e. $\mathcal{A}_\text{raw}$ and $\mathcal{A}$, respectively, denote the outliers, true inliers, and quasi-inliers~(see Section~\ref{sec:rot_estimation}), respectively. (a) The preprocessing step using ground segmentation (gray points denote the estimated ground points). (b) Correspondence estimation by using descriptor extraction and matching, which can be interchangeable with other methods. (c) Graph-based outlier pruning to initially reject outlier correspondences. (d)~Quasi-SO(3) estimation based on graduated non-convexity~(GNC). (e) Component-wise translation estimation.~(best viewed in color).}
\label{fig:overview_quatropp}
\end{figure*}

In our earlier work, i.e. Quatro~\citep{lim2022single}, we demonstrated that Quatro overcomes the aforementioned problems by leveraging the decoupling-based method that reduces the minimum number of required correspondences for the estimation from three to one.
However, Quatro was still not exploited as a loop closing module in SLAM;
thus, the influence of robust global registration on the performance of graph-based SLAM was still not closely examined.

Therefore, we propose a robust global registration framework, called \textit{Quatro++}, and analyze the effect of the global registration from the loop closing-centric perspective. 
As shown in Fig.~\ref{fig:overview_quatropp}, Quatro++ is a superset of Quatro that exploits ground segmentation to be more robust against outlier pairs and degeneracy issues.
In Quatro++, we introduce ground segmentation to allow global registration to be more robust against distant cases by filtering ground points, which are most likely to be featureless, before the pose estimation.
Even though many researchers empirically use ground segmentation as a preprocessing step~\citep{kim2018scancontext, yue2019data, komorowski2021minkloc3d}, the effect of ground segmentation on the performance of global registration has seldom been quantitatively analyzed.
Thus, we validate that ground segmentation substantially increases the success rate of global registration even in degenerate environments, such as corridor-like scenes, or distant cases.
%To the best of our knowledge, it is the first attempt to analyze the effect of global registration on the loop closing module of LiDAR SLAM.
Finally, as an extension of our previous work~\citep{lim2022single}, we additionally compare our approach with state-of-the-art learning-based approaches~\citep{shi2021keypoint,cattaneo2022lcdnet} and show that our approach achieves substantially higher performance than other approaches in loop closing level.

In summary, our ultimate goal is to achieve robust global registration when spurious correspondences caused by the degradation of feature extraction and matching are given.
Unlike other studies that focus more on improving the expressibility of feature descriptors to increase the quality of correspondences~\citep{ao2021spinnet,shi2021keypoint,chen2022overlapnet,cattaneo2022lcdnet,yin2023segregator}, what we aim for is robustly estimating the relative pose even though undesirable imprecise correspondences are given.
The contribution of this paper is fourfold:

\begin{itemize}
\item We propose a novel global registration framework, Quatro++, that robustly achieves an initial alignment, overcoming the effect of the gross outliers and degeneracy issue as well.
\item We demonstrate that employing ground segmentation makes global registration more robust and reduces the computational cost, allowing more accurate and fast loop closing for LiDAR SLAM.
\item Our proposed method was analyzed from various perspectives, showing a superior performance compared with state-of-the-art methods, including deep learning-based approaches. By doing so, we demonstrate the ease of exploiting our Quatro++ because our method is a learning-free approach.
\item By integrating our approach with various LiDAR SLAM frameworks, we showed that our Quatro++ successfully improves the performance of SLAM by providing more precise loop constraints as an initial alignment.
\end{itemize}

%

%%%%%%%%%%%%%%%%%%%%%%%%%%%%%%%%%%%%%%%%%%%%%%%%%%%%%%%%%%%%%%%%%%%%%%%%%%%%%%%%
% Related works
%
\section{Related Works}
\subsection{Local Registration}

As mentioned earlier, point cloud registration is mainly classified into two categories depending on how to find correspondences between the two point clouds: one is the local registration \citep{besl1992method,rusinkiewicz2001efficienticp,chetverikov2002trimmed_icp, segal2009gicp,pomerleau2013comparing,koide2020vgicp,dellenbach2022cticp} and the other is the global registration~\citep{fischler1981ransac,dong2017gh-icp,yang2015goicp,zhou2016fast,yang2020teaser,bernreiter2021phaser}. 

First, the local registration methods heavily rely on the nearest neighbor search~\citep{greenspan2001nearestneighbor}, which assumes that a target point correlates with the source point closest to the target point.
ICP~\citep{besl1992method} and its subsequent studies are renowned local registration methods. 
Unfortunately, the assumption above is invalid once two point clouds are far apart and slightly overlapped. 
For this reason, ICP variants in loop closing situations often fail to estimate relative pose when the pose discrepancy between the two point clouds is large.  

\subsection{Global Registration}

Unlike these local registration methods, global registration is more likely to be appropriate to estimate the relative pose in the case where the pose discrepancy is large. 
Accordingly, the output of the global registration can be used as an initial guess, allowing the estimate of the local registration to successfully converge into the global minimum.
Note that the two types of global registration methods exist: a)~correspondence-based ~\citep{fischler1981ransac,yang2015goicp,zhou2016fast,dong2017gh-icp,tzoumas2019outlier,yang2020teaser,lin2022kcp} and b)~correspondence-free methods~\citep{rouhani2011correspondence,brown2019family,bernreiter2021phaser}. 
In this study, we place more emphasis on the correspondence-based methods.

\subsection{Correspondence-Based Global Registration}

The correspondence-based methods utilize feature extraction and matching to obtain correspondences between the two point clouds. 
However, the outliers inevitably occur within the putative correspondences. To tolerate the effect of outliers on pose estimation, numerous researchers have studied outlier-robust global registration methods. 
Typical examples are random sample consensus~(RANSAC) proposed by~\cite{fischler1981ransac}, and its variants \citep{papazov2012rigid,chum2003locally,choi1997performance,schnabel2007efficient}. However, the RANSAC-based methods are likely to be brittle with high outlier rates.
Empirically, these methods often fail if the ratio of outliers is over 50$~\%$~\citep{tzoumas2019outlier}.

The other examples are branch-and-bound (BnB)-based methods~\citep{lawler1966branch,olsson2008branch,hartley2009global,pan2019multibnb}. 
These methods have an advantage in that they guarantee theoretical optimality. 
Unfortunately, these BnB-based methods are too slow to be exploited in real-world applications, e.g. it takes over 50$~\text{sec}$ for a single registration~\citep{lei2017fast}.

\subsection{Fast and Outlier-Robust Global Registration}

Unlike the RANSAC-variants or BnB-based methods, GNC has been introduced to achieve outlier-robust and fast registrations. 
The GNC-based methods estimate pose while rejecting the outlier measurements simultaneously~\citep{black1996unification,zhou2016fast}, overcoming up to 70-80\% of outliers and providing faster speed compared with the previous works. 
Consequently, GNC has been widely used to robustly estimate the relative pose against the gross outliers~\citep{tzoumas2019outlier,yang2020gnc,yang2020teaser,sun2021iron,song2022dynavins}.

In addition, the certifiable methods that guarantee their optimality in polynomial-time have been proposed, such as semidefinite programming and sums of squares relaxation-based methods ~\citep{vandenberghe1996semidefinite,briales2017convex,maron2016point,carlone2016planar}. 

\subsection{Geometry-Aware Traditional Outlier Rejection Modules}

Further, some researchers have studied novel traditional outlier rejection modules. 
To check the geometric consistency and thus prune spurious correspondences effectively, these outlier rejection methods leverage spectral techniques~\citep{leordeanu2005spectral}, maximum clique~\citep{shi2020robin,sun2021ransic}, voting schemes~\citep{glent2014search,yang2019ranking}, or game theory~\citep{rodola2013scale}.

\subsection{Deep Learning-Based Registration}

As deep learning era has come, deep learning-based methods have shown remarkable achievements. 
The research directions can be categorized into three groups: a) improving the quality of feature extraction and matching~\citep{zeng20173dmatch,yew20183dfeat,deng2018ppf,gojcic2019perfect}, b) end-to-end registration~\citep{wang2019deep,gojcic2020learning}, c) learning-based outlier rejection techniques within the deep learning architecture~\citep{Choy_2020_CVPR,Pais_2020_CVPR, bai2021pointdsc}. 
Among them, the last group shares our philosophy that outliers are always inevitable. 

However, these data-driven methods are too fitted into the situations included in the training dataset; 
thus, the learning-based methods can result in catastrophic failure if they encounter untrained situations or the data captured by other sensor configuration are taken as inputs.
As we pursue a versatile method that is applicable in any environment, the learning-based methods are thus beyond our scope.

\section{Preliminaries}

In this section, we present the notation of variables and the problem definition of global registration. 
In addition, the challenges of global registration using sparse point clouds captured by a 3D LiDAR sensor are explained. In particular, we focus more on the mechanically spinning 3D LiDAR sensor because its uniform beam forming triggers a severe sparsity issue~\citep{lim2021patchwork}, potentially causing more outlier correspondences.

\subsection{Notation}

First of all, source and target clouds are denoted as $\mathbf{P}$ and $\mathbf{Q}$, respectively, and let us define their coordinate frames as $\mathcal{P}$ and $\mathcal{Q}$, respectively.
These point clouds consist of a number of cloud points, which can be expressed as $\mathbf{P}=\{\mathbf{p}_{1}, \mathbf{p}_{2}, \dots, \mathbf{p}_{N}\}$ and $\mathbf{Q}=\{\mathbf{q}_{1}, \mathbf{q}_{2}, \dots, \mathbf{q}_{M}\}$. 
Each point is expressed in the Cartesian coordinates, so the elements of $\mathbf{p}_{i}~(1 \leq i \leq N)$ and $\mathbf{q}_j~{(1 \leq j \leq M)}$ are comprised of $x, y,$ and $ z$ values, respectively. 

Next, we aim for correspondence-based global registration, so the correspondences set, $\mathcal{A}$, is given by feature extraction and matching; $\mathcal{A}$ consists of the indices pairs,~$(i, j)$, which indicate that the $i$-th point in $\mathbf{P}$ and the $j$-th point in $\mathbf{Q}$ are matched to each other. 
In general, it is natural that $\mathcal{A}$ inevitably includes inherent outlier pairs due to the false matching~\citep{rusu2009fpfh, yew20183dfeat, tzoumas2019outlier}.
Accordingly, $\mathcal{A}$ can be classified into the outlier set, $\mathcal{O}$, and inlier set,~$\mathcal{I}$, such that $\mathcal{O} \cup \mathcal{I}=\mathcal{A}$. Note that these two sets are disjoint, i.e. $\mathcal{O} \cap \mathcal{I} = \varnothing$.

Then, the relative rotation matrix and translation vector between $\mathcal{P}$ and $\mathcal{Q}$ are denoted as $\mathbf{R} \in \mathrm{SO}(3)$ and $\mathbf{t}\in\mathbb{R}^{3}$, respectively. 
Finally, the relationship between the $i$-th point in $\mathbf{P}$ and the $j$-th point in $\mathbf{Q}$ is defined as follows: 

\begin{equation}
	\mathbf{q}_{j}=\mathbf{R}\mathbf{p}_{i}+\mathbf{t}+\boldsymbol{\epsilon}_{ij}
	\label{eq:pt_relation}
\end{equation}

\noindent where $\boldsymbol{\epsilon}_{ij}\in\mathbb{R}^{3}$ denotes the unknown measurement noise.
That is, $\boldsymbol{\epsilon}_{ij}$ depends on whether the pair $(i, j)$ is a true inlier or not; 
$\boldsymbol{\epsilon}_{ij}$ is Gaussian noise if $(i,j) \in \mathcal{I}$ and is irregular error, otherwise. 

\subsection{Problem Definition}

Based on the notations above, the objective function of correspondence-based global registration \color{black} can \color{black} be defined as follows:

\begin{equation}
	\hat{\mathbf{R}}, \hat{\mathbf{t}} =\argmin_{\mathbf{R} \in \mathrm{SO}(3), \mathbf{t} \in \mathbb{R}^{3}}  \sum_{(i,j) \in \mathcal{A} }\rho\Big( r(\mathbf{q}_{j}-\mathbf{R} \mathbf{p}_{i}-\mathbf{t})\Big) 
	\label{eqn:goal}
\end{equation}

\noindent where $\rho(\cdot)$ denotes a robust loss function, such as Cauchy or Huber norm \citep{tzoumas2019outlier}, intended to suppress undesirable large errors caused by false correspondences; 
$r(\cdot)$ denotes the squared residual function, i.e. $\abs{\cdot}^2$ for the scalar and $\left\|\cdot\right\|^2$ for the vector. 

\begin{figure*}[t!]
	\captionsetup{font=footnotesize}
	\begin{subfigure}[b]{0.96\textwidth}
		\centering
		\includegraphics[width=1\textwidth]{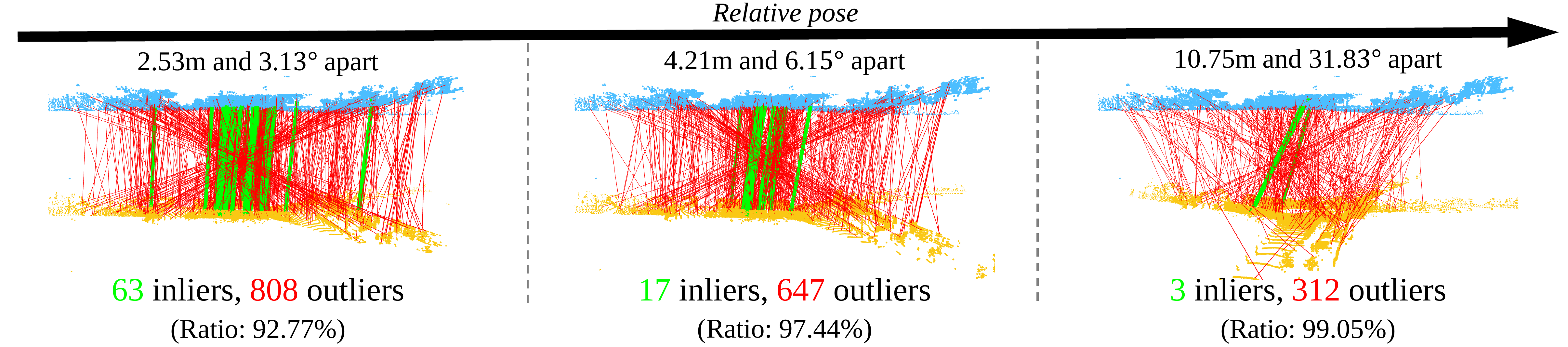}
	\end{subfigure}
	\caption{(L-R): Feature matching results between frames 939 as a source cloud~(cyan), and 4,206, 4,204 and 4,195 as target clouds~(yellow), respectively, for Seq.~\texttt{02} in the KITTI dataset. The farther the pose discrepancy is, the worse quality of the estimated correspondences is, showing an increase in the ratio of outliers (red lines) and a decrease in the number of actual inliers (green lines).}
% 	This degradation occurs owing to two issues. First, the density of a 3D point cloud captured by a mechanically spinning LiDAR sensor dramatically becomes too sparse as the range from the origin goes farther. Accordingly, for each point in the point cloud, the neighboring points are insufficient to precisely describe the actual geometrical inter-relations; thus, the expressibility of the descriptor inevitably decreases. Second, as the pose discrepancy between the two point clouds becomes larger, the actual overlapped area between two point clouds is reduced~(best viewed in color).
	\label{fig:challenge_of_feature_matching}
\end{figure*}

\subsection{Challenge of 3D Point Cloud Registration in Distant and Partially Overlapped Cases}\label{sec:challenge_of_global_registration}

Unlike object-level point cloud data whose points are evenly distributed, such as bunny dataset~\citep{curless1996volumetric}, a 3D point cloud captured by a spinning LiDAR sensor has somewhat different characteristics. 
That is, as the distance from the origin of a point cloud becomes \color{black} longer, the density dramatically decreases, \color{black} which is referred to as sparsity issue~\citep{lim2021patchwork}.
The sparsity issue directly has a negative impact on the quality of feature descriptors and thus induces false matching results.

To be more concrete, most feature descriptor methods, including deep learning-based methods, utilize the geometrical inter-relations between each point and its neighboring points to generate descriptors. 
However, as the density becomes sparse, these neighboring points are insufficient to describe the actual geometrical inter-relations because the distance between the two points becomes large. 
As a result, the expressibility of descriptors becomes worse. Finally, if these degraded descriptors are taken as the inputs of descriptor matching, it will deteriorate the matching performance, reducing the number of actual inliers while increasing the ratio of outliers, as shown in Fig.~\ref{fig:challenge_of_feature_matching}.

Based on these observations, (\ref{eqn:goal}) is redefined to achieve robust global registration as follows: 

\begin{equation}
	\hat{\mathbf{R}}, \hat{\mathbf{t}} =\argmin_{\mathbf{R} \in \mathrm{SO}(3), \mathbf{t} \in \mathbb{R}^{3}}  \sum_{(i,j) \in \mathcal{A} \setminus \hat{\mathcal{O}}}\rho\Big( r(\mathbf{q}_{j}-\mathbf{R} \mathbf{p}_{i}-\mathbf{t})\Big) 
	\label{eqn:final_goal}
\end{equation}

\noindent where $\hat{\mathcal{O}}$ denotes the estimated outlier pairs. 
In short, our ultimate goal is to robustly estimate relative pose although imprecise correspondences that contain the outliers are given.

\section{Quatro++: Quasi-SE(3) Estimation Leveraging Ground Segmentation}

%%%%%%%%%%%%%%%%%%%%%%%%%%%%%%
\subsection{System Overview}\label{sec:overview}

Our proposed method is based on two premises: 
a)~most terrestrial mobile robots and autonomous driving vehicles inevitably come into contact with the ground and thus 
b)~the yaw rotation is dominant in relative rotation motion, which are also presumed in \cite{scaramuzza20111dransac} and \cite{kim2019gp}. 

Based on these assumptions, our Quatro++ mainly consists of five parts, as shown in Fig.~\ref{fig:overview_quatropp}.
First, our proposed method begins with ground segmentation~\citep{lim2021patchwork} to filter ground points within the source and target clouds, respectively, because these points are usually geometrically featureless compared with the non-ground points~\citep{shan2018lego}.
Empirically, it was found that ground removal improves the quality of feature matching and thus dramatically increases the success rate of global registration~(see Sections~\ref{exp:quat_and_quatpp} and~\ref{exp:effect_of_ground_segmentation}). 
Second, feature extraction and matching are performed to estimate correspondences.

However, as mentioned in Section~\ref{sec:challenge_of_global_registration}, the potential outliers are unavoidable; thus, the outliers are initially filtered out by using MCIS as the third step. 
Next, the relative rotation and translation are estimated by the decoupling estimation method. 
Accordingly, in the fourth step, quasi-SO(3) estimation is performed by utilizing the concept of GNC to estimate the relative rotation in an alternating optimization manner, while rejecting the potential outliers. 
Fifth, component-wise translation estimation~(COTE) is exploited to estimate the relative $x$, $y$, and $z$, respectively, based on the concept of consensus set~\citep{li2009consensus}.
 
The methodologies and rationale for each module in Quatro++ are explained in the following subsections.

%%%%%%%%%%%%%%%%%%%%%%%%%%%%%%
\subsection{Fast and Robust Ground Segmentation as a Preprocessing Step}

\begin{figure}[t!]
	\centering
	\begin{subfigure}[b]{0.46\textwidth}
		\includegraphics[width=1.0\textwidth]{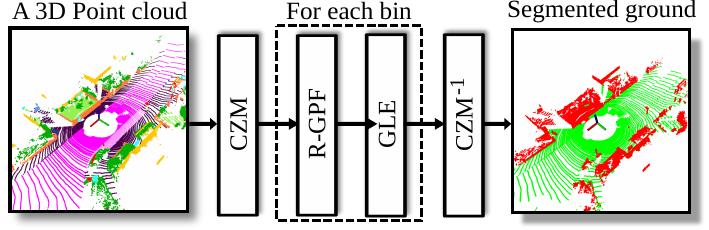}
		\caption{\centering}
	\end{subfigure}
	\captionsetup[subfigure]{justification=centering}
	\subfloat[\centering Ground segmentation module \\ in LeGO-LOAM]{
		\includegraphics[width=0.23\textwidth]{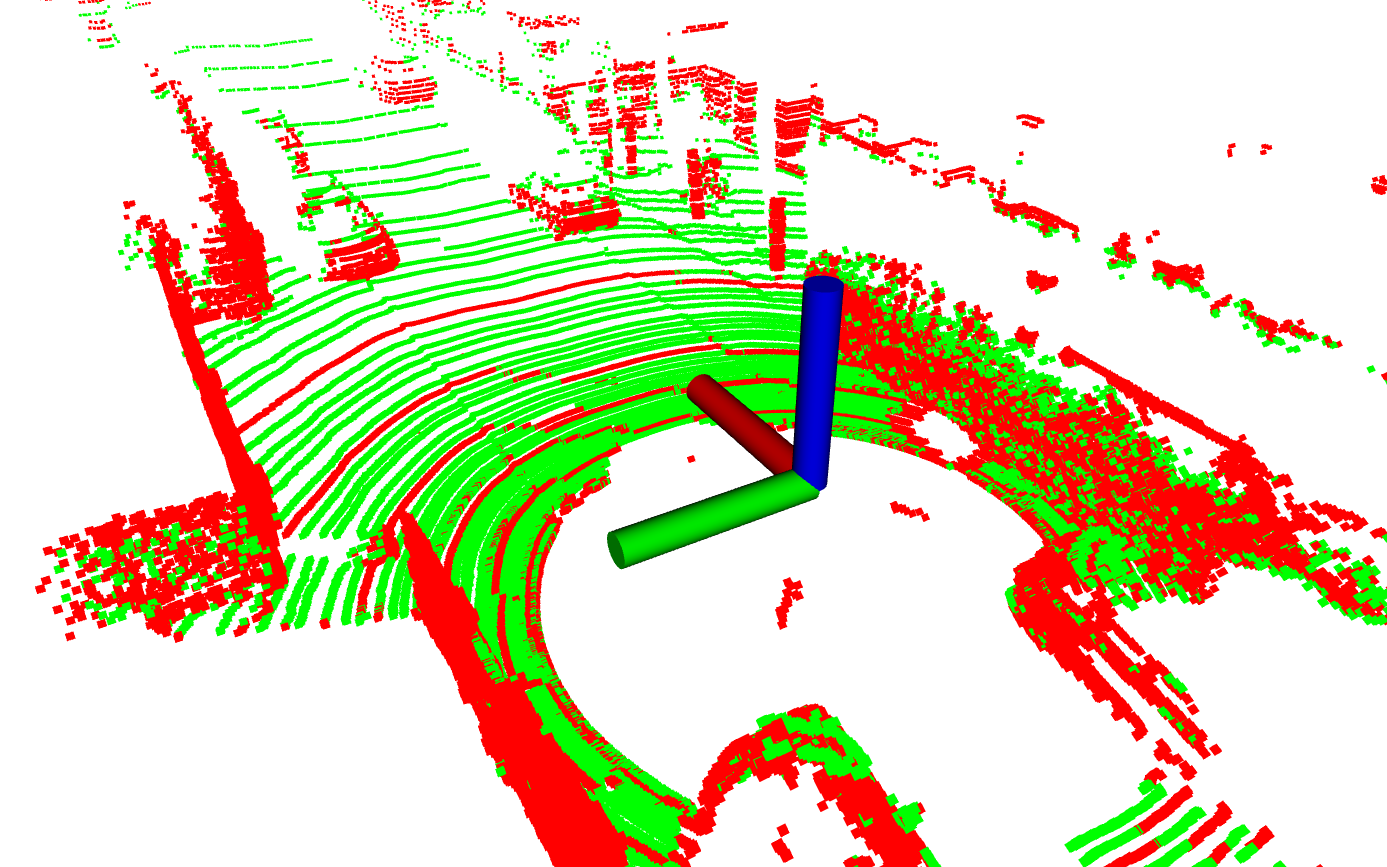}%
	}
	\captionsetup[subfigure]{justification=centering}
	\subfloat[Patchwork]{
		\includegraphics[width=0.23\textwidth]{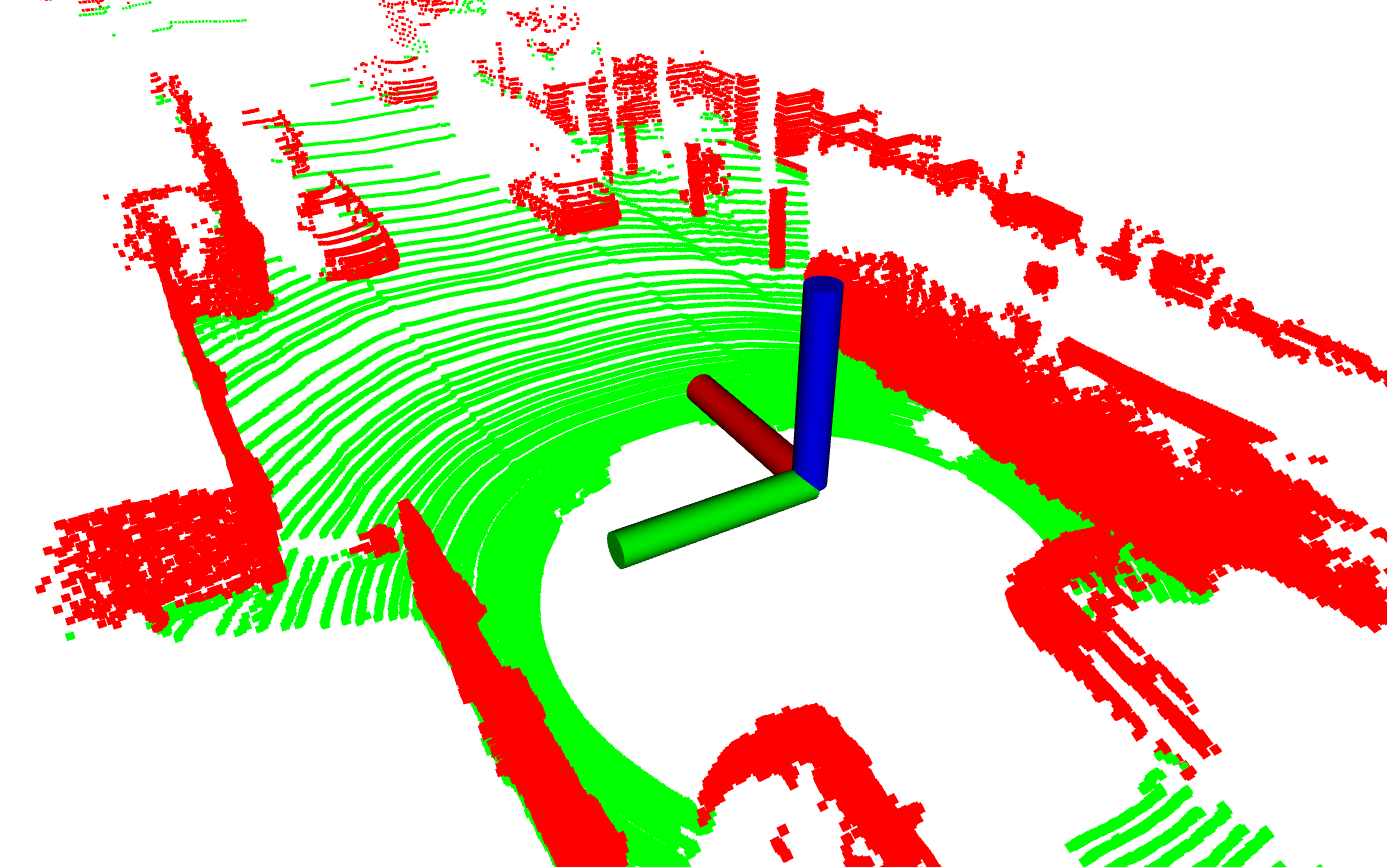}%
	}
	\captionsetup{font=footnotesize}
	\caption{(a)~Illustration of Patchwork~\citep{lim2021patchwork}. CZM, R-GPF, and GLE are abbreviations of concentric zone model, region-wise ground plane fitting, and ground likelihood estimation, respectively. (b)-(c)~Visualization of the ground segmentation output in Seq.~\texttt{00} in the KITTI dataset for comparison. The green and red colors denote the estimated ground and non-ground points, respectively (best viewed in color). (b)~Estimated ground points by using ground segmentation in LeGO-LOAM~\citep{shan2018lego} and (c)~by using Patchwork~\citep{lim2021patchwork}. Empirically, it was shown that Patchwork shows more robust segmentation with fewer false positives and false negatives (best viewed in color).}
	\label{fig:comparison_ground_seg}
\end{figure}

First, ground segmentation is performed by using Patchwork~\citep{lim2021patchwork} to extract non-ground points from the source and target cloud, respectively, as illustrated in Fig.~\ref{fig:comparison_ground_seg}(a).
Patchwork is a region-wise ground segmentation method that firstly divides all the cloud points into multiple subsets based on the concentric zone model~(CZM). 
% By doing so, even though the overall ground is non-flat, the ground in a local region can be approximated as a single plane.
Next, region-wise ground plane fitting~(R-GPF) is used to output the central point, normal vector, and flatness for each subregion. 
The flatness indicates the variance of points in the subset along the normal vector. 
Finally, ground likelihood estimation~(GLE) is used to check whether the estimated ground plane is likely to be actual ground or not. 
Note that the Patchwork is utilized because it shows consistent and precise ground segmentation performance compared with other well-known ground segmentation methods~\citep{lim2022pago}, as presented in Figs.~\ref{fig:comparison_ground_seg}(b) and~\ref{fig:comparison_ground_seg}(c).

In summary, ground segmentation has two major advantages. 
First, it increases the matching accuracy. 
In fact, the ground points usually produce lots of false matching because ground points have more ambiguous geometrical characteristics than non-ground points. 
Therefore, by rejecting these geometrically indistinguishable points beforehand, it is shown that ground segmentation significantly increases the success rate of the global registration methods in the distant cases~(see Sections~\ref{exp:quat_and_quatpp} and~\ref{exp:effect_of_ground_segmentation}).

%%%%%%%%%%%%%%%%%%%%%%%%%%%%%%
\begin{figure*}[t!]
	\captionsetup{font=footnotesize}
	\centering
	\begin{subfigure}[b]{1\textwidth}
		\includegraphics[width=1.0\textwidth]{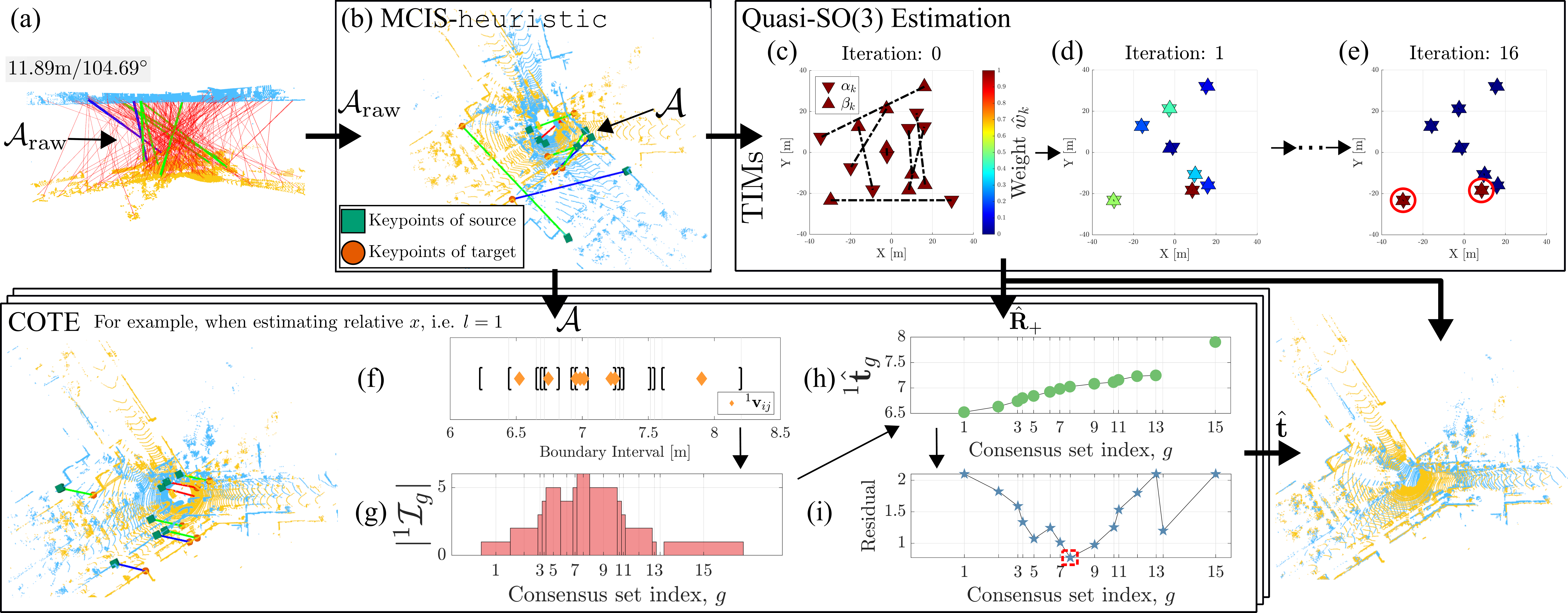}
	\end{subfigure}
	\caption{Visual description of the procedure of \textit{Quatro} when the frames 2,328 and 3,268 for Seq.~\texttt{00} in the KITTI dataset are given as source (cyan) and target (yellow), respectively, which is a distant and partially overlapped case. (a)~Estimated correspondences, $\mathcal{A}_\text{raw}$, which include potential outliers owing to false matching. The left-top text represents the pose discrepancy. (b)~Filtered correspondences by using MCIS-\texttt{heuristic}, which is a graph-based outlier pruning method. (c)-(e)~Procedure of Quasi-SO(3) estimation based on graduated non-convexity~(GNC). (c)~On the initialization step, $\mathcal{A}$ is transformed into translation invariant measurements~(TIMs) all of whose weights are set to one~(brown color). (d)~Next, GNC-based optimization is performed to estimate relative rotation while rejecting outliers by reweighting the weights. (e)~The problem is that less than three pairs are occasionally left, which is highlighted by red circles, by assigning near-zero values to the weights of correspondences. Despite this degeneracy, our quasi-SO(3) estimation robustly outputs the relative rotation because its minimum required DoF is one. (f)-(i)~Examples of component-wise translation estimation~(COTE) to estimate the relative $x$ value. (f) Boundary interval set ${^1\mathcal{E}}$, where the brackets [ and ] denote the lower and upper bounds, respectively. (g)~Cardinality of the $g$-th consensus set,~$\abs{{^1\mathcal{I}}_g}$. (h)~Each value of the optimal solutions, i.e.~${^1\hat{\mathbf{t}}}_g$, which is associated with each ${^1\mathcal{I}}_g$. Note that ${^1\hat{\mathbf{t}}}_{14}$ is not assigned because ${ }^1 \mathcal{I}_{14}=\varnothing$. (i)~Residual of the objective function for all correspondences when ${^1\mathbf{t}}_g$ is given. The residual becomes lowest when $g = 8$, thus ${^1 \hat{\mathbf{t}}}_{8}$ is selected as the final solution for the 1st element, i.e. ${^1 \hat{\mathbf{t}}} \leftarrow {^1 \hat{\mathbf{t}}}_{8}$ (best viewed in color).}
	\label{fig:illustration}
\end{figure*}
%%%%%%%%%%%%%%%%%%%%%%%%%%%%%%

Second, rejecting these ground points in advance reduces the computational cost of the subsequent extraction and matching procedure because almost 40$\sim$60~\% of points in a point cloud are from the ground once the point clouds are captured by the ground vehicles~\citep{lim2021patchwork}.
As a result, ground segmentation usually reduces the number of cloud points.

\subsection{Feature Extraction and Matching}

Next, feature extraction and matching are employed to estimate correspondences. 
In this study, fast point feature histogram~(FPFH) is exploited to generate descriptors~\citep{rusu2009fpfh}.
Of course, any relevant feature extraction and matching method can be used, yet we chose FPFH-based matching to guarantee the generality without any training procedure and to emphasize that global registration must be robust even though imprecise correspondences are given.
For these reasons, improving feature extraction and matching itself is beyond our scope. 

The procedure to estimate correspondences is summarized as follows. 
Given the estimated non-ground points of source and target, voxel sampling is firstly employed with voxel size~$\nu$. 
% By doing so, hundreds of thousands of points are approximated into tens of thousands of points. 
% As a result, the computational burden of the following algorithms decreases. 
Then, FPFH is used by taking a voxel sampled cloud as an input to generate the feature descriptor of each point with the radius for normal estimation,~$r_\text{normal}$, and that for FPFH descriptor, $r_\text{FPFH}$. 
Finally, the reciprocal test~\citep{zhou2016fast} is performed to obtain $\mathcal{A}_{\text{raw}}$, as represented in Fig.~\ref{fig:illustration}(a). 
More details about the parameters are presented in Section~\ref{sec:paramter_of_q}.

\subsection{Graph-Based Outlier Rejection}

Next, MCIS~\citep{rossi2013pmc} is applied to reject outlier correspondences in advance. Given $\mathcal{A}_{\text{raw}}$ as an input, MCIS outputs the filtered correspondences,~$\mathcal{A}$ (see Fig.~\ref{fig:illustration}(b))
However, finding the exact maximal clique \color{black} is NP-complete and its complexity increases in exponential time, so it is \color{black} likely to be a bottleneck in the pipeline and thus occasionally takes lots of time.
For this reason, we modify it to heuristically find a maximal clique, named MCIS-\texttt{heuristic} to reduce computational burden.
\color{black} That is, we estimate the maximal clique found within a threshold time as the optimal result.
More details about MCIS can be found in \cite{yang2020teaser} and \cite{yin2023segregator}.
\color{black}

\subsection{\color{black} Overview of \color{black} Decoupling Rotation and Translation Estimation}\label{sec:decoupling}

After initially pruning outlier correspondences, the relative rotation and translation are estimated, respectively. Sharing the philosophy of \cite{horn1987closed} and \cite{arun1987least} that $\hat{\mathbf{R}}$ and $\hat{\mathbf{t}}$ can be estimated in a decoupled manner, the relative rotation is firstly estimated in the translation-invariant space, followed by translation estimation. Note that this decoupling estimation strongly assumes that the relative rotation is successfully estimated~\citep{yang2020teaser}.

Here, a brief explanation of the decoupling method is presented~\citep{lim2022single}. 
\color{black}Let us assume that $\mathcal{A}$ is a tuple that has an order. \color{black}
Then, to cancel out the effect of relative translation, two consecutive pairs ($i$, $j$) corresponding to \color{black}the $n$-th element of $\mathcal{A}$ and $(i^\prime, j^\prime)$ corresponding to the ($n+1$)-th element of $\mathcal{A}$ \color{black} are subtracted in a chain form, which are expressed as $\boldsymbol{\alpha}_k~=~\mathbf{p}_{i^{\prime}} - \mathbf{p}_{i}$ and $\boldsymbol{\beta}_k~=~\mathbf{q}_{j^{\prime}} - \mathbf{q}_{j}$, respectively.
By doing so, the number of correspondences is preserved, preventing an increase in the computational cost.
Note that the last element is made by the subtraction between the last and first pair in~$\mathcal{A}$.

$\boldsymbol{\alpha}_k$ and $\boldsymbol{\beta}_k$ are referred to as translation invariant measurement\color{black}s (TIMs) that satisfy \color{black} $\boldsymbol{\beta}_{k}=\mathbf{R}\boldsymbol{\alpha}_{k}+\boldsymbol{\epsilon}_{k}$, whose translation term is canceled out;~$\boldsymbol{\epsilon}_k$ denotes Gaussian noise if both ($i$, $j$) and ($i^\prime$, $j^\prime$) pairs are true inliers and $\boldsymbol{\epsilon}_k$ has a large irregular error, otherwise.

Then, the rotation is estimated in translation invariant space as follows:

\begin{equation}
	\hat{\mathbf{R}}=\underset{\mathbf{R} \in \mathrm{SO}(3)}{\argmin } \sum_{k=1}^{K} \min \Big( w_k r({\boldsymbol{\beta}}_{k}- \mathbf{R}{\boldsymbol{\alpha}}_{k}), \, \bar{c}^{2}\Big)
	\label{eqn:decouple_rot}
\end{equation}

\noindent where $K$ denotes the cardinality of $\mathcal{A}$, i.e.~$\abs{\mathcal{A}}$; $w_k$ is the weight to control the influence of each TIM pair; $\bar{c}$~is a truncation parameter. Thus, (\ref{eqn:decouple_rot}) means that if either $\boldsymbol{\alpha}_{k}$ or $\boldsymbol{\beta}_{k}$ is outlier,  
$r({\boldsymbol{\beta}}_{k}- \mathbf{R}{\boldsymbol{\alpha}}_{k})$ has large residual. Then, $w_k$ is exploited to suppress the effect of the potential outlier. However, if the first summand, i.e.~$w_k r({\boldsymbol{\beta}}_{k}- \mathbf{R}{\boldsymbol{\alpha}}_{k})$, is still large, this residual term is substituted with the predefined constant $\bar{c}^{2}$ to cut off the influence of the outliers on the estimation. Consequently, the undesirable effect of these potential outliers can be safely suppressed.
\color{black} The details of how to estimate $w_k$ are presented in Section~\ref{sec:rot_estimation}. \color{black}

Then, the rotation estimation is followed by COTE~\citep{yang2020teaser}. As the \color{black} term COTE \color{black} itself indicates, 3D relative translation is estimated in an element-wise manner as follows:

\begin{equation}
	^{l}{\hat{\mathbf{t}}}=\underset{^{l}\mathbf{t} \in \mathbb{R}}{\argmin } \sum_{(i,j) \in \mathcal{A}} \min \bigg( \frac{r({^{l}\mathbf{t}}-{^{l}\mathbf{v}_{ij}})}{\sigma_{ij}^2}, \, {\bar{c}^{2}} \bigg)
	\label{eqn:decouple_trans}
\end{equation}

\noindent where \color{black} $\mathbf{v}_{ij}=\mathbf{q}_{j}-{\hat{\mathbf{R}}} \mathbf{p}_{i}$ denotes the translation discrepancy, $\sigma_{ij}$ is the noise bound, and $^l(\cdot)$ denotes the $l$-th element of a 3D vector. \color{black} That is, $l = 1,2,3$ and each value corresponds to $x$, $y$, and $z$ translation in the ascending order, respectively.

In summary, (\ref{eqn:final_goal}) is decomposed into (\ref{eqn:decouple_rot}) and (\ref{eqn:decouple_trans}). More details are explained in Sections~\ref{sec:rot_estimation} and~\ref{sec:trans_estimation}.

% Thus, (\ref{eqn:goal}) is paraphrased into optimization of {(\ref{eqn:decouple_rot})} (Section~\rom{2}.\textit{D}) followed by optimization of {(\ref{eqn:decouple_trans})} using ${\mathbf{R}}^*$ (Section~\rom{2}.\textit{E}).

\subsection{Quasi-SO(3) in Urban Environments} \label{sec:quasi_so3_sacrifices_rp}

\begin{figure}[t!]
	\centering
	\begin{subfigure}[b]{0.46\textwidth}
		\includegraphics[width=1.0\textwidth]{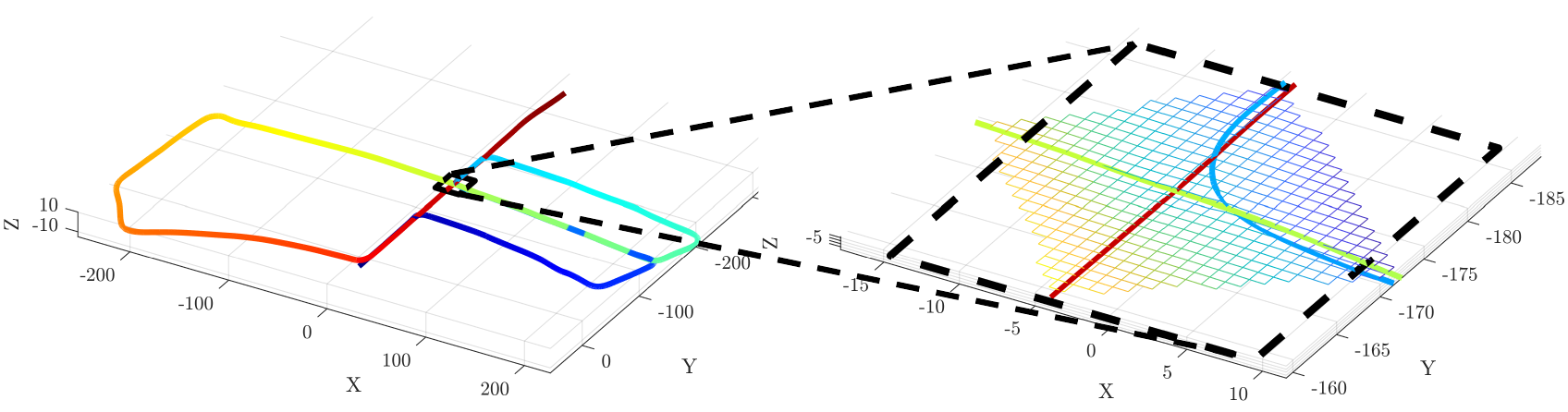}
		\caption{\centering}
	\end{subfigure}
	\begin{subfigure}[b]{0.23\textwidth}
		\includegraphics[width=1.0\textwidth]{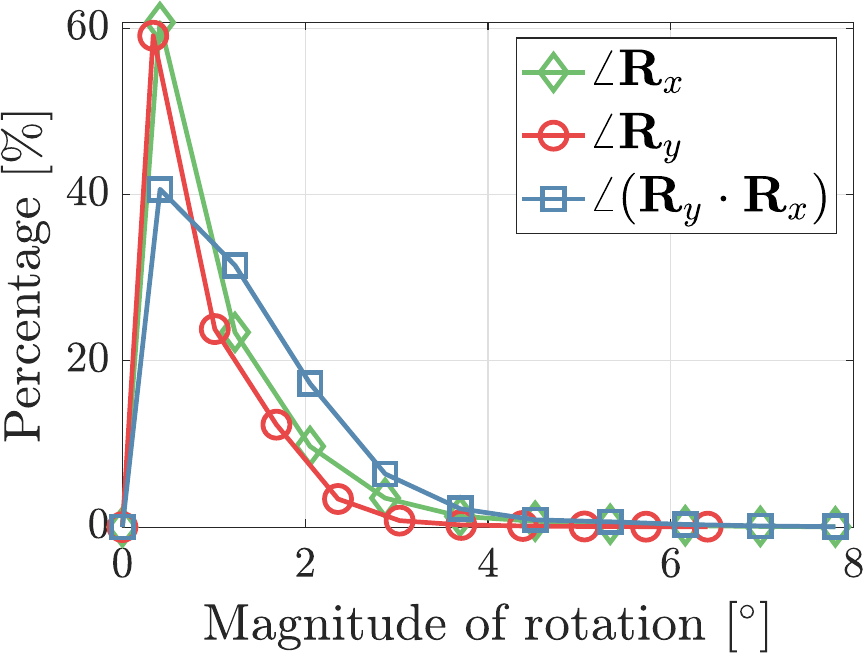}
		\caption{\centering}
	\end{subfigure}
	\begin{subfigure}[b]{0.23\textwidth}
		\includegraphics[width=1.0\textwidth]{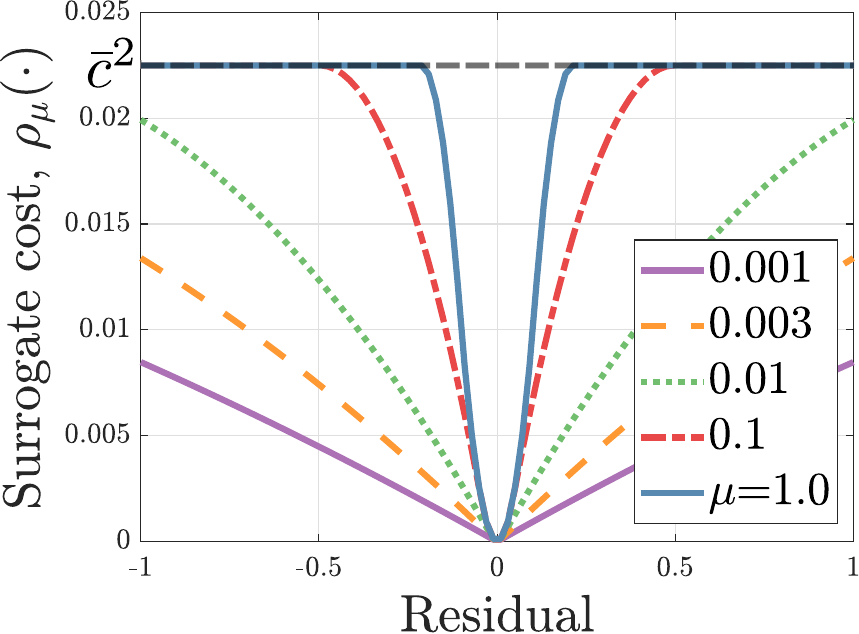}
		\caption{\centering}
	\end{subfigure}
	\captionsetup{font=footnotesize}
	\caption{(a) Visual description to show the geometrical characteristics of loop closing situations in urban canyons. (a) Revisit usually occurs in either forward, reverse, or perpendicular direction. Accordingly, the roll and pitch differences between the two viewpoints of source and target clouds are sufficiently small. (b) Qualitative analysis of the actual geodesic magnitude of relative roll ($\angle \mathbf{R}_x$), pitch ($\angle \mathbf{R}_y$), and the combined rotations ($\angle (\mathbf{R}_y \cdot \mathbf{R}_x)$) except yaw rotation, where $\angle \mathbf{R}=\cos^{-1}{\frac{\Tr(\mathbf{R}) - 1}{2}}$. The loop pairs whose distances are between 0.5 to 10~m apart in the KITTI dataset were used. (c) Surrogate function $\rho_{\mu}(\cdot)$ when $\bar{c}=0.15$. As the parameter $\mu$ gradually grows, the kernel has more non-linearity. It becomes truncated least squares once $\mu$ reaches $\infty$ (best viewed in color).}
	\label{fig:exp_evidences}
\end{figure}

Next, the concept of quasi-SO(3) is proposed based on the premise in Section~\ref{sec:overview}. 
To be more robust against degeneracy, Quasi-SO(3) aims to reduce the DoF of rotation estimation from three to one, which is built upon a key observation that a pure yaw rotation is relatively dominant than roll and pitch rotations in a loop closing situation of urban environments.
As shown in Fig.~\ref{fig:exp_evidences}(a), once a revisit occurs, the roll and pitch differences between the source and target clouds are not large. 
Accordingly, two viewpoints of the source and target clouds are likely to be located in the coplanar space. 

The experimental evidence also supports our rationale that the rotation affected by roll and pitch motions is negligible in loop closing situations. 
To be more concrete, the probability distribution function empirically shows that the effect of roll and pitch rotations is usually smaller than $8^\circ$~(Fig.~\ref{fig:exp_evidences}(b)). 
In general, the error of small angle assumption does not exceed 1.0 \% if the angle is lower than $8^\circ$ \citep{youn2021state}. 
In that respect, it was demonstrated that these angles are sufficiently insignificant. 

Therefore, let us express the relative rotation as $\mathbf{R}=\mathbf{R}_{z} \cdot \mathbf{R}_{y} \cdot \mathbf{R}_{x}$, where $\mathbf{R}_{z}$, $\mathbf{R}_{y}$, and $\mathbf{R}_{x}$ denote the yaw, pitch, and roll rotations, respectively. 
Then, $\mathbf{R}$ can be approximated as $\mathbf{R} \approx \mathbf{R}_{z}$ based on our assumption that the magnitudes of roll and pitch angles are insignificant in urban canyons~\citep{chen2022overlapnet}, so the effect of $\mathbf{R}_{y}$ and  $\mathbf{R}_{x}$ can be negligible as~$\mathbf{R}_{y} \cdot \mathbf{R}_{x} \approx \mathbf{I}_3$, where $\mathbf{I}_n$ denotes the $n \times n$ identity matrix. 
Consequently, the DoF of the relative rotation is reduced from three to one through this assumption.
For brevity, the approximated $\mathbf{R}$ is denoted by~$\mathbf{R}_{+}$.

One might argue that our assumption occasionally does not hold once the pose discrepancy is somewhat large and the loop closing is performed in non-flat areas. 
However, this problem can be easily resolved by utilizing the estimated roll and pitch angles from an inertial navigation system (INS).
\color{black} We propose to \color{black} tackle this problem using the INS in Section~\ref{sec:ins}.

\subsection{Quasi-SO(3) Estimation using Graduated Non-Convexity to Avoid Degeneracy}\label{sec:rot_estimation}

Based on the assumption above, GNC-based truncated least square~\citep{tzoumas2019outlier} is explained to estimate $\mathbf{R}_{+}$.
To this end, (\ref{eqn:decouple_rot}) can be modified as $\argmin _{\mathbf{R}_{+}} \sum_{k=1}^{K} \rho_{\mu}\left(r\left( \boldsymbol{\beta}_{k}-\mathbf{R}_{+}\boldsymbol{\alpha}_{k}\right)\right)$ by using a surrogate function, $\rho_{\mu}(\cdot)$, that controls the non-linearity of loss function governed by parameter $\mu$ (Fig.~\ref{fig:exp_evidences}(c)).

Next, this equation is redefined based on Black-Rangarajan duality as follows~\citep{zhou2016fast, yang2020teaser}:
\begin{equation}
\hat{\mathbf{R}}_{+} = \argmin_{\begin{array}{c}\scriptstyle \mathbf{R}_{+} \in \mathrm{SO}(3); \\[-4pt]
\scriptstyle w_{k} \in[0,1]
\end{array}
}
\sum_{k=1}^{K}\left[w_{k} r\left(\boldsymbol{\beta}_{k}-\mathbf{R}_{+}\boldsymbol{\alpha}_{k}\right)+\Phi_{\rho_{\mu}}\left(w_{k}\right)\right]
\label{eq:black_rangarajan}
\end{equation}
\noindent where $\Phi_{\rho_{\mu}}\left(w_{k}\right)=\frac{\mu\left(1-w_{k}\right)}{\mu+w_{k}} \bar{c}^{2}$ is a penalty term \citep{yang2020gnc, lim2022single}. 
However, minimizing the residuals while simultaneously assigning the corresponding weights does not guarantee optimality~\citep{zhou2016fast}. 

To tackle this problem, the alternating optimization is exploited to solve (\ref{eq:black_rangarajan}).
That is, the optimal rotation is estimated with the fixed weights updated in the previous iteration and then the weights are updated with the fixed optimized rotation. 
To be more concrete, the alternating optimization is expressed as follows:

\begin{equation}
	\hat{\mathbf{R}}^{(t)}_{+}=\underset{\mathbf{R}_{+} \in \mathrm{SO}(3)}{\argmin} \sum_{k=1}^{K} \hat{w}^{(t-1)}_{k} r\left(\boldsymbol{\beta}_{k}-\mathbf{R}_{+}\boldsymbol{\alpha}_{k}\right),
	\label{eq:rotation}
\end{equation}

\begin{equation}
	\hat{\mathbf{W}}^{(t)}=\underset{w_{k} \in[0,1]}{\argmin } \sum_{k=1}^{K}\left[w_{k} r\left(\boldsymbol{\beta}_{k}-\hat{\mathbf{R}}^{(t)}_{+}\boldsymbol{\alpha}_{k}\right)+\Phi_{\rho_{\mu}}\left(w_{k}\right)\right]
	\label{eq:weight}
\end{equation}

\noindent where the superscript ${(t)}$ denotes the $t$-th iteration and $\hat{\mathbf{W}}$ is a matrix form of the weights, i.e. $\hat{\mathbf{W}}=\operatorname{diag}\left(\hat{w}_{1}, \hat{w}_{2}, \ldots, \hat{w}_{K}\right)$. Each \textcolor{qwr}{$\hat{w}_{k}$} can be obtained by solving the following equation:

\begin{equation}
	\frac{\partial}{\partial w_{k}} \big(w_{k} r\left(\boldsymbol{\beta}_{k}-\hat{\mathbf{R}}^{(t)}_{+}\boldsymbol{\alpha}_{k}\right) + \Phi_{\rho_{\mu}}\left(w_{k}\right) \big) = 0.
	\label{eq:closes_form}
\end{equation}

\noindent \color{black} Once we expand \eqref{eq:closes_form} in terms of $\hat{w}_{k}$, \color{black} $\hat{w}_{k}$ is expressed in a truncated manner as follows:

\begin{equation}
	\hat{w}_{k}^{(t)}= \begin{cases}0 & \text { if } \tilde{r}_{k} \in\left[\frac{\mu+1}{\mu} \bar{c}^{2},+\infty\right) \\ \bar{c} \sqrt{\frac{\mu(\mu+1)}{\tilde{r}_{k}}}-\mu & \text { if } \tilde{r}_{k} \in\left[\frac{\mu}{\mu+1} \bar{c}^{2}, \frac{\mu+1}{\mu} \bar{c}^{2}\right) \\ 1 & \text { otherwise } \end{cases}
	\label{eq:truncated_weight}
\end{equation}

\noindent where $\tilde{r}_k \geq 0$ is the simplified notation of $r(\boldsymbol{\beta}_k- \mathbf{R}_{+}^{(t)}\boldsymbol{\alpha}_k)$. 
% The overall procedure of quasi-SO(3) estimation is illustrated in Figs.~\ref{fig:illustration}(c). 

Note that the value of $\mu$ grows for every iteration to gradually increase the non-linearity of the loss function. That is, $\mu$ is assigned as $\mu^{(t)} \leftarrow \kappa \cdot \mu^{(t-1)}$, where $\kappa~>~1$ is a constant parameter. By doing so, the kernel becomes more non-linear, as presented in Fig.~\ref{fig:exp_evidences}(c). 
For initialization, $\mu^{(0)}$ is set to be ${\bar{c}^2}/({\max(r(\boldsymbol{\beta}_k -\boldsymbol{\alpha}_k))-\bar{c}^2})$. The optimization ends if either the differential of the loss function is sufficiently small or the iteration reaches the maximum number, $N_\text{iter}$.

In summary, the estimation of $\mathbf{R}_{+}$ has two major advantages over that of $\mathbf{R}$. 
First, estimation of the $\mathbf{R}_{+}$ helps to avoid the degeneracy situations in itself by reducing the DoF of rotation from three to one. 
Sometimes, the number of remaining inliers is less than three, because GNC does not guarantee to leave more than three inliers.
Even though the degeneracy occurs, our proposed rotation estimation enables to tolerate the case where only one or two correspondences remain because a single correspondence is enough to estimate $\mathbf{R}_{+}$.

\begin{figure}[t!]
 	\centering 
 	\begin{subfigure}[b]{0.15\textwidth}
	 		\includegraphics[width=1.0\textwidth]{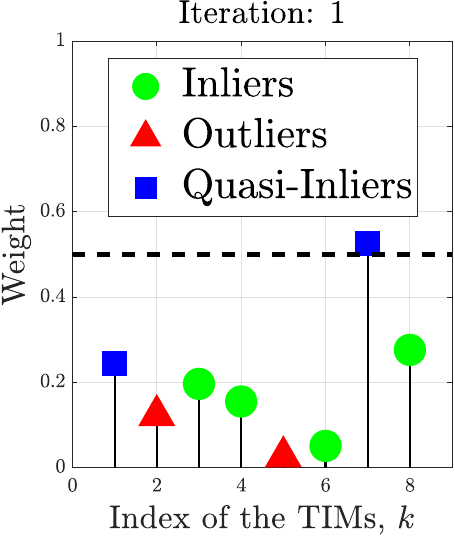}
	 	\end{subfigure}
 	\begin{subfigure}[b]{0.15\textwidth}
	 		\includegraphics[width=1.0\textwidth]{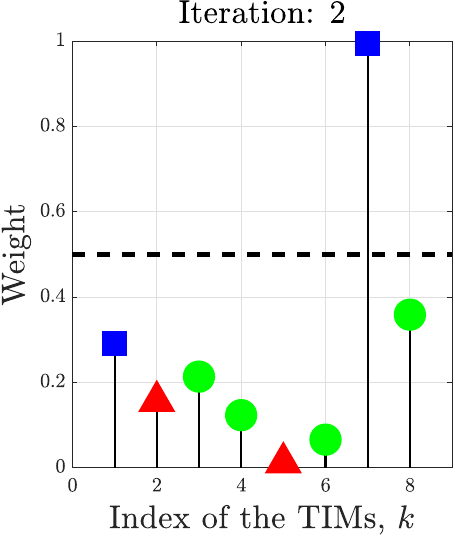}
	 	\end{subfigure}
 	\begin{subfigure}[b]{0.15\textwidth}
	 		\includegraphics[width=1.0\textwidth]{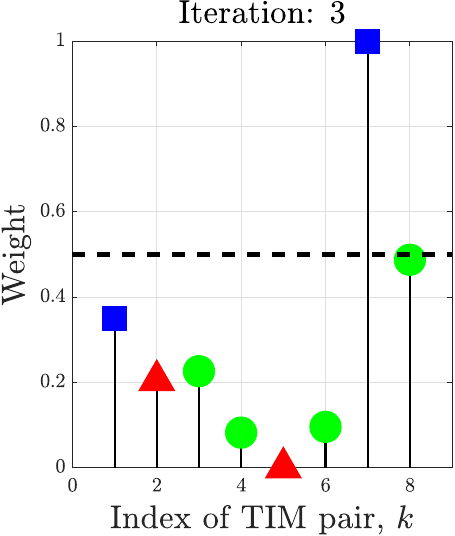}
	 	\end{subfigure}
 	\vspace{-0.1cm}
 	\captionsetup{font=footnotesize}
 	\caption{Changes of weights by GNC in quasi-SO(3) estimation in a degeneracy case: under the situation where the points having weights below 0.5 (dotted line) are filtered out as outliers, quasi-inliers help to estimate quasi-SO(3) by acting as inliers, thus our proposed method enables to tolerate the degeneracy.}
 	\label{fig:inlier_weight_analysis}
 \end{figure}

It may be thought that this phenomenon does not occur often; 
however, as shown in Figs.~\ref{fig:illustration}(d), ~\ref{fig:illustration}(e), and \ref{fig:inlier_weight_analysis}, the degeneracy occurs once the distance between the two viewpoints becomes more than 9 meters away in outdoor scenes, empirically. 
Therefore, being robust against the degeneracy can be interpreted as being able to help the convergence of local registration better when performing loop closing in the distant cases~(see Sections~\ref{sec:why_quatro} and~\ref{sec:exp_slam}).

Second, $\mathbf{R}_{+}$ allows measurements that are originally outliers to be treated as inliers. To explain this reason, we introduce the concept of \textit{quasi-inliers} each of whose error only exists along the perpendicular direction of the ground plane. That is, let us decompose the noise term in the translation invariant space into two components: a) a perpendicular term and b)~a parallel term to the ground plane. Then, the quasi-inlier represents the measurement whose former error term has a value greater than zero, yet the latter one has a near-zero value. 

Note that these quasi-inliers often occur owing to the ambiguity of the descriptors in urban environments. That is, the majority of urban structures are generally orthogonal to the ground, so the local geometrical properties along the perpendicular direction are highly likely to be similar, resulting in false matching.

The point is that the quasi-inliers behave like the actual inliers during the estimation of $\mathbf{R}_{+}$. 
That is, the perpendicular error terms rarely affect the estimation of $\mathbf{R}_{+}$ because $\mathbf{R}_{+}$ primarily considers the effect of yaw rotation. %, so $\mathbf{R}_{+}$ is robust against the error terms along the perpendicular direction. 
In other words, the estimation of $\mathbf{R}_{+}$ is equivalent to the estimation of yaw rotation with the measurements projected into the $xy$-plane, so the effect of error along the perpendicular direction can be negligible.

In short, as these quasi-inliers play a role as the actual inliers, the quasi-inliers help to increase the number of inliers and prevent degeneracy, as shown in Fig.~\ref{fig:inlier_weight_analysis}. 

%%%%%%%%%%%%%%%%%%%%%%%%%%%5 

%%%%%%%%%%%%%%%%%%%%%%%%%%%5 

\subsection{Component-Wise Translation Estimation}\label{sec:trans_estimation}

Finally, COTE is exploited to estimate the component-wise relative translation by the following four steps. For each $l$-th element, boundary interval set, which is a $2\abs{\mathcal{A}}$-tuples and denoted as ${^{l}\mathcal{E}}$, is firstly set which comprises the lower bound, ${^{l}\mathbf{v}}_{ij} - \sigma_{ij}\bar{c}$, and the upper bound, $^{l}\mathbf{v}_{ij} + \sigma_{ij}\bar{c}$ ($\boldsymbol{[}$ and $\boldsymbol{]}$ in Fig.~\ref{fig:illustration}(f), respectively). Note that it is assumed that all the elements of $^{l}\mathcal{E}$ are sorted in ascending order.
That is, let the $g$-th element in ${^{l}\mathcal{E}}$ be ${^{l}\mathcal{E}}(g)$, whose value could be the lower or upper bound, then it satisfies ${^{l}\mathcal{E}}(g)<{^{l}\mathcal{E}}(g+1)$ for $g=1,2,\dots,2\abs{\mathcal{A}}-1$.

Second, the $g$-th consensus set is assigned as follows: 

\begin{equation}
^{l}\mathcal{I}_g=\{(i,j)| \frac{({^{l}\phi_{g}}-{^{l}\mathbf{v}}_{ij})^{2}}{\sigma_{ij}^{2}} \leq \bar{c}^{2}\}
\end{equation}

\noindent where ${^{l}\phi_{g}}\in\mathbb{R}$ denotes an arbitrary value which satisfies \textcolor{qwr}{${^{l}\mathcal{E}}(g)<{^{l}\phi_{g}}<{^{l}\mathcal{E}}(g+1)$}. % for $g=1,2,\dots,2\abs{\mathcal{A}}-1$~(Fig.~\ref{fig:illustration}(g)).

Third, the optimal solution for each non-empty $^{l}\mathcal{I}_g$, which is denoted as $^{l}{\hat{\mathbf{t}}}_g$, is estimated as follows:

\begin{equation}
	^l{\hat{\mathbf{t}}_{g}}=\Big(\sum_{(i, j) \in ^l\mathcal{I}_{g} } \frac{1}{\sigma_{ij}^2}\Big)^{-1}\sum_{(i, j) \in \mathcal{I}_{g}} \frac{1}{\sigma_{ij}^2} {^l\mathbf{v}}_{ij}.
	\label{eq:optimal_solution_for_each_consensus}
\end{equation}

Finally, among the optimal solution candidates estimated by (\ref{eq:optimal_solution_for_each_consensus}), the one that minimizes the truncated objective function, i.e.~(\ref{eqn:decouple_trans}), is chosen as the global optimum~(the red dashed rectangle in Fig.~\ref{fig:illustration}(i)) as follows:

\begin{equation}
	^{l}{\hat{\mathbf{t}}}=\underset{^{l}\mathbf{t}_g \in \mathbf{H}}{\argmin } \sum_{(i,j) \in \mathcal{A}} \min \bigg( \frac{r({^{l}\mathbf{t}_g}-{^{l}\mathbf{v}_{ij}})}{\sigma_{ij}^2}, {\bar{c}^{2}} \bigg)
	\label{eqn:trans_final}
\end{equation}

\noindent where $\mathbf{H}$ is a set of the optimal solution candidates whose size is up to $2\abs{\mathcal{A}}-1$. Note that COTE presumes that $\hat{\mathbf{R}}_{+}$ is sufficiently precise. 

\section{Quatro++ in Back-End of LiDAR SLAM}

In this section, the details of how Quatro++ is integrated into the LiDAR SLAM frameworks. 
Originally, our proposed method is a front-end agnostic method, so it can be easily exploited in any LiDAR SLAM frameworks.
In doing so, Quatro++ is utilized as an initial alignment method to help the estimate of local registration successfully converge into the global optimum.

\subsection{Generic Pose Graph Optimization in LiDAR SLAM}

As mentioned in Section~\ref{sec:intro}, PGO is used to minimize global trajectory errors, which are caused by local drift of odometry. 
That is, these errors are accumulated as time goes by, resulting in a large pose discrepancy between the estimated poses and the actual poses. 

To tackle this problem, loop detection and loop closing are utilized, which mainly consists of three steps.
First, loop detection is performed to identify inter-related nodes~\citep{kim2018scancontext, chen2022overlapnet, cattaneo2022lcdnet}. 
Second, a registration algorithm is exploited to estimate the actual pose difference. 
Finally, global pose errors are minimized via PGO by utilizing the pose differences estimated by the registration between the two non-consecutive nodes. 

In summary, the objective function of PGO is expressed as follows:

\begin{equation}
	\begin{aligned}
		\hat{\mathcal{X}}=\underset{\mathbf{X} \in \mathcal{X}}{\operatorname{argmin}} & \sum_{i}\left\|\mathbf{X}^{-1}_{i} \mathbf{X}_{i+1} \boxminus \mathbf{z}^{\text{odom}}_{i}\right\|_{\Sigma^\text{odom}_{i} }^{2} \\
		&+\sum_{(j, k) \in \mathcal{L}}\left\|\rho_{j k}\left( \mathbf{X}_{j}^{-1} \mathbf{X}_{k} \boxminus \mathbf{z}^{\text{reg}}_{j k}\right)\right\|_{\Sigma^\text{loop}_{j k} }^{2}
	\end{aligned}
	\label{eqn:pgo_optim}
\end{equation}

\noindent where \color{black} the first summand denotes the summation of Mahalanobis distances estimated by the odometry and the last summand denotes that by the loop closing. \color{black}
The definitions of the variables are as follows: $\mathcal{X}$ and $\hat{\mathcal{X}}$ are sets of 3D poses and the optimized poses, respectively;
$i$, $j$, and $k$ are the indices of the node in a graph structure where the $j$-th and $k$-th nodes are not the successive nodes; 
$\mathcal{L}$ denotes an indices set of loops and $\mathbf{X}_i \in \text{SE}(3)$ is the $i$-th pose that corresponds to the $i$-th node; 
$\boxminus$ denotes box-minus operation~\citep{he2021kalman} which is an equivalent operator of subtraction between two manifolds, i.e. in a 3D space, $\boxminus: \text{SE}(3) \times \text{SE}(3) \rightarrow \mathbb{R}^6$; $\mathbf{z}^\text{odom}_{i}$ \color{black} denotes the measurement from the odometry and $\mathbf{z}^\text{reg}_{jk}$ from the registration; \color{black}
$\Sigma^\text{odom}_{i}$ and $\Sigma^\text{reg}_{jk}$ denote the covariances that correspond to $\mathbf{z}^\text{odom}_{i}$ and $\mathbf{z}^\text{reg}_{jk}$, respectively. 

\subsection{Mean Squared Error-Based False Loop Rejection}

 In general, the loops, i.e. $\mathcal{L}$, can be redefined as $\mathcal{D} \backslash \hat{\mathcal{O}}_{\mathcal{L}}$ where $\mathcal{D}$ is all the loop candidates searched by a loop detection method and $\hat{\mathcal{O}}_{\mathcal{L}}$ denotes the rejected loop candidates by considering them as false loops. 

Most LiDAR SLAM methods~\citep{shan2018lego, shan2020lio, li2021LiLiOM, ramezani2022wildcat} heavily rely on mean squared error~(MSE) of local registration methods to determine whether the loop candidate is considered as a false loop or not, i.e. the output of \texttt{getFitnessScore()} in point cloud library~\citep{rusu20113d,aldoma2012tutorial}. 
The MSE is the average of squared Euclidean distances from the transformed source cloud points by the estimated relative rotation and translation to the corresponding target cloud points~\citep{rusu20113d}.
Accordingly, the MSE indicates how the source cloud is tightly matched to the target cloud.

\begin{figure}[t!]
	\centering 
	\begin{subfigure}[b]{0.15\textwidth}
		\includegraphics[width=1.0\textwidth]{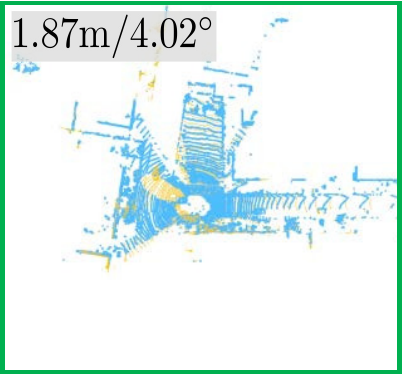}
		\caption{\centering}
	\end{subfigure}
	\begin{subfigure}[b]{0.15\textwidth}
		\includegraphics[width=1.0\textwidth]{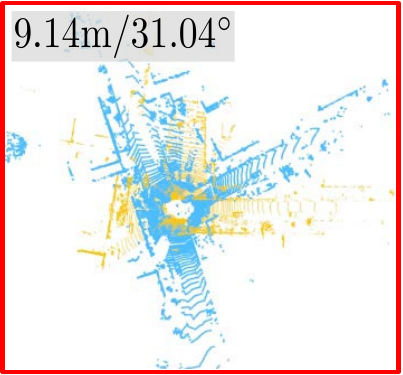}
		\caption{\centering}
	\end{subfigure}
	\begin{subfigure}[b]{0.15\textwidth}
		\includegraphics[width=1.0\textwidth]{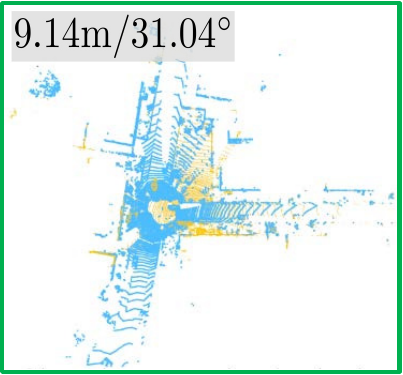}
		\caption{\centering}
	\end{subfigure}
	\vspace{-0.1cm}
	\captionsetup{font=footnotesize}
	\caption{Registration results once (a)-(b) local registration is only applied and (c) a coarse-to-fine alignment is applied by utilizing our proposed method. The texts in the upper left corner represent the initial pose differences. The green and red boxes indicate whether the registration succeeds or not, respectively. The cyan and yellow colors denote the transformed source cloud and the target cloud, respectively (best viewed in color).}
	\label{fig:example_c2f}
	\vspace{-0.3cm}
\end{figure}

Once the estimate successfully converges into the actual global minimum~(Fig.~\ref{fig:example_c2f}(a)), the MSE usually shows a small value because each distance between the target point and the warped source point is near zero. 
In contrast, the MSE shows a larger value in the divergence cases if the estimate goes to the local minimum and thus two point clouds are not aligned~(Fig.~\ref{fig:example_c2f}(b)).
Based on these observations, the loop is rejected if the MSE is somewhat large because a number of distances between the transformed source and target are not zero. 

Thus, this MSE-based rejection effectively filters out most false loops, leaving some actual loops among the loop candidates.

\subsection{Potential Limitations of the Mean Squared Error-Based False Loop Rejection}\label{sec:limit_loop_closing}

% false positive & false negative
 However, there are two problems in terms of a)~false negative and b)~false positive loops. 
 First, the narrow convergence region of the local registration methods triggers too many rejections even though these loops are the actual loops. 
 As mentioned earlier, the local registration methods estimate correspondences based on the strong assumption that the nearest points between the source and target clouds are valid. 
 %This assumption is safe if the pose difference between the viewpoints of the source and target is small and thus two point clouds are sufficiently overlapped. 
 %In that case, the estimate generally converges. 
 However, once the pose difference becomes large and only a small portion of two point clouds is overlapped, this assumption does not hold, so the estimate generally diverges.
 For this reason, many loop candidates whose pose discrepancies are large are highly likely to be rejected, causing lots of false negative loops.
 
 Second, there are some failure cases having low MSEs, causing false positive loops.
 This phenomenon usually occurs in corridor-like scenes where the surroundings have few geometrical characteristics and high ambiguity.
 For this reason, it is hard to discern whether the registration succeeded or not solely based on the MSE.
 As a result, these false positive loops deteriorate the quality of loop constraints.
 That means, $\mathbf{z}^{\text{reg}}_{jk}$~in~(\ref{eqn:pgo_optim}) becomes more imprecise.

\subsection{Why Quatro++?: For More Loops and Accurate Measurements}\label{sec:why_quatro}

To tackle these two problems, our proposed method can be utilized as a coarse alignment to provide an initial guess, helping the local registration methods to successfully estimate the relative pose, as presented in Fig.~\ref{fig:example_c2f}. 
As the performance of Quatro++ is less affected by the initial pose difference, our global registration can initially transform the source cloud to the region where the estimate of local registration can converge into the global minimum. 

In summary, the coarse-to-fine alignment-based loop closing has two main advantages over solely using a local registration method.
First, the loop closing with our proposed method significantly lessens wrong rejection of the loop candidates, i.e. false negatives, owing to the lowered MSEs. 
Second, our proposed coarse-to-fine-based loop closing in itself increases the accuracy of measurements. 
In particular, it was shown that our proposed method is robust against the corridor-like environments, so that the local minima cases can be reduced as well~(see Section~\ref{exp:quat_and_quatpp}).
Consequently, the coarse-to-fine loop closing increases the quality of $\mathbf{z}^{\text{reg}}_{jk}$~in~(\ref{eqn:pgo_optim}). 

Therefore, these two merits result in more accurate and precise PGO results (see Section~\ref{sec:exp_slam}). 

% \begin{figure}[t!]
% 	\centering
% 	\begin{subfigure}[b]{0.46\textwidth}
% 		\includegraphics[width=1.0\textwidth]{figs/local_global_registration_optimum.pdf}
% 	\end{subfigure}
% 	\captionsetup{font=footnotesize}
% 	\caption{Visual description of the local and global registration in the viewpoint of convex optimization, where the space is parameterized by 3D twist, $\zeta \in \mathbb{R}^6$. (a) .}
% 	\label{fig:local_global_registration_optimum}
% \end{figure}
 
\subsection{Quatro++ With Roll-Pitch Compensation Using INS}\label{sec:ins}

As mentioned earlier, our proposed method sacrifices the estimation of relative roll and pitch angles to improve robustness. However, this limitation can be easily resolved by leveraging the INS system. 
That is, the relative roll and pitch angles between the two point clouds are easily obtained by the INS system. 
These angles are highly relevant to gravity direction, which can be fully observable in the INS system. 
In other words, the roll and pitch angles can be estimated with respect to the world frame, which indicates that these estimated angles are drift-free.

Based on these observations, the source point cloud is first transformed into the roll-pitch-compensated frame by using the estimated rotation, $\hat{\mathbf{R}}_{\text{INS}} = \hat{\mathbf{R}}_y \cdot \hat{\mathbf{R}}_x$, where $\hat{\mathbf{R}}_y$ and $\hat{\mathbf{R}}_x$ denote the elements of relative rotation affected by pitch and roll angles, respectively. 
%Then, our it is followed by out global registration. 
The combination of using our proposed method with $\hat{\mathbf{R}}_{\text{INS}}$ is very simple, yet this compensation significantly reduces the relative pose errors while preserving the robustness of our proposed method~(see Sections~\ref{exp:quat_and_quatpp} and~\ref{sec:exp_ins}).

\section{Experiments}

\subsection{Dataset}

We conducted experiments with various datasets to evaluate the performance of Quatro++~(and Quatro). Accordingly, the following datasets were used: KITTI dataset \citep{geiger2012kitticvpr,geiger2013vision} to evaluate the success rate and robustness of global registration, {NAVER LABS localization dataset} \citep{lee2021large_naverlabs} to check the applicability to more sparse LiDAR scans, MulRan dataset~\citep{kim2020mulran} to integrate our proposed method as a loop closing module, and Hilti-Oxford dataset~\citep{zhang2022hilti} to show feasibility of our Quatro++ on hand-held sensor configurations.
These datasets were captured by Velodyne HDL-64E, Velodyne VLP-16, Ouster OS1-64, and HESAI XT32, respectively. 
That is, our experiments were conducted by using various mechanically-spinning-type LiDAR sensors to check the generality and versatility.

We look closely at three points: a)~performance comparison with other state-of-the-art~(SOTA) global registration methods, b)~the effect of ground segmentation on global registration, and c)~application of our proposed method to LiDAR SLAM frameworks.
In particular, we place more emphasis on the distant cases where the viewpoints of two point clouds are far apart. Accordingly, the experiments are further categorized into three parts: a)~loop closing test, b)~odometry test, and c)~augmented rotation test.

\color{black}
That is, first, let us define all the loop pairs as $\mathcal{S}_\text{All}$ as follows:

% \left\{
\begin{equation}
	\begin{split}
	\mathcal{S}_\text{All}= & \Bigl\{(s, t) | r_{\min} \leq \left\|\mathbf{t}_s-\mathbf{t}_t \right\|^2  \leq r_{\max}, \\
	& |s-t| \geq m, \forall s, t \in \mathbb{N}_\text{traj} \Bigl\},
	\end{split}
\end{equation}

\noindent where $s$ and $t$ are the indices of source and target point clouds, respectively;
$\mathbf{t}_s$ and $\mathbf{t}_t$ are the translation vectors of ground truth poses that correspond to the source and target point clouds, respectively;
$r_{\min}$ and $r_{\max}$ are the minimum and maximum distances between the source and target point clouds, respectively;
$m$ is the minimum number of frames between the source and target point clouds to prevent the loop pairs from being too close to each other;
and $\mathbb{N}_\text{traj}$ is the set of all the indices of the point cloud and ground truth pose pairs.

\begin{figure}[t!]
	\captionsetup{font=footnotesize}
	\centering
	\begin{subfigure}[b]{0.40\textwidth}
		\includegraphics[width=1\textwidth, trim={0 5cm 0 5cm}, clip]{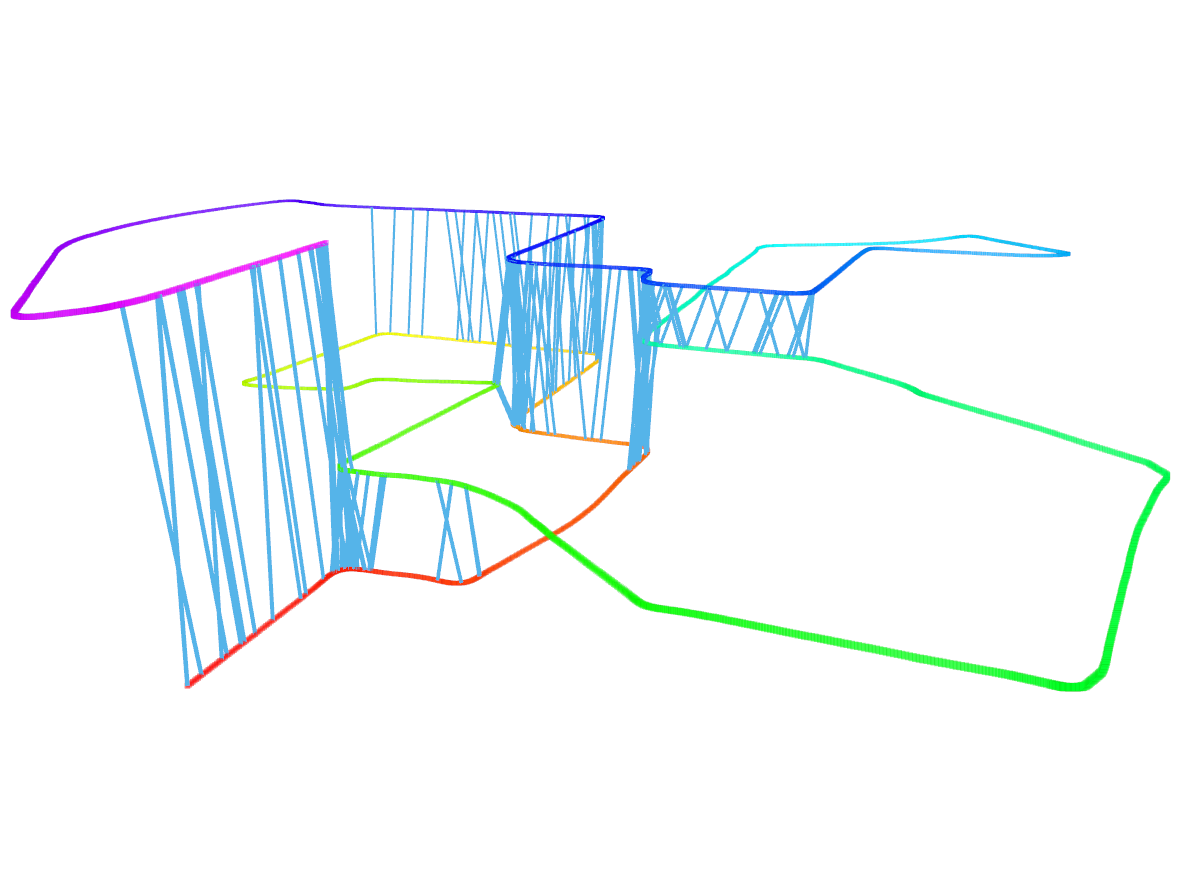}
		\caption{\centering}
	\end{subfigure}
	\begin{subfigure}[b]{0.40\textwidth}
		\includegraphics[width=1\textwidth, trim={0 8cm 0 8cm}, clip]{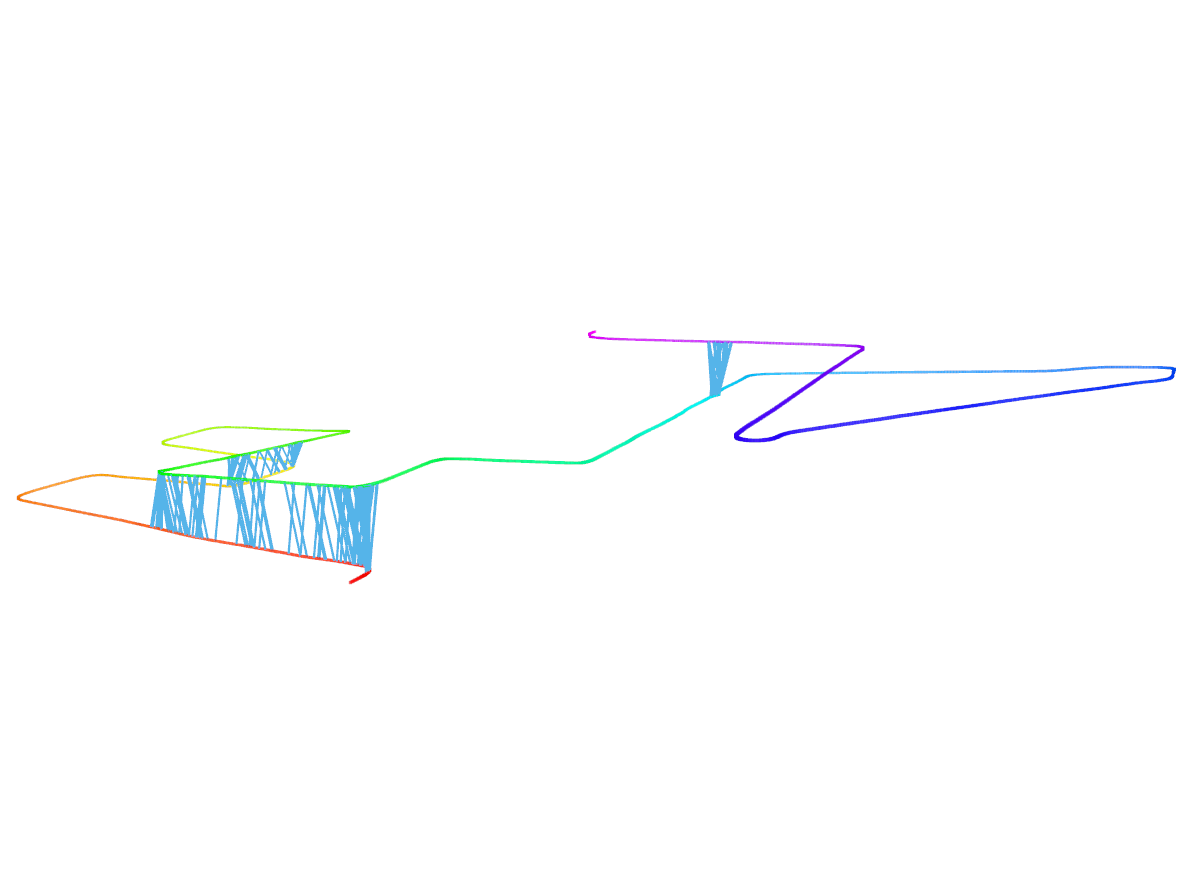}
		\caption{\centering}
	\end{subfigure}
	\caption{\color{black}(a)-(b) Examples of loop pairs of Seq.~\texttt{00} and Seq.~\texttt{08} in the KITTI dataset, each of whose distances is between 10 and 12\,m, i.e. $r_{\min} = 10$ and $r_{\max} = 12$\,m. The rainbow color indicates the poses over time and the trajectory is airborne for better understanding. The cyan-colored lines are the sampled loop pairs, $\mathcal{S}_L$, whose size is 1,000, between two poses in loop closing situations~(best viewed in color).}
	\label{fig:kitti_loop_viz}
\end{figure}

Then, loop pairs for the loop closing test, $\mathcal{S}_L$, is sampled from $\mathcal{S}_\text{All}$, i.e.~$\mathcal{S}_L \subset \mathcal{S}_\text{All}$,
where $\abs{\mathcal{S}_L} = N_\text{sample}$. In this paper, we set $N_\text{sample} = 1,000$ and the symbol ``A $\sim$ B'' denotes that $r_{\min} = \text{A}$ and $r_{\max} = \text{B}$.
$\mathcal{S}_L$ is presented in Fig.~\ref{fig:kitti_loop_viz} for better understanding.
In this paper, the maximum distance we aim to achieve is set to 12~m, i.e.~B$=12$, which is double the criteria used in previous researches~\citep{kim2018scancontext, cattaneo2022lcdnet} when setting true loop pairs.
Thus, our experiments present the robustness of registration against large pose discrepancy.
In addition, when the pose difference exceeds 12~m, two point clouds overlap so partially that the output of local registration as fine alignment may not reach the global minimum, in turn leading to inaccurate loops.

In case of the odometry test, the set of pairs, $\mathcal{S}_O$, is defined as follows:

\begin{equation}
	\begin{split}
	\mathcal{S}_O= & \Bigl\{(\Delta n + 1, \Delta (n+1) + 1) | n \in \mathbb{Z}, \\
	& 0 \leq n < \frac{\mathbb{N}_{\text{traj}}-1}{\Delta} - 1 \Bigl\},
	\end{split}
	\label{eq:odom_pairs}
\end{equation}

\noindent where $\Delta > 0$ denotes the interval between two frames. Finally, the pairs for augmented rotation test, $\mathcal{S}_A$, is similar to~$\mathcal{S}_O$, but we additionally rotate the target cloud along the yaw direction to check the robustness against large angle discrepancy.

\color{black}

\subsection{Error Metrics}

To evaluate algorithms quantitatively, both average pose errors (average translation error, $t_{\text{avg}}$, and average rotation error, $r_{\text{avg}}$) and the relative pose errors~(relative translation error, $t_\text{rel}$, and relative rotation error, $r_\text{rel}$) were exploited; $t_{\text{avg}}$ and $r_{\text{avg}}$ are defined as follows:

\begin{itemize}
	\item $t_{\text{avg}}= \sum_{n=1}^{N} (\mathbf{t}_{n, \text{GT}}-{\hat{\mathbf{t}}}_{n})^{2} / N$,
	\item $r_{\text{avg}}= \frac{180}{\pi} \cdot \sum_{n=1}^{N} \cos^{-1} (\frac{\operatorname{Tr}\left({\hat{\mathbf{R}}}_{n}^{\intercal} \mathbf{R}_{n, \text{GT}}\right)-1}{2}) / N  $ 
\end{itemize}
\noindent where $\mathbf{t}_{n, \text{GT}}$ and $\mathbf{R}_{n, \text{GT}}$ are the $n$-th ground truth translation and rotation, respectively; 
$N$ denotes the number of total samples; 
$t_\text{rel}$ and $r_\text{rel}$ are used for the odometry test, which are calculated by RPG evaluation tools~\citep{Zhang18rpg_eval_tool}.

\color{black}
Furthermore, we evaluate global registration methods in terms of success rate, which is a more important metric to directly evaluate the robustness of the global registration needed in loop closing situations of LiDAR SLAM.
The criteria of the success rate is based on~\cite{kim2019gp}: a pose error that is sufficiently within the narrow convergence region where the estimate of local registration successfully converges into the global optimum. Thus, the registration is counted as a success if both translation and rotation errors are lower than 2~m and 10$^\circ$, respectively.
\color{black}

\subsection{Parameters of Quatro++}\label{sec:paramter_of_q}

Parameter settings of our proposed method in our experiments are presented in Table~\ref{table:quatro_param}. 
It should be noted that parameters of FPFH are set to satisfy the following relations: $\nu < r_\text{normal} < r_\text{FPFH}$. 
Further implementation details can be found in our open-source code.

\begingroup
\begin{table}[b!]
	\captionsetup{font=footnotesize}
	\centering
	\caption{Parameters of each module in our global registration pipeline. Note that the parameters of FPFH  should be set differently depending on the sensor configuration or the number of laser scans. \color{black}The values before, in the middle of, and after the slash are the parameters for Velodyne VLP-16, HDL-64E, and Ouster OS1-64 sensors, respectively, to consider different sparsity characteristices, i.e. differences of vertical resolution between two adjacent rays. \color{black}}
	\setlength{\tabcolsep}{3pt}
	{\scriptsize
		\begin{tabular}{l|l|ccc}
			\toprule \midrule
			& & Param. & Description & Value  \\ \midrule
			\parbox[t]{2mm}{\multirow{6}{*}{\rotatebox[origin=c]{90}{Quatro++}}} & 
			\parbox[t]{2mm}{\multirow{3}{*}{\rotatebox[origin=c]{90}{FPFH}}} &$\nu$  & Voxel sampling size & 0.1 / 0.3 / \color{black} 0.6 \color{black} m   \\
			 & &$r_\text{normal}$  & Radius for normal estimation & 0.3  / 0.5  / \color{black} 1.5 \color{black} m   \\
			 & &$r_\text{FPFH}$  & Radius of FPFH descriptor      & 0.45 / 0.65 / \color{black} 2.25 \color{black} m  \\  \cmidrule{2-5}
			& \parbox[t]{2mm}{\multirow{3}{*}{\rotatebox[origin=c]{90}{Quatro}}} &$\bar{c}$  & Noise bound & 0.3   \\
			 & &$N_\text{iter}$  & Maximum iteration of quasi-SO(3) estimation & 50 \\
			 & &$\kappa$  & A factor to control the increase of GNC & 1.4 \\ \midrule \bottomrule
		\end{tabular}
	}
	\label{table:quatro_param}
\end{table}
\endgroup

\section{Experimental Results}

\begin{figure*}[t!]
	\captionsetup{font=footnotesize}
	\centering
	\begin{subfigure}[b]{0.30\textwidth}
		\includegraphics[width=1\textwidth]{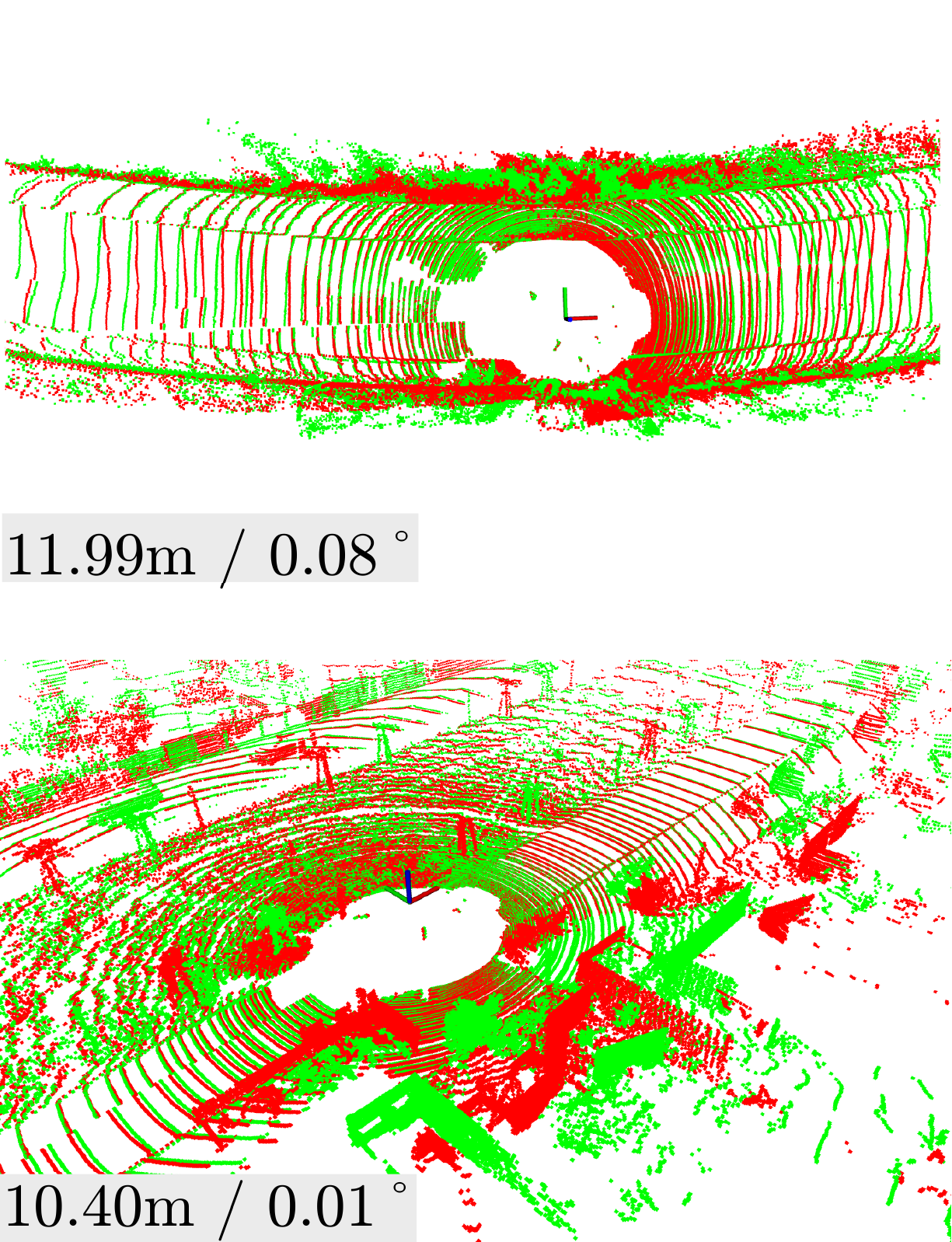}
		\caption{\centering Source and target}
	\end{subfigure}
	\begin{subfigure}[b]{0.30\textwidth}
		\includegraphics[width=1\textwidth]{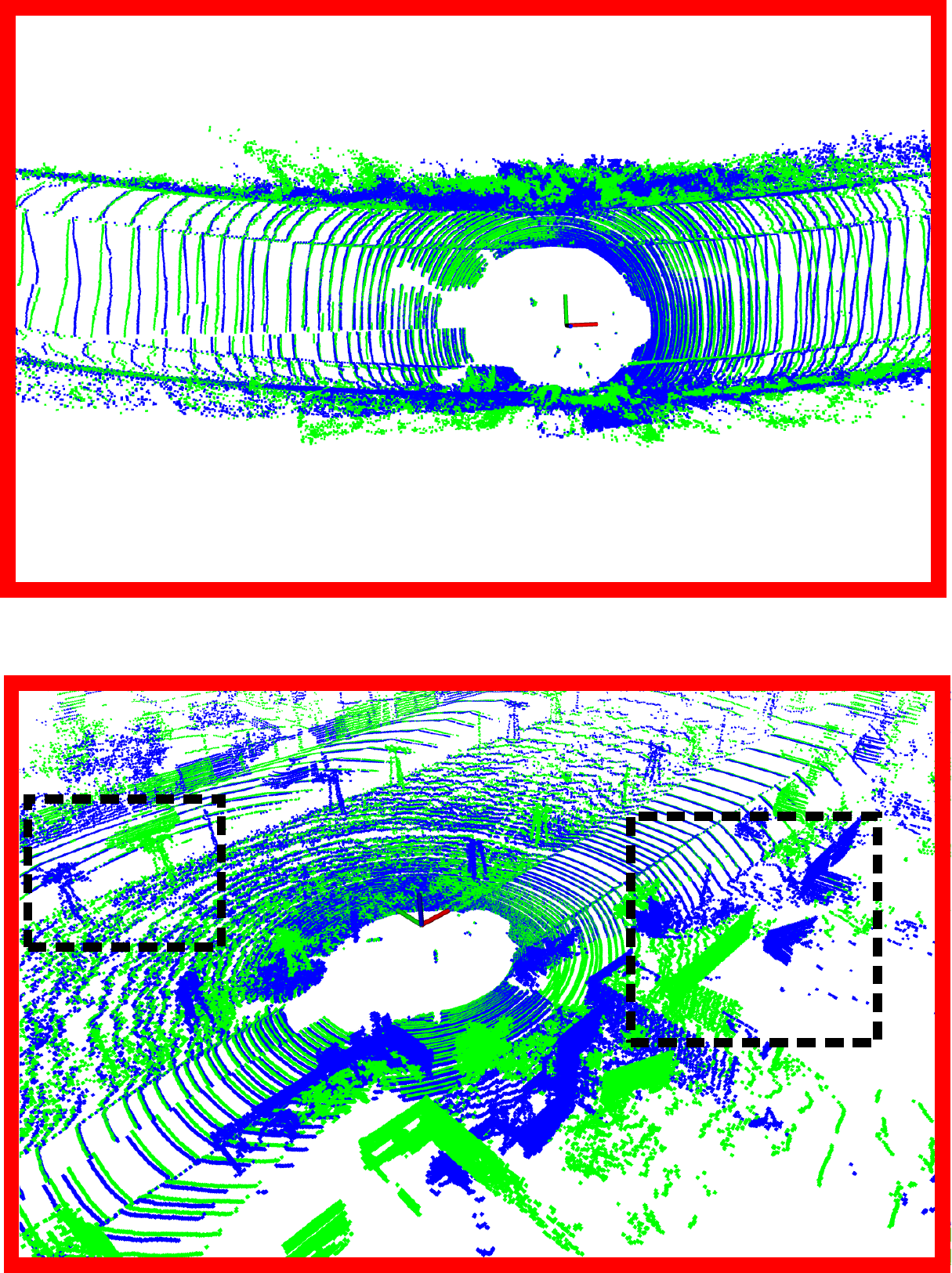}
		\caption{\centering Quatro~(Ours)}
	\end{subfigure}
	\begin{subfigure}[b]{0.30\textwidth}
		\includegraphics[width=1\textwidth]{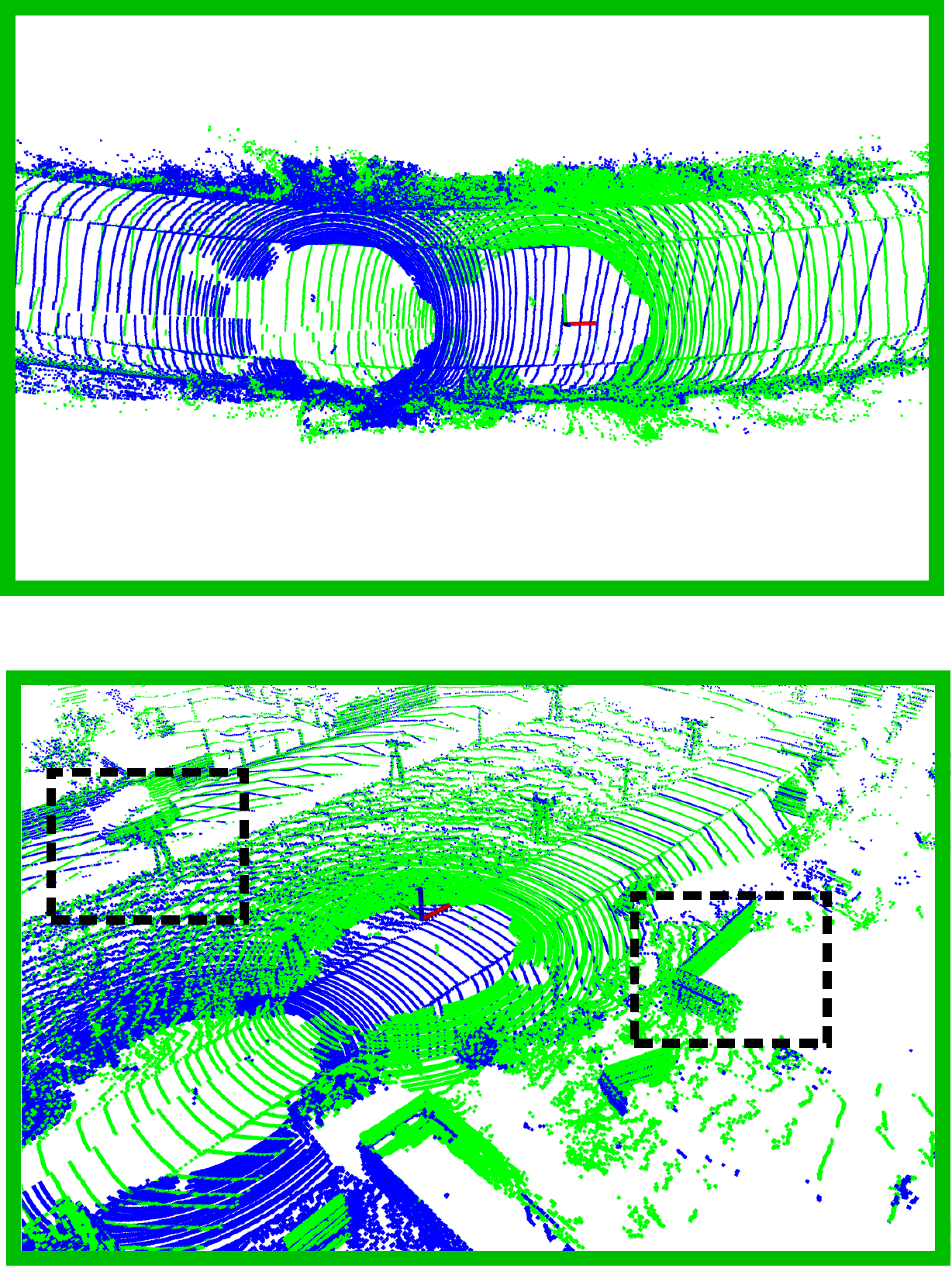}
		\caption{\centering Quatro++~(Ours)}
	\end{subfigure}
	\caption{(a) Source (red) and target clouds (green), where the left-bottom texts represent the pose discrepancy in each scene. (b)-(c) Registration results of our Quatro and Quatro++ between 1,939 to 4,531 frames in Seq.~\texttt{00} and between 165 to 999 frames in Seq.~\texttt{08} in the KITTI dataset, where the blue points denote the warped source cloud by the estimated pose. Quatro++ successfully registers the source and target clouds even though both cloud points are measured in distant and unobstructed cases, where few artificial structures are located. 
	Black dashed boxes highlight the non-ground objects that have to be tightly aligned. The solid red and green boxes indicate the algorithms failed and succeeded, respectively~(best viewed in color)}
	\label{fig:comp_btw_quatro_and_quatropp}
\end{figure*}

\begingroup
\begin{table*}[t!]
	\captionsetup{font=footnotesize}
	\centering
	\caption{\color{black}Comparison of success rates on the KITTI dataset. The bold and the gray-highlight denote the best and the second-best performance, respectively~(unit: \%).}
	\setlength{\tabcolsep}{4pt}
	{\scriptsize
	\color{black}
		\begin{tabular}{l|l|cccccccccc}
			\toprule \midrule
%			\multirow{2}[3]{*}{Method} & \multicolumn{2}{c}{0 $\sim$ 2m} & \multicolumn{2}{c}{4 $\sim$ 6m} & \multicolumn{2}{c}{8 $\sim$ 10m} \\  \cmidrule(lr){2-3} \cmidrule(lr){4-5} \cmidrule(lr){6-7}
%			& $t_{\text{avg}}$ & $r_{\text{avg}}$ & $t_{\text{avg}}$ & $r_{\text{avg}}$ & $t_{\text{avg}}$  & $r_{\text{avg}}$   \\ \midrule
			& Sequence & \multicolumn{3}{c}{\texttt{00}} & \multicolumn{3}{c}{\texttt{02}} & \multicolumn{3}{c}{\texttt{05}}  \\ \cmidrule(lr){3-5} \cmidrule(lr){6-8} \cmidrule(lr){9-11}
			& Diff. of viewpoint & 2 $\sim$ 6\,m & 6 $\sim$ 10\,m & 10 $\sim$ 12\,m & 2\,m $\sim$ 6\,m & 6 $\sim$ 10\,m & 10 $\sim$ 12\,m & 2 $\sim$ 6\,m & 6 $\sim$ 10\,m & 10 $\sim$ 12\,m \\ \midrule
			\parbox[t]{2mm}{\multirow{\numconvention}{*}{\rotatebox[origin=c]{90}{Conv.}}} & RANSAC & 20.3 & 8.5 & 3.3 & 32.3 & 12.0 & 5.8 & 30.3 & 9.3 & 7.7 \\
			& FGR & 99.2 & 79.3 & 45.5 & 71.2 & 41.8 & 21.2 & 99.5 & 83.7 & 48.1 \\
			& TEASER++ & \textbf{100.0} & \hl{94.7} & 84.6 & 95.8 & 78.7 & 71.8 & \hl{99.9} & 99.1 & 93.9 \\
			& Quatro (Ours) & \hl{99.9} & 94.5 & \hl{84.7} & \hl{96.1} & \hl{79.5} & \hl{71.9} & \textbf{100.0} & \hl{99.3} & \hl{94.5} \\
			& Quatro++ (Ours) & \hl{99.9} & \textbf{94.8} & \textbf{87.1} & \textbf{98.7} &  \textbf{97.1} & \textbf{90.8} & \hl{99.9} & \textbf{99.7} & \textbf{98.5} \\ \midrule
			\parbox[t]{2mm}{\multirow{\numdeep}{*}{\rotatebox[origin=c]{90}{Deep}}}
			& MDGAT-Matcher & 94.0 & 87.3 & 69.9 & 12.0 & 0.0 & 0.0 & 94.5 & 89.9 & 81.4  \\
			& LCDNet (fast) & 78.5 & 12.8 & 0.8 & 70.5 & 15.8 & 2.9 & 87.7 & 21.2 & 0.9   \\
			& LCDNet & \textbf{100.0} & 90.3 & 62.1 & 91.4 & 66.4 & 41.3 & \textbf{100.0} & 98.8 & 78.8  \\ \midrule \midrule
			% Here, Segregator: w/o veg
%			& Segregator  & 100.0 & 94.8 & 86.5 & 95.6 & 90.4 & 83.2 & 99.3 & 97.1 & 93.4  \\ \midrule \midrule
			& Sequence & \multicolumn{3}{c}{\texttt{06}} & \multicolumn{3}{c}{\texttt{07}} & \multicolumn{3}{c}{\texttt{08}}  \\ \cmidrule(lr){3-5} \cmidrule(lr){6-8} \cmidrule(lr){9-11}
			& Diff. of viewpoint & 2 $\sim$ 6\,m & 6 $\sim$ 10\,m & 10 $\sim$ 12\,m & 2\,m $\sim$ 6\,m & 6 $\sim$ 10\,m & 10 $\sim$ 12\,m & 2 $\sim$ 6\,m & 6 $\sim$ 10\,m & 10 $\sim$ 12\,m \\ \midrule
			\parbox[t]{2mm}{\multirow{\numconvention}{*}{\rotatebox[origin=c]{90}{Conv.}}} & RANSAC & 19.1 & 8.7 & 3.3 & 17.0 & 5.6 & 5.3 & 18.6 & 6.4 & 2.7 \\
			& FGR       & 99.4     & 92.6 & 59.2 &   99.7     & 81.8 & 41.2 & 36.2 & 24.0 & 15.9 \\
			& TEASER++ & \hl{99.9} & 99.5 & 96.0 & \hl{99.9} & \hl{99.7} & \hl{95.8}                           & {\textbf{100.0}} & \hl{99.7} & 96.6 \\
			& Quatro (Ours) & {\textbf{100.0}} & 99.7 & \hl{97.7}   & {\textbf{100.0}} & \hl{99.7} & \hl{95.8} & {\textbf{100.0}} & {\textbf{99.8}} & \hl{98.0} \\
			& Quatro++ (Ours) & {\textbf{100.0}} & {\textbf{100.0}} & {\textbf{100.0}} & {\textbf{100.0}} & {\textbf{99.9}} & {\textbf{98.0}} & \hl{99.9} & {\textbf{99.8}} & {\textbf{98.9}}  \\ \midrule
			\parbox[t]{2mm}{\multirow{\numdeep}{*}{\rotatebox[origin=c]{90}{Deep}}}
			& MDGAT-Matcher & 99.1 & \hl{99.9} & {91.7} & 92.1 & 94.1 & 68.3 & 17.9 & 55.6 & 7.7  \\
			& LCDNet (fast) & 47.4 & 4.5 & 3.1 & 70.4 & 11.3 & 0.0 & 56.5 & 5.7 & 0.2  \\
			& LCDNet & {\textbf{100.0}} & 99.8 & 85.2 & {\textbf{100.0}} & {98.6} & {80.3} & {\textbf{100.0}} & {95.9} & {75.7}  \\
%			& Segregator   & 100.0 & 99.9 & 99.6 & 99.9 & 99.7 & 98.7 & 98.3 & 97.9 & 94.6  \\
			\midrule\bottomrule
		\end{tabular}
	}
	\label{table:success_rate_in_kitti}
	% 	\vspace{-0.15cm}
\end{table*}
\endgroup

\begingroup
\begin{table*}[t!]
	\captionsetup{font=footnotesize}
	\centering
	\caption{\color{black}Comparison of success rates on the MulRan dataset. The bold and the gray-highlight denote the best and the second-best performance, respectively~(unit: \%).}
	\setlength{\tabcolsep}{4pt}
	{\scriptsize \color{black}
		\begin{tabular}{l|l|cccccccccc}
			\toprule \midrule
%			\multirow{2}[3]{*}{Method} & \multicolumn{2}{c}{0 $\sim$ 2m} & \multicolumn{2}{c}{4 $\sim$ 6m} & \multicolumn{2}{c}{8 $\sim$ 10m} \\  \cmidrule(lr){2-3} \cmidrule(lr){4-5} \cmidrule(lr){6-7}
%			& $t_{\text{avg}}$ & $r_{\text{avg}}$ & $t_{\text{avg}}$ & $r_{\text{avg}}$ & $t_{\text{avg}}$  & $r_{\text{avg}}$   \\ \midrule
			& Sequence & \multicolumn{3}{c}{\texttt{DCC01}} & \multicolumn{3}{c}{\texttt{DCC02}} & \multicolumn{3}{c}{\texttt{DCC03}}  \\ \cmidrule(lr){3-5} \cmidrule(lr){6-8} \cmidrule(lr){9-11}
			& Diff. of viewpoint & 2 $\sim$ 6\,m & 6 $\sim$ 10\,m & 10 $\sim$ 12\,m & 2\,m $\sim$ 6\,m & 6 $\sim$ 10\,m & 10 $\sim$ 12\,m & 2 $\sim$ 6\,m & 6 $\sim$ 10\,m & 10 $\sim$ 12\,m \\ \midrule
			\parbox[t]{2mm}{\multirow{\numconvention}{*}{\rotatebox[origin=c]{90}{Conv.}}} & RANSAC & 74.2 & 46.8 & 32.2 & 78.0 & 48.0 & 34.6 & 76.2 & 50.3 & 30.0 \\
			& FGR             & 63.5 & 43.9 & 41.0 & 63.3 & 50.9 & 44.1 & 65.4 & 45.7 & 35.7 \\
			& TEASER++        & 95.2 & \hl{92.1} & 86.4 & \textbf{99.0} & \hl{93.8} & \hl{88.4} & \textbf{99.4} & \hl{97.8} & \textbf{91.8} \\
			& Quatro (Ours) & \hl{95.3} & \textbf{92.6} & \hl{86.5} & \hl{98.8} & 93.7 & \textbf{90.5} & \hl{99.3} & \textbf{97.9} & \hl{91.3} \\
			& Quatro++ (Ours) & \textbf{95.4} & \textbf{92.6} & \textbf{86.9} & \textbf{99.0} & \textbf{95.0} & \textbf{90.5} & 99.0 & 97.7 & \textbf{91.8}  \\ \midrule
			\parbox[t]{2mm}{\multirow{2}{*}{\rotatebox[origin=c]{90}{Deep}}}
			& LCDNet (fast) & 16.6 & 0.4 & 0.1 & 28.3 & 0.0 & 0.0 & 19.2 & 0.0 & 0.0 \\
			& LCDNet        & 93.8 & 76.1 & 48.0 & 98.6 & 85.2 & 50.0 & 98.6 & 87.0 & 61.7 \\ \midrule \midrule
			% & Segregator    & 93.4 & 82.7 & 68.4 & 98.0 & 85.9 & 75.8 & 98.6 & 91.7 & 80.9  \\ \midrule \midrule
			& Sequence & \multicolumn{3}{c}{\texttt{Riverside01}} & \multicolumn{3}{c}{\texttt{Riverside02}} & \multicolumn{3}{c}{\texttt{Riverside03}}  \\ \cmidrule(lr){3-5} \cmidrule(lr){6-8} \cmidrule(lr){9-11}
			& Diff. of viewpoint & 2 $\sim$ 6\,m & 6 $\sim$ 10\,m & 10 $\sim$ 12\,m & 2\,m $\sim$ 6\,m & 6 $\sim$ 10\,m & 10 $\sim$ 12\,m & 2 $\sim$ 6\,m & 6 $\sim$ 10\,m & 10 $\sim$ 12\,m \\ \midrule
			\parbox[t]{2mm}{\multirow{\numconvention}{*}{\rotatebox[origin=c]{90}{Conv.}}} & RANSAC & 60.2 & 40.7 & 29.1 & 65.2 & 42.6 & 36.0 & 46.7 & 18.9 & 25.0 \\
			& FGR             &     85.9    & 77.3 & 62.6 & \hl{96.9} & \hl{91.0} & \hl{76.9} & \hl{80.9} & 45.5 & 49.5 \\
			& TEASER++        &     85.5    & 78.1 & 73.4 & 92.3 & 83.9 & 73.5 & 80.2 & 48.2 & 65.4 \\
			& Quatro (Ours)   &     85.7    & \hl{79.9} & \hl{78.9} & 93.1 & 86.3 & 76.2 & 80.2 & \hl{49.0} & \hl{68.3} \\
			& Quatro++ (Ours) & \textbf{91.2} & \textbf{87.0} & \textbf{84.1} & \textbf{98.2} & \textbf{93.1} & \textbf{85.8} & \textbf{91.1} & \textbf{70.7} & \textbf{75.8}  \\ \midrule
			\parbox[t]{2mm}{\multirow{2}{*}{\rotatebox[origin=c]{90}{Deep}}}
			& LCDNet (fast) & 27.5 & 0.1 & 0.0 & 65.8 & 10.3 & 0.1 & 20.4 & 2.5 & 0.1 \\
			& LCDNet        & \hl{86.0} & 67.4 & 38.0 & 93.2 & 76.1 & 51.4 & 79.8 & 48.1 & 56.8  \\
%			& Segregator    & 80.8 & 69.2 & 57.6 & 83.9 & 63.7 & 57.2 & 78.9 & 50.7 & 56.0  \\
			\midrule \midrule
		\end{tabular}
		\begin{tabular}{l|l|ccccccc}
%			\midrule \midrule
%			\vspace{-0.1cm}
%			\multirow{2}[3]{*}{Method} & \multicolumn{2}{c}{0 $\sim$ 2m} & \multicolumn{2}{c}{4 $\sim$ 6m} & \multicolumn{2}{c}{8 $\sim$ 10m} \\  \cmidrule(lr){2-3} \cmidrule(lr){4-5} \cmidrule(lr){6-7}
%			& $t_{\text{avg}}$ & $r_{\text{avg}}$ & $t_{\text{avg}}$ & $r_{\text{avg}}$ & $t_{\text{avg}}$  & $r_{\text{avg}}$   \\ \midrule
			& Sequence & \multicolumn{3}{c}{\texttt{KAIST02}} & \multicolumn{3}{c}{\texttt{KAIST03}} \\ \cmidrule(lr){3-5} \cmidrule(lr){6-8}
			& Diff. of viewpoint & 2 $\sim$ 6\,m & 6 $\sim$ 10\,m & 10 $\sim$ 12\,m & 2\,m $\sim$ 6\,m & 6 $\sim$ 10\,m & 10 $\sim$ 12\,m \\ \midrule
			\parbox[t]{2mm}{\multirow{5}{*}{\rotatebox[origin=c]{90}{Conv.}}} & RANSAC & 74.8 & 60.8 & 37.8 & 87.3 & 69.9 & 44.3 \\
			& FGR & 73.8 & 72.3 & 68.8 & \hl{99.9} & 98.8 & 92.2 \\
			& TEASER++ & \textbf{94.7} & \textbf{93.0} & \textbf{88.5} & \hl{99.9} & \hl{98.9} & 93.3 \\
			& Quatro (Ours) & 94.0 & \hl{92.5} & \hl{87.8} & \hl{99.9} & \hl{98.9} & \hl{94.3} \\
			& Quatro++ (Ours) & 93.5 & 91.3 & \textbf{88.5} & \textbf{100.0} & \textbf{99.6} & \textbf{98.0}  \\ \midrule
			\parbox[t]{2mm}{\multirow{2}{*}{\rotatebox[origin=c]{90}{Deep}}}
			& LCDNet (fast) & 16.3 & 0.7 & 0.4 & 20.1 & 2.2 & 0.0 \\
			& LCDNet        & \hl{94.4} & 86.2 & 71.9 & \hl{99.9} & 93.6 & 73.7 \\  \midrule \bottomrule
			% & Segregator    & 94.5 & 89.3 & 84.8 & 99.8 & 98.9 & 94.6 \\ \midrule\bottomrule
		\end{tabular}
	}
	\label{table:success_rate_in_mulran}
\end{table*}
\endgroup

In our experiments, other well-known global registration methods are employed as follows: RANSAC~\citep{fischler1981ransac} whose maximum iteration is set to 10,000, FGR~\citep{zhou2016fast}, and TEASER++~\citep{yang2020teaser}. 
Here, the main differences between our proposed method and the baseline methods are explained. 
First, RANSAC and FGR have no outlier rejection modules like MCIS. 
FGR and TEASER++ are also GNC-based methods, so alternating optimization is performed to reject the outlier pairs. 
The biggest difference between our proposed method and each baseline method is as follows: 
FGR uses linearization in SE(3) manifold and the pose difference is estimated by gradient-based optimization;
Similar to our proposed method, TEASER++ also utilizes the decoupling method.
However, TEASER++ estimates SO(3) estimation, unlike our quasi-SO(3) estimation, \color{black} and does not exploit ground segmentation.
To avoid confusion, we stress that the meanings of ``++'' in Quatro++ and TEASER++ are different:
TEASER++ is a speeded up version of TEASER at an optimization level and Quatro++ denotes the combination of Quatro with ground segmentation. \color{black}

\color{black}
Furthermore, we compare our Quatro++ with deep learning-based methods: MDGAT-Matcher~\citep{shi2021keypoint} and LCDNet~\citep{cattaneo2022lcdnet}.
We evaluate two versions of LCDNet: one using an unbalanced optimal transport algorithm, which estimates the transformation matrix in a learning-based manner, denoted as LCDNet~(fast),
and the other one uses RANSAC with 50,000 iterations, denoted as LCDNet.

In particular, we compare the generalization capabilities of our approach and the deep learning-based methods, so the networks trained on the KITTI dataset are tested on the MulRan dataset.
Note that both datasets were acquired by 64-channel LiDAR sensors, but the KITTI dataset was acquired by the Velodyne HDL-64E sensor, while the MulRan dataset was acquired by the OS1-64 sensor.
Therefore, we can check the generalization capabilities in terms of both environmental changes and different sensor configuration.

When evaluating the coarse-to-fine alignment performance, we use the term Quatro++-\texttt{c2f}, which is comprised of the proposed Quatro++ as a coarse alignment and G-ICP~\citep{segal2009gicp} as a fine alignment.

\color{black}

%% Quatro++라고 하고 가정하자!
\subsection{Performance Comparison Between Quatro and Quatro++}\label{exp:quat_and_quatpp}

First, the performance of Quatro and Quatro++ was analyzed. Note that the difference between Quatro++ and Quatro is with and without the ground segmentation module before the feature extraction and matching steps. 
As shown in Fig.~\ref{fig:comp_btw_quatro_and_quatropp}, Quatro occasionally fails to perform registration once the pose discrepancy between the source and target clouds is large, so the correspondences from the central ground points are dominant~(Fig.~\ref{fig:comp_btw_quatro_and_quatropp}(b)).
% This is because the sparsity becomes more severe as the pose difference between two viewpoints is more distant. 
In contrast, Quatro++ showed more robust performance under these situations, successfully warping the source cloud into the target cloud~(Fig.~\ref{fig:comp_btw_quatro_and_quatropp}(c)). 

Consequently, as shown in \color{black} Tables~\ref{table:success_rate_in_kitti} and ~\ref{table:success_rate_in_mulran}, \color{black} Quatro++ presented a higher success rate, which is the most important evaluation metric to check the suitability of global registration methods as an initial alignment.
In particular, \color{black}we demonstrate that Quatro++ becomes more robust in distant cases, whose difference in viewpoints between the source and target clouds ranges between 10 to 12\,m.
Furthermore, Quatro++ showed \color{black} a remarkable improvement when our proposed method is used in Seq.~\texttt{02}, which is a rural environment that contains few non-ground objects and the ground points are more dominant.

%\begin{figure}[t!]
%	\captionsetup{font=footnotesize}
%	\centering
%	\begin{subfigure}[b]{0.45\textwidth}
%		\includegraphics[width=1\textwidth]{imgs/SuccessRate1.png}
%		\includegraphics[width=1\textwidth]{imgs/SuccessRate2.png}
%	\end{subfigure}
%	\caption{Success rate of the global registration methods in the loop closing situations of the KITTI dataset whose position differences are between 10$\sim$12 m away. The criteria of the success rate is based on~\cite{kim2019gp}: a
%    pose error that is sufficiently within the narrow convergence region where the estimate of local registration successfully converges into the global optimum. Thus, the registration was counted as a success if both relative translation and rotation errors are lower than 2~m and 10$^\circ$, respectively.}
%	\label{fig:success_rate}
%\end{figure}

\subsection{Effect of Ground Segmentation on Global Registration}\label{exp:effect_of_ground_segmentation}

In addition, we conducted \color{black}three in-depth analyses \color{black} of the effect of the ground segmentation on the global registration \color{black} as follows:
impact of ground segmentation a)~when two point clouds with large discrepancy of viewpoints are given and b)~when more spurious correspondences are given, and c)~influence of the performance of ground segmentation on global registration. \color{black}

\begin{figure}[t!]
	\captionsetup{font=footnotesize}
	\centering
    \begin{subfigure}[b]{0.5\textwidth}
		\centering
		\includegraphics[width=0.92\textwidth]{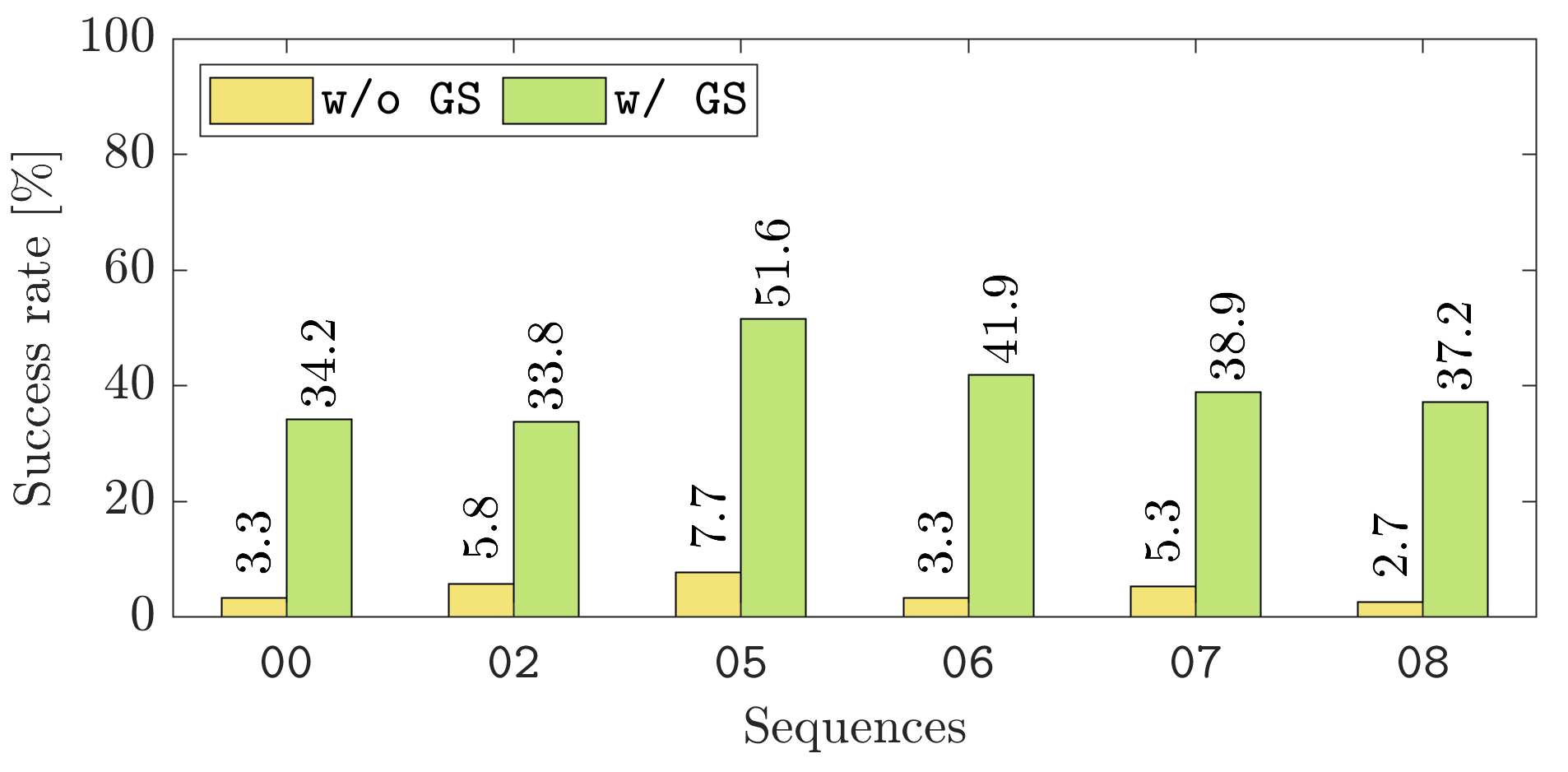}
		\caption{\centering RANSAC}
	\end{subfigure}
	\begin{subfigure}[b]{0.5\textwidth}
		\centering
		\includegraphics[width=0.92\textwidth]{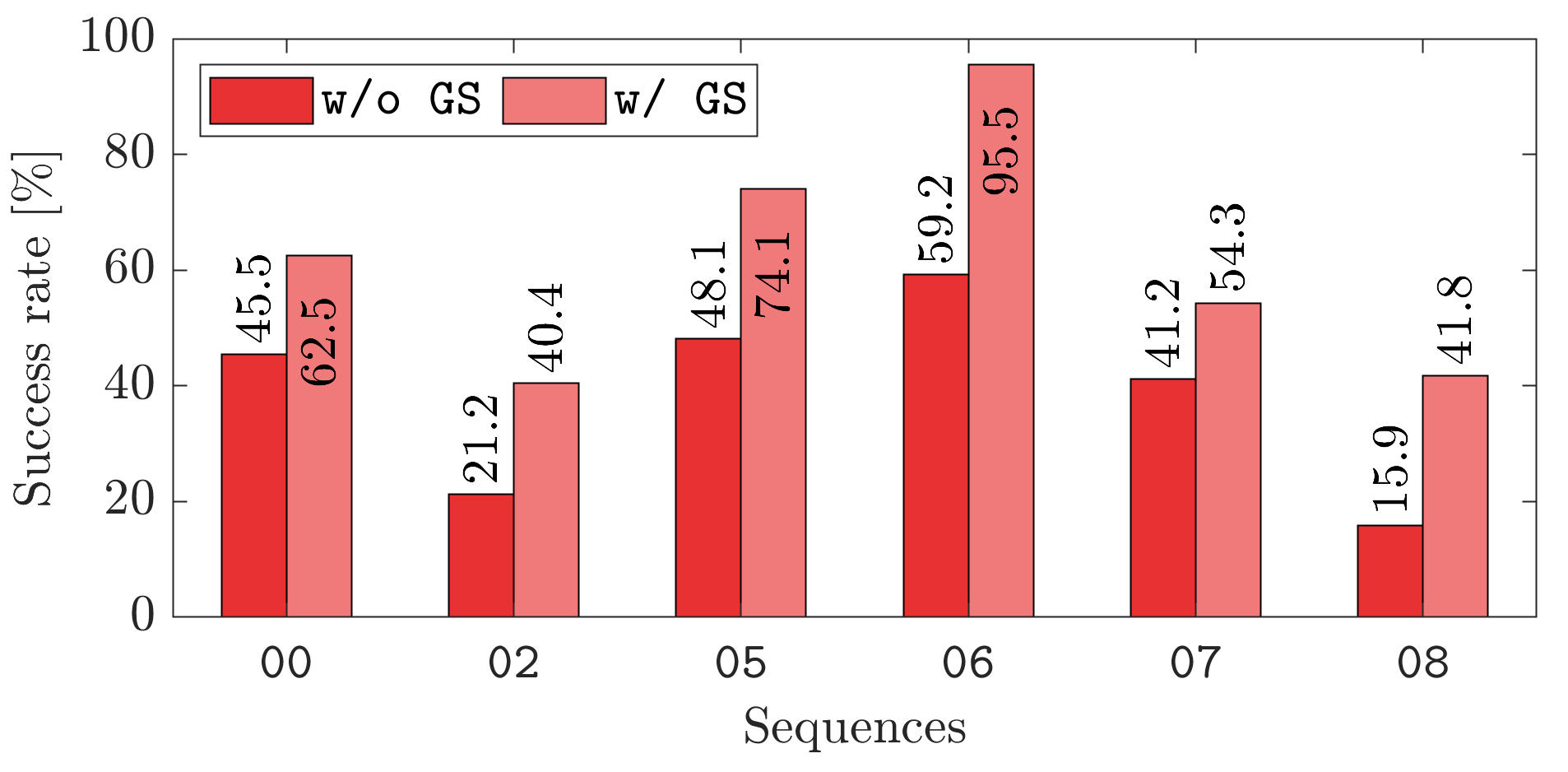}
		\caption{\centering FGR}
	\end{subfigure}
	\begin{subfigure}[b]{0.5\textwidth}
		\centering
		\includegraphics[width=0.92\textwidth]{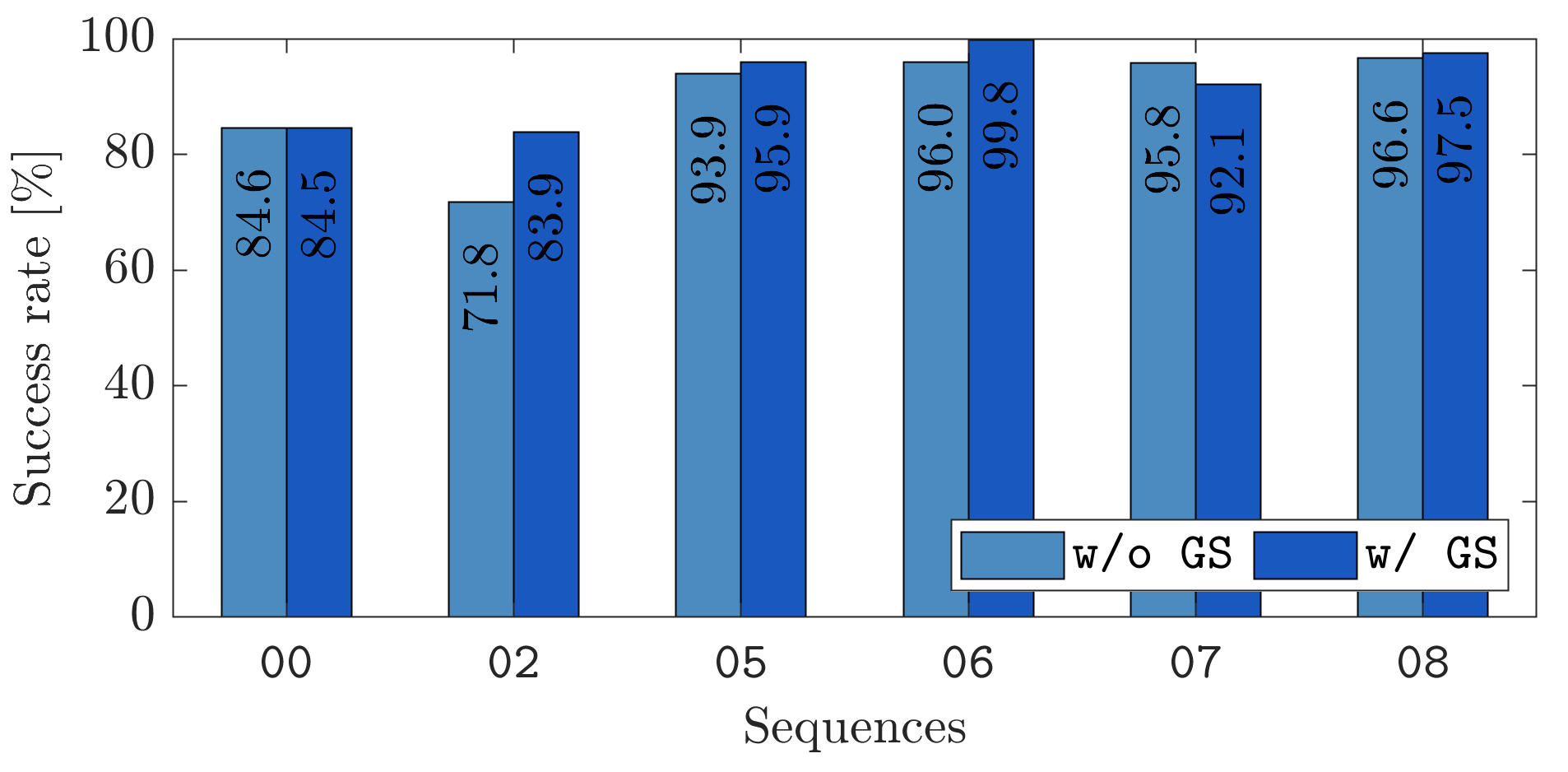}
		\caption{\centering TEASER++}
	\end{subfigure}
	\caption{Average success rate changes of other global registration methods depending on the absence~(\texttt{w/o}) and presence~(\texttt{w/}) of ground segmentation~(\texttt{GS}) module in the loop closing situations of the KITTI dataset where the position discrepancy is between 10$\sim$12 m.}
	\label{fig:success_rate_each}
\end{figure}

First, as shown in Fig.~\ref{fig:success_rate_each} \color{black} and Table~\ref{table:success_rate_mulran_w_wo_gseg}, ground segmentation substantially \color{black} increases the performance of all global registration methods \color{black} when the viewpoint discrepancy between source and target is large. \color{black}
The main reason for the poor performance without the ground segmentation is because of the too many correspondences from the central ground points of the source to those of the target. 
That is, both ground points of the source and target close to the origin are sufficiently dense regardless of the large pose discrepancy. 
Accordingly, even though these central points are not measured from the same viewpoint, the high density makes descriptors from the central ground points similar, leading to gross outlier correspondences.
Consequently, these false correspondences finally lead to wrong pose estimation~(empirically, global registration methods estimate the relative translation as a near-zero vector in this case as shown in Fig.~\ref{fig:comp_btw_quatro_and_quatropp}(b)).
In summary, the high density of the central points is misunderstood as a geometrical characteristic itself.

\color{black}
While the ground segmentation module occasionally decreases the performance slightly, Quatro with ground segmentation showed better performance when the viewpoint discrepancy between source and target is large~(note that the combination of Quatro and ground segmentation is equal to Quatro++).
Because Quatro estimates the relative rotation by using the quasi-SO(3) estimation, its reduced DoF allows us to perform robust global registration even though few correspondences are given.
That means Quatro is more robust to the reduced number of correspondences due to the rejection of many ground points than TEASER++.
In contrast, TEASER++, whose DoF for estimation of rotation is three, sometimes showed worse performance because rejection of ground points substantially reduces the number of correspondences, resulting in degeneracy.
Finally, the degeneracy potentially leads to large rotation errors of TEASER++.
Therefore, these results demonstrate the potential of synergy between our Quatro and ground segmentation.

% While the ground segmentation module slightly decreases the performance in the 2 $\sim$ 6\,m case in Table~\ref{table:success_rate_mulran_w_wo_gseg} because the these central points are likely to be true correspondences when the pose discrepancy is small, so
% yet the performance degradation is negligible.

Second, we validate that ground segmentation especially improves the performance of global registration even when more spurious correspondences are given, as shown in Table~\ref{table:success_rate_mulran_w_wo_gseg}.
We deliberately set the parameters of FPFH as $\nu= 0.3$\,m, $r_\text{normal} = 0.5$\,m, and $r_\text{FPFH}=0.65$\,m in the MulRan dataset~(note that the appropriate parameters for MulRan dataset are $\nu= 0.6$\,m, $r_\text{normal} = 1.5$\,m, and $r_\text{FPFH}=2.25$\,m, as presented in Table~\ref{table:quatro_param}).
Once more spurious correspondences are given, the success rates of global registration methods without ground segmentation are more dramatically decreased, as shown in Table~\ref{table:success_rate_mulran_w_wo_gseg}.
However, after the application of ground segmentation, global registration approaches showed better success rates, which supports that ground segmentation helps to increase the coverage of the global registration methods.
In particular, ground segmentation brought much greater performance increase when the spurious correspondences were given than when accurate ones were given, which also corroborates that ground segmentation effectively prevents the catastrophic failure caused by imprecise feature matching.

% To be updated
Finally, we analyzed the results of the global registration according to the performance of the ground segmentation.
To this end, we employed GPF~\citep{zermas2017fast}, LineFit~\citep{himmelsbach2010fast}, CascadedSeg~\citep{narksri2018slope}, R-GPF~\citep{lim21erasor}, and Patchwork~\citep{lim2021patchwork}.
As reported in \cite{lim2021patchwork} and \cite{lim2022pago}, while GPF, LineFit, and CascadedSeg showed high precision with low recall, R-GPF showed high recall with low precision.
In contrast, Patchwork showed both high precision and recall with little perturbation of ground segmentation performance, showing ths highest F$_1$-score.

\begin{figure}[t!]
	\captionsetup{font=footnotesize}
	\centering
%	\begin{subfigure}[b]{0.32\textwidth}
%		\includegraphics[width=1.0\textwidth]{imgs/gs_precision-eps-converted-to.pdf}
%		\caption{\centering}
%	\end{subfigure}
%	\begin{subfigure}[b]{0.32\textwidth}
%		\includegraphics[width=1.0\textwidth]{imgs/gs_recall-eps-converted-to.pdf}
%		\caption{\centering}
%	\end{subfigure}
	\begin{subfigure}[b]{0.47\textwidth}
		\includegraphics[width=1.0\textwidth]{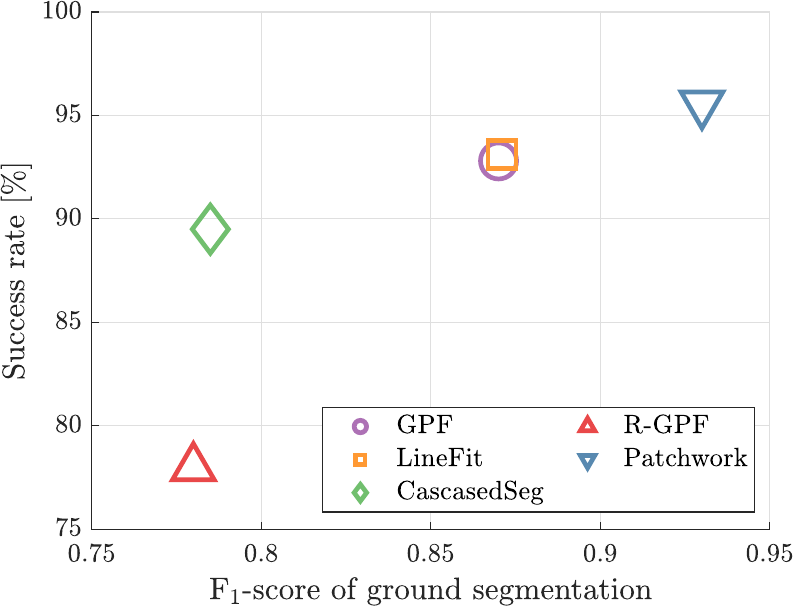}
	\end{subfigure}
	\caption{\color{black}Influence of the performance of ground segmentation on the success rate of our Quatro. Note that Quatro with Patchwork is equal to Quatro++~(best viewed in color). \color{black}}
	\label{fig:gseg_on_quatro}
\end{figure}

As shown in Fig.~\ref{fig:gseg_on_quatro}, the performance of ground segmentation and global registration exhibits a logarithmic relationship.
This is because higher F$_1$-score implies that ground points are precisely and repeatably rejected.
By successfully rejecting the ground points from point clouds observed at other viewpoints, ground segmentation with high F$_1$-score allows global registration to take filtered points whose most parts are consistently overlapped as input.
In contrast, if we employ ground segmentation with low F$_1$-score, ground points are inconsistently rejected.
By doing so, even though the ground points are rejected from the source or target cloud, the ground may not be clearly rejected from the other point cloud.
This phenomenon rather increases the number of outliers. For this reason, our approach with other ground segmentation approaches showed lower performance than one with Patchwork.

Therefore, \color{black} we conclude \color{black} that exploiting ground segmentation enables global registration to robustly estimate the relative pose between the two point clouds whose pose discrepancy is large.
\color{black} In particular, the experimental evidence supports that ground segmentation with high F$_1$-score effectively prevents failure cases by rejecting the gross outliers from the ground points in advance.
\color{black}

\begingroup
\begin{table*}[t!]
	\captionsetup{font=footnotesize}
	\centering
	\caption{\color{black}Comparison of success rate changes before and after the application of ground segmentation~\citep{lim2021patchwork} on MulRan dataset when more spurious correspondences are given by setting the parameters of FPFH as $\nu= 0.3$\,m, $r_\text{normal} = 0.5$\,m, and $r_\text{FPFH}=0.65$\,m~(unit: \%).}
	\setlength{\tabcolsep}{2pt}
	{\scriptsize \color{black}
		\begin{tabular}{l|l|c|cccccccccc}
			\toprule \midrule
%			\multirow{2}[3]{*}{Method} & \multicolumn{2}{c}{0 $\sim$ 2m} & \multicolumn{2}{c}{4 $\sim$ 6m} & \multicolumn{2}{c}{8 $\sim$ 10m} \\  \cmidrule(lr){2-3} \cmidrule(lr){4-5} \cmidrule(lr){6-7}
%			& $t_{\text{avg}}$ & $r_{\text{avg}}$ & $t_{\text{avg}}$ & $r_{\text{avg}}$ & $t_{\text{avg}}$  & $r_{\text{avg}}$   \\ \midrule
			& \multirow{2}{*}{Method} & \multirow{2}{*}{\begin{tabular}{@{}c@{}}Ground \\ seg.\end{tabular}} & \multicolumn{3}{c}{\texttt{DCC}} & \multicolumn{3}{c}{\texttt{Riverside}} & \multicolumn{3}{c}{\texttt{KAIST}}  \\ \cmidrule(lr){4-6} \cmidrule(lr){7-9} \cmidrule(lr){10-12}
			&  &  & 2 $\sim$ 6\,m & 6 $\sim$ 10\,m & 10 $\sim$ 12\,m & 2\,m $\sim$ 6\,m & 6 $\sim$ 10\,m & 10 $\sim$ 12\,m & 2 $\sim$ 6\,m & 6 $\sim$ 10\,m & 10 $\sim$ 12\,m \\ \midrule
			\parbox[t]{9mm}{\multirow{8}{*}{\rotatebox[origin=c]{90}{\begin{tabular}{@{}c@{}}\!Params: $0.3 / 0.5 / 0.65$\,m\; \\ \!(more imprecise \\ correspondences are given)\; \end{tabular}}}} & \multirow{2}{*}{RANSAC}            &            & 22.3 & 10.2 & 7.6 & 15.6 & 4.8 & 4.0 & 25.9 & 10.9 & 7.7 \\
			&& \checkmark & \textbf{44.5} (22.2\,\textuparrow) & \textbf{27.8} (17.6\,\textuparrow) & \textbf{24.9} (17.3\,\textuparrow) & \textbf{39.3} (23.7\,\textuparrow) & \textbf{24.0} (19.2\,\textuparrow) & \textbf{17.2} (13.2\,\textuparrow) & \textbf{53.7} (27.8\,\textuparrow) & \textbf{36.9} (26.0\,\textuparrow) & \textbf{23.2} (15.5\,\textuparrow) \\ \cmidrule{2-12}
			&\multirow{2}{*}{FGR}               &            & 50.8 & 16.4 & 4.7 & 56.9 & 9.1 & 1.6 & 67.1 & 33.0 & 12.6 \\
			&& \checkmark & \textbf{56.5} (5.7\,\textuparrow) & \textbf{37.5} (21.1\,\textuparrow) & \textbf{24.9} (20.2\,\textuparrow) & \textbf{88.6} (31.7\,\textuparrow)  & \textbf{56.7} (47.6\,\textuparrow) & \textbf{28.7} (27.1\,\textuparrow) & \textbf{81.2} (14.1\,\textuparrow) & \textbf{76.6} (43.6\,\textuparrow) & \textbf{57.1} (44.5\,\textuparrow) \\ \cmidrule{2-12}
			&\multirow{2}{*}{TEASER++}          &            & \textbf{95.6} & 72.8 & 44.0  & 82.5 & 46.5 & 22.4  & 89.8  & 63.4  & 39.4 \\
			&& \checkmark & 95.4 (0.2\,\textdownarrow) & \textbf{78.0} (5.2\,\textuparrow) & \textbf{52.9} (8.9\,\textuparrow) & \textbf{89.8} (7.3\,\textuparrow) & \textbf{66.4} (19.9\,\textuparrow) & \textbf{41.3} (18.9\,\textuparrow) & \textbf{96.8} (7.0\,\textuparrow) & \textbf{86.8} (23.4\,\textuparrow) & \textbf{63.0} (23.6\,\textuparrow) \\ \cmidrule{2-12}
			&\multirow{2}{*}{Quatro (Ours)}     &            & 95.9 & 75.0 & 48.0 & 82.9 & 48.1 & 24.2 & 90.1 & 64.4 & 41.7 \\
			&& \checkmark & \textbf{96.4} (0.5\,\textuparrow) & \textbf{83.2} (8.2\,\textuparrow) & \textbf{61.1} (13.1\,\textuparrow) & \textbf{91.6} (8.7\,\textuparrow) & \textbf{74.5} (26.4\,\textuparrow) & \textbf{49.8} (25.6\,\textuparrow)  & \textbf{97.0} (6.9\,\textuparrow)  & \textbf{90.8} (26.4\,\textuparrow) & \textbf{71.3} (29.6\,\textuparrow) \\ \midrule \midrule
			%%%%%%%%%%%%%%%%%%%%%%%%% Better parameters
			\parbox[t]{9mm}{\multirow{8}{*}{\rotatebox[origin=c]{90}{\begin{tabular}{@{}c@{}}\!Params: $0.6 / 1.5 / 2.25$\,m\;\; \\ \!(less imprecise \\ correspondences are given)\;\; \end{tabular}}}} & \multirow{2}{*}{RANSAC}            &            & 76.1 & 48.4 & 32.3 & 67.2 & 44.5 & 31.7 & 81.1 & 65.4 & 41.1 \\
			&& \checkmark & \textbf{83.6} (7.5\,\textuparrow) & \textbf{55.4} (7.0\,\textuparrow) & \textbf{35.3} (3.0\,\textuparrow) & \textbf{81.8} (14.6\,\textuparrow) & \textbf{61.9} (17.4\,\textuparrow) & \textbf{45.1} (13.4\,\textuparrow) & \textbf{89.7} (8.6\,\textuparrow) & \textbf{74.5} (9.1\,\textuparrow) & \textbf{56.0} (14.9\,\textuparrow) \\ \cmidrule{2-12}
			&\multirow{2}{*}{FGR}               &            & 64.1 & 46.8 & 40.3 & 87.9 & 71.3 & 63.0 & 86.9 & 85.6 & 80.5 \\
			&& \checkmark & \textbf{68.8} (4.7\,\textuparrow) & \textbf{51.1} (4.3\,\textuparrow) & \textbf{44.9} (4.6\,\textuparrow) & \textbf{94.4} (6.5\,\textuparrow)  & \textbf{84.2} (12.9\,\textuparrow) & \textbf{81.0} (18.0\,\textuparrow) & \textbf{87.8} (0.9\,\textuparrow) & \textbf{86.0} (0.4\,\textuparrow) & \textbf{85.7} (5.2\,\textuparrow) \\ \cmidrule{2-12}
			&\multirow{2}{*}{TEASER++}          &            & \textbf{97.9} & \textbf{94.6} & 88.3  & 92.4 & 86.6 & 79.6  & \textbf{97.3}  & \textbf{96.0}  & 90.9 \\
			&& \checkmark &		                             97.7 (0.2\,\textdownarrow) & 94.4 (0.2\,\textdownarrow) & \textbf{86.4} (1.9\,\textdownarrow) & \textbf{95.1} (2.7\,\textuparrow) & \textbf{87.1} (0.5\,\textuparrow) & \textbf{80.2} (0.6\,\textuparrow) & 97.2 (0.1\,\textdownarrow) & 94.2 (1.8\,\textdownarrow)  & \textbf{91.0} (0.1\,\textuparrow)  \\ \cmidrule{2-12}
			&\multirow{2}{*}{Quatro (Ours)}     &            & \textbf{97.8} & 94.7 & 89.4 & 92.7 & 88.0 & 82.1 & \textbf{97.0} & \textbf{95.7} & 91.1 \\
			&& \checkmark & \textbf{97.8} (0.0\,$-$) & \textbf{95.1} (0.4\,\textuparrow) & \textbf{89.7} (0.3\,\textuparrow) & \textbf{96.1} (3.4\,\textuparrow) & \textbf{92.6} (4.6\,\textuparrow) & \textbf{87.2} (5.1\,\textuparrow)  & 96.8 (0.2\,\textdownarrow)  & 95.5 (0.3\,\textdownarrow) & \textbf{93.3} (2.2\,\textuparrow) \\ \midrule \bottomrule
		\end{tabular}
	}
	\label{table:success_rate_mulran_w_wo_gseg}
	% 	\vspace{-0.15cm}
\end{table*}
\endgroup

\subsection{Performance Comparison With State-of-the-Art Methods}\label{sec:exp_sota}

\noindent \textbf{In Loop Closing Situations} \; Next, the performance of global registration methods is compared in loop closing situations. 
As shown in \color{black} Tables~\ref{table:success_rate_in_kitti} and \ref{table:success_rate_in_mulran}, all the baseline methods showed \color{black} high success rates \color{black} once the relative pose between two viewpoints of source and target was sufficiently close \color{black} (2 $\sim$ 6\,m case). \color{black}
However, as the relative pose between two viewpoints of source and target became distant, our Quatro++ showed higher success rate than other global registration methods \color{black} (10 $\sim$ 12\,m case). \color{black}
Again, we place more emphasis on the success rate that can directly check whether the global registration methods can successfully estimate the relative pose as an initial alignment even though distant loop pairs are given. 
Thus, the higher the success rate is, the more suitable method is for performing loop closing. 

In addition, it is also noticeable that our preliminary version, Quatro, was on par with TEASER++ \color{black} (Tables~\ref{table:success_rate_in_kitti} and~\ref{table:success_rate_in_mulran}) \color{black} or showed higher success rate \color{black} when more imprecise correspondences are given (Table~\ref{table:success_rate_mulran_w_wo_gseg}). \color{black}
This is because quasi-SO(3) estimation, which reduces the minimum of DoF from three to one, enables to be robust against degeneracy. By doing so, our proposed method prevents catastrophic failure when estimating rotation, unlike TEASER++. For these reasons, Quatro showed higher success rate and smaller errors in the distant cases~(8$\sim$ 10 m case, Table~\ref{table:kitti_IMU}).

\begingroup
\begin{table}[t!]
	\captionsetup{font=footnotesize}
	\centering
	\caption{Comparison of average translation and rotation errors, i.e.~$t_\text{avg}$ and $r_\text{avg}$, with the state-of-the-art methods in loop closing situations of Seq.~\texttt{06} in the KITTI dataset. The ranges in the first row represent position discrepancy boundaries of loop pairs. \textcolor{qwr}{The bold and the \textcolor{qwe}{gray-highlight} denote the best and the second-best performance, respectively} (units for $t_\text{avg}$:~m, $r_\text{avg}$:~$\deg$).}  
	\setlength{\tabcolsep}{4pt}
	{\scriptsize
		\begin{tabular}{l|cccccc}
			\toprule \midrule
			\multirow{2}[3]{*}{Method} & \multicolumn{2}{c}{0 $\sim$ 2\,m} & \multicolumn{2}{c}{4 $\sim$ 6\,m} & \multicolumn{2}{c}{8 $\sim$ 10\,m} \\  \cmidrule(lr){2-3} \cmidrule(lr){4-5} \cmidrule(lr){6-7}
			& $t_{\text{avg}}$ & $r_{\text{avg}}$ & $t_{\text{avg}}$ & $r_{\text{avg}}$ & $t_{\text{avg}}$  & $r_{\text{avg}}$   \\ \midrule
			RANSAC & 2.369 & 14.22 & 5.010 & 27.35 & 8.341 & 33.34  \\
			FGR & \textbf{0.057} & 0.222 & 0.103 & 0.301 & 1.821 & 1.828  \\
			TEASER++ & 0.070 & 0.285 & 0.131 & 0.481 & 0.498 & 1.469  \\
			Quatro (Ours) & 0.067 & 0.324 & 0.120 & {0.465} & 0.471 & 0.724   \\ 
			Quatro++ (Ours) & 0.078 & 0.338 & 0.110 & 0.452 & \hl{0.163} & 0.561   \\ \midrule
			Quatro w/ INS (Ours) & \hl{0.059} & \textbf{0.207} & \hl{0.101} & \hl{0.230} & 0.429 & \hl{0.346}   \\
			Quatro++ w/ INS (Ours) & 0.067 & \hl{0.208} & \textbf{0.089} & \textbf{0.222} & \textbf{0.117} & \textbf{0.243}   \\
			\midrule\bottomrule
		\end{tabular}
	}
	\label{table:kitti_IMU}
	% 	\vspace{-0.15cm}
\end{table}
\endgroup

\begin{figure*}[t!]
	\captionsetup{font=footnotesize}
	\centering
	\begin{subfigure}[b]{0.19\textwidth}
		\includegraphics[width=1.0\textwidth]{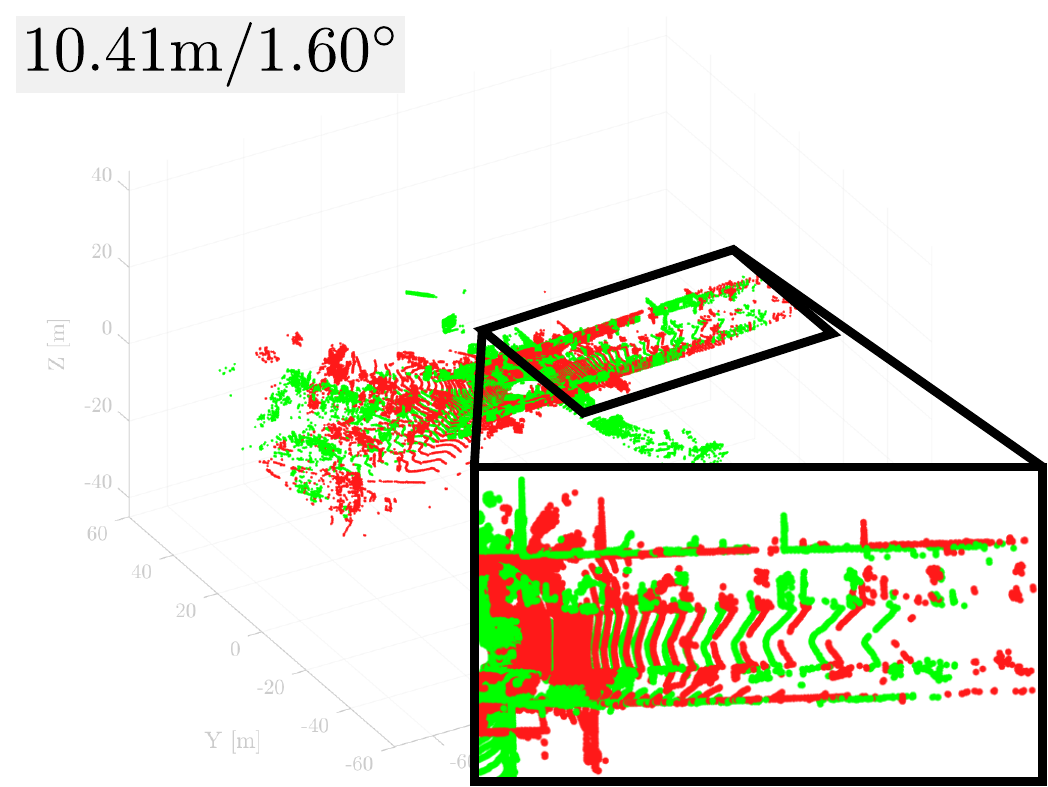}
		\includegraphics[width=1.0\textwidth]{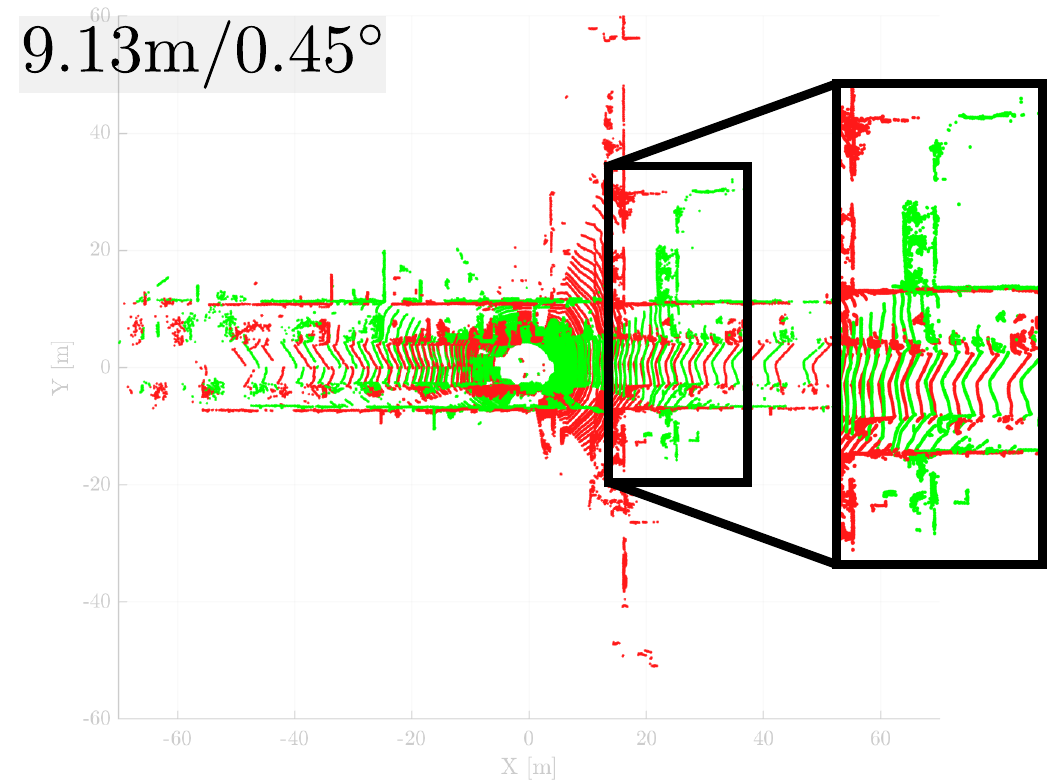}
		\includegraphics[width=1.0\textwidth]{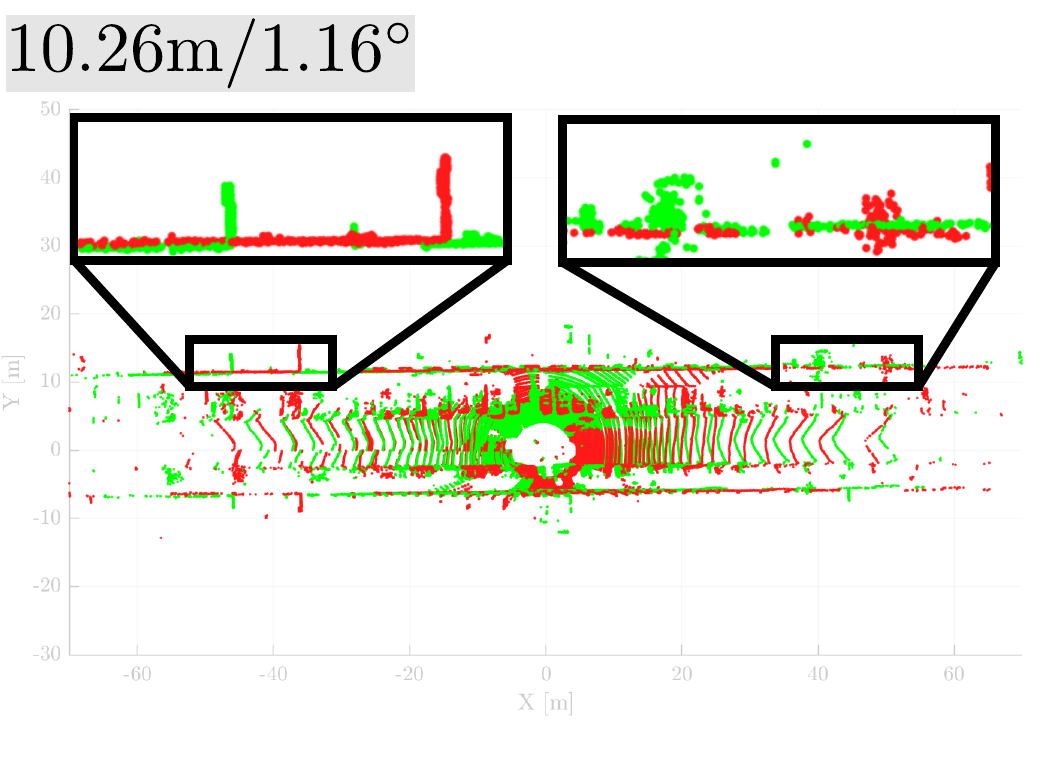}
		\includegraphics[width=1.0\textwidth]{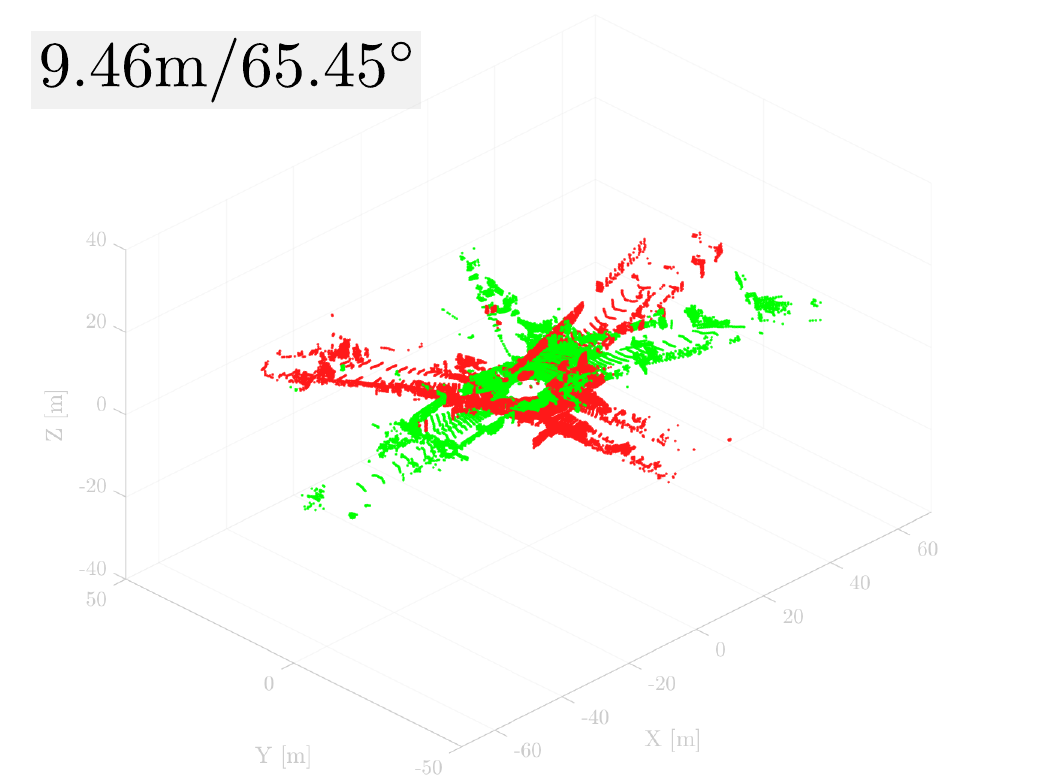}
		\includegraphics[width=1.0\textwidth]{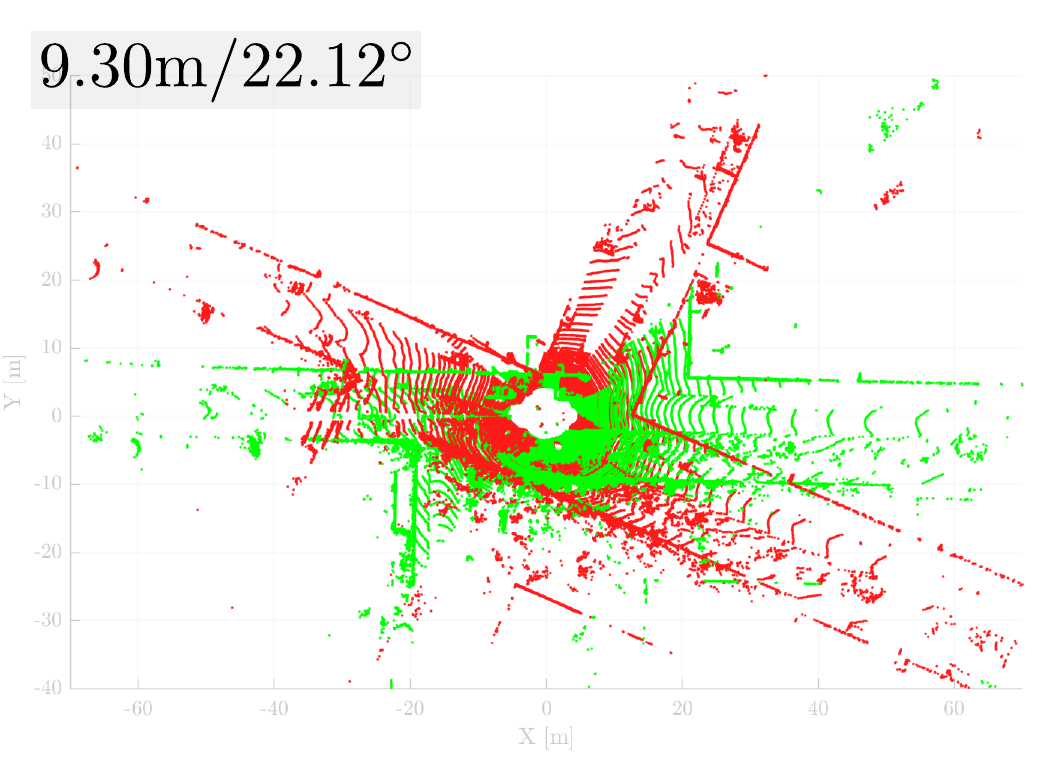}
		\includegraphics[width=1.0\textwidth]{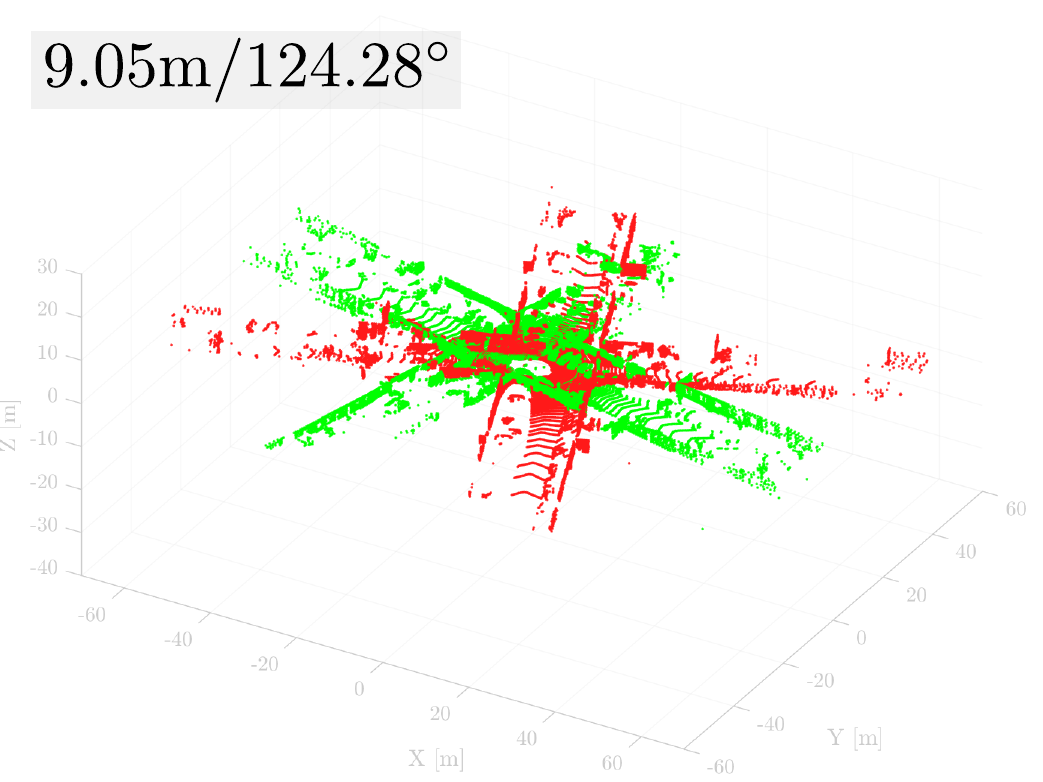}
		\includegraphics[width=1.0\textwidth]{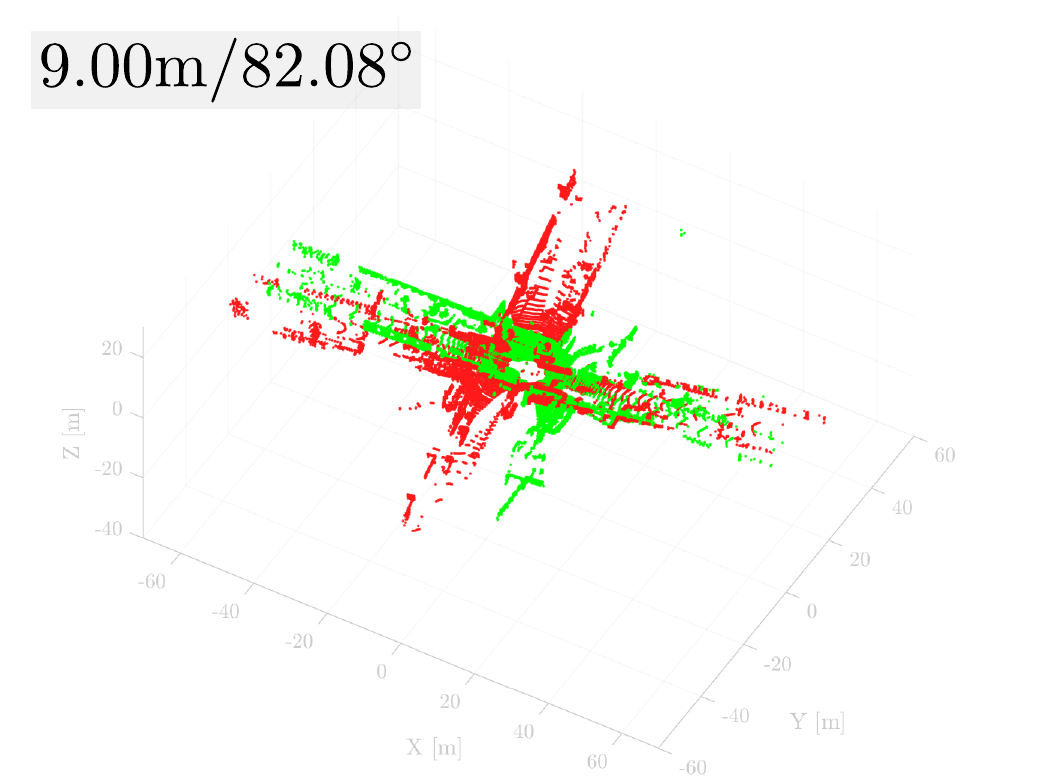}
		\includegraphics[width=1.0\textwidth]{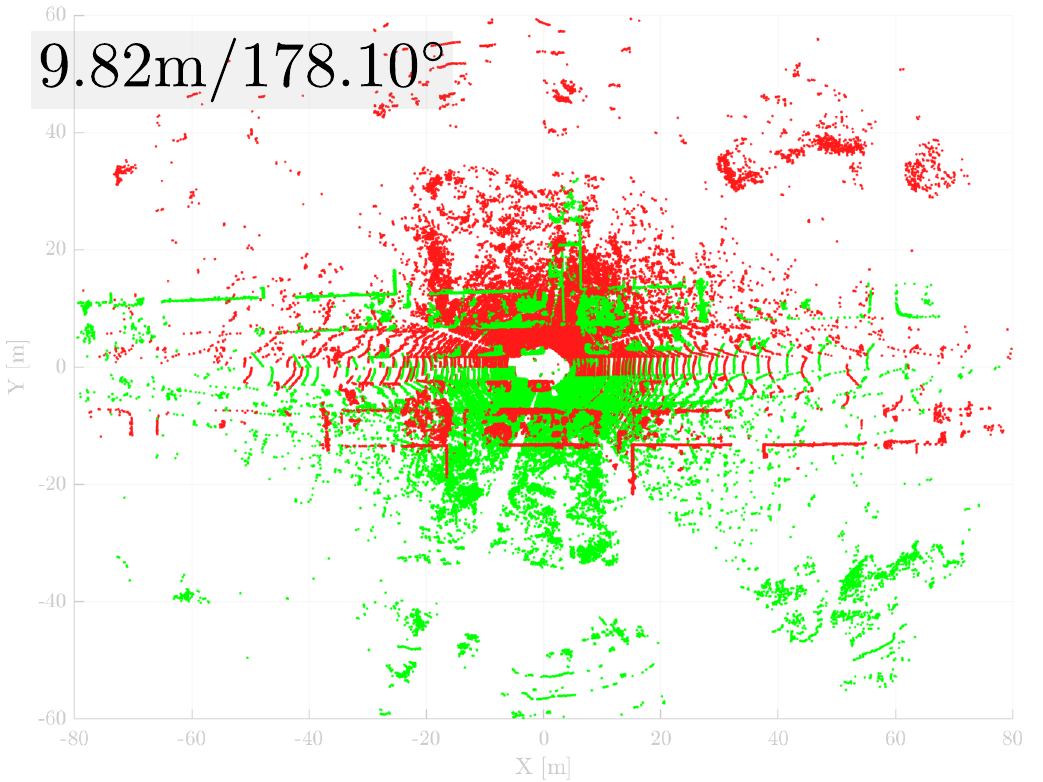}
		\caption{\centering Source and target}
	\end{subfigure}
	\begin{subfigure}[b]{0.19\textwidth}
		\includegraphics[width=1.0\textwidth]{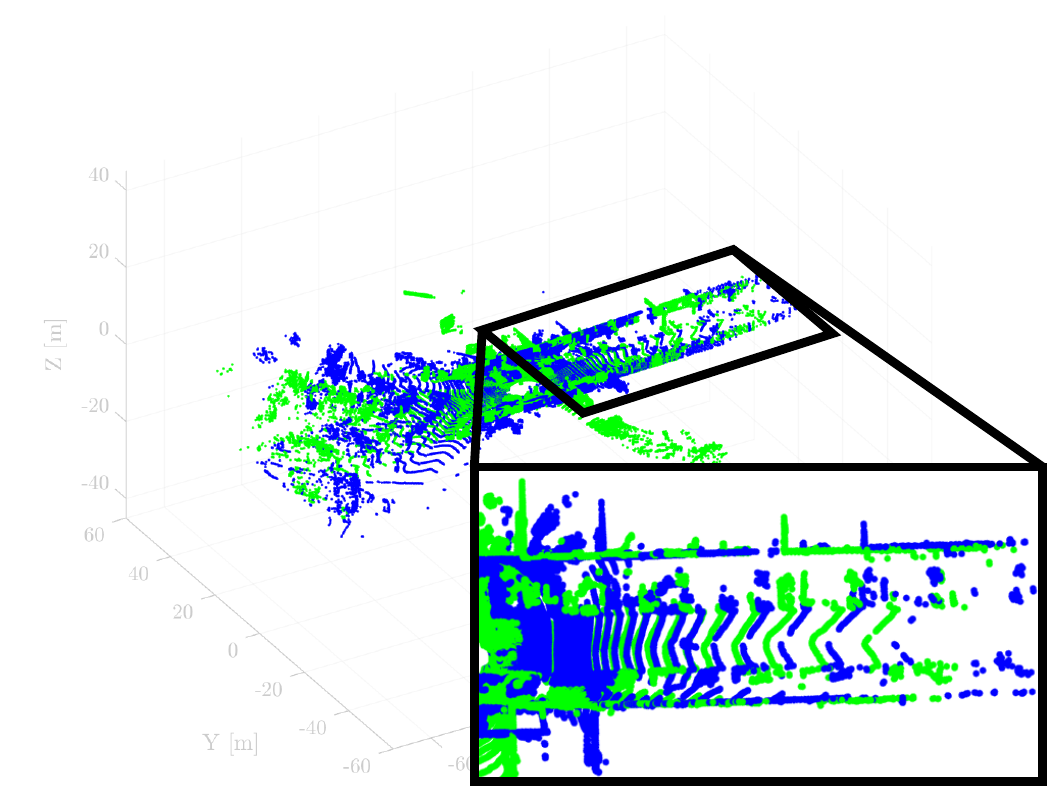}
		\includegraphics[width=1.0\textwidth]{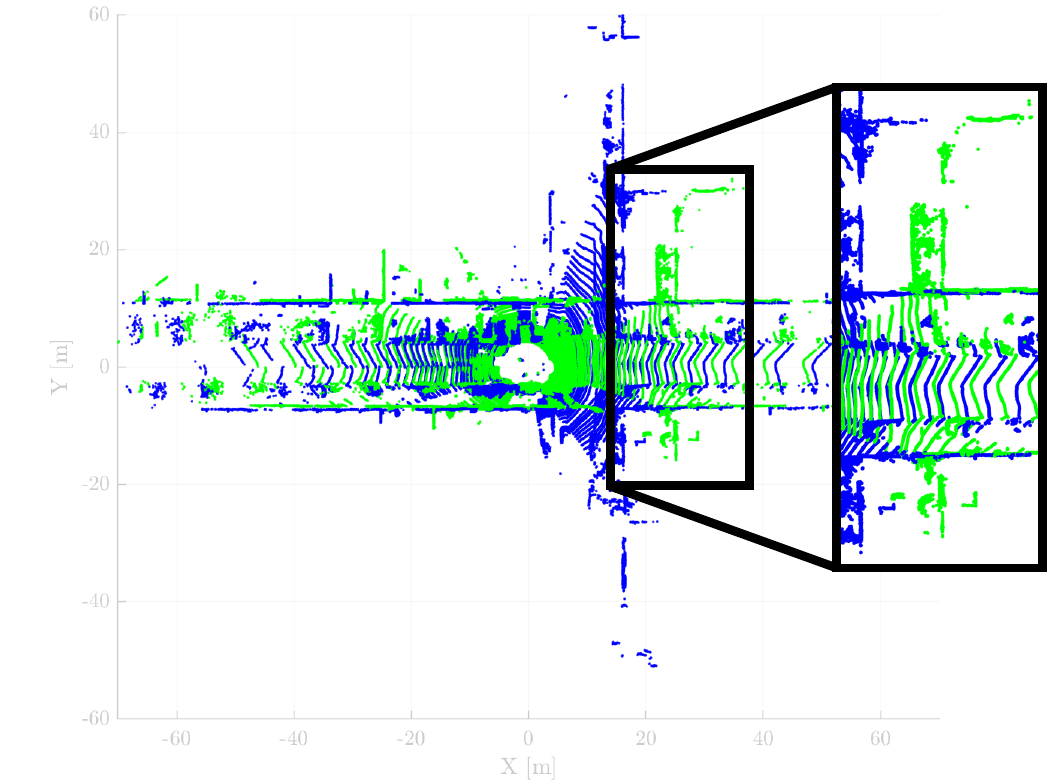}
		\includegraphics[width=1.0\textwidth]{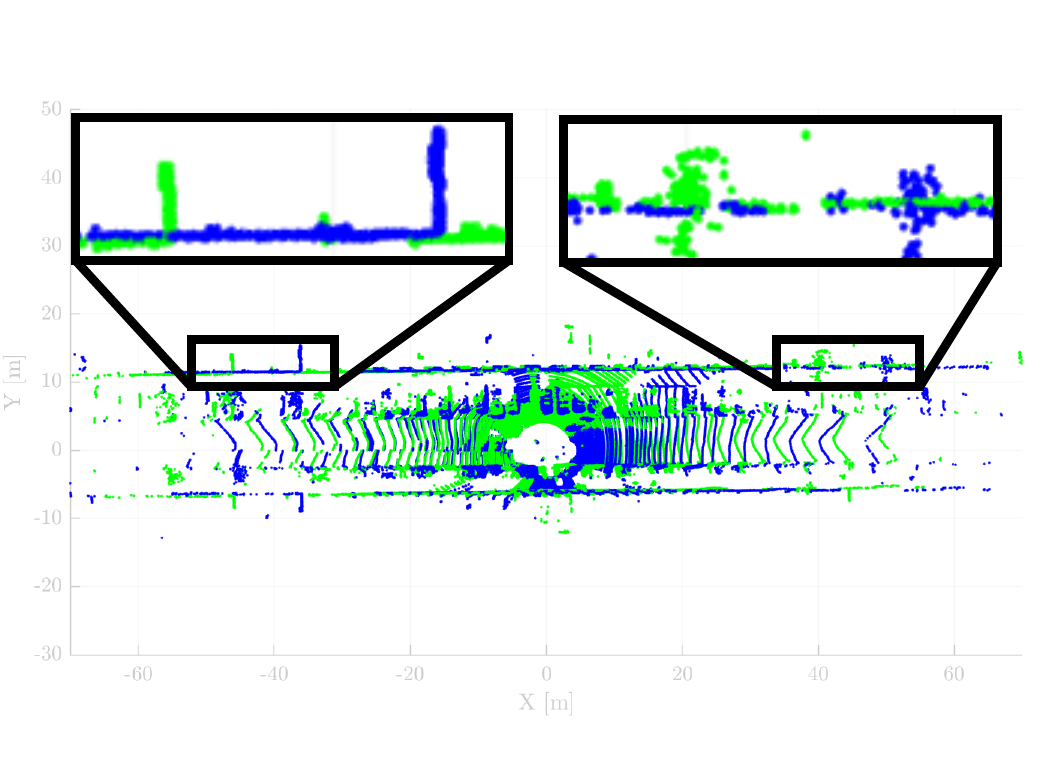}
		\includegraphics[width=1.0\textwidth]{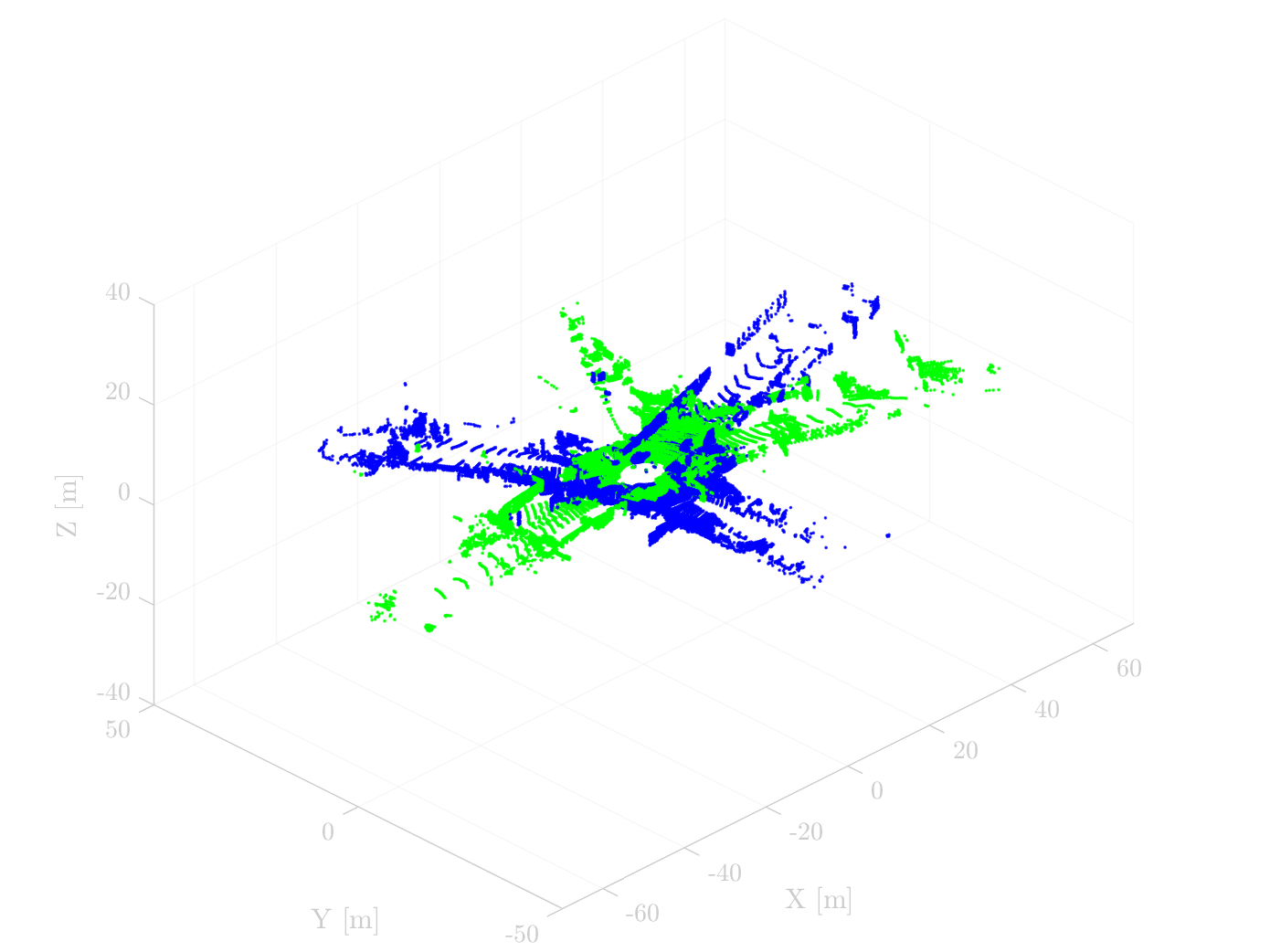}
		\includegraphics[width=1.0\textwidth]{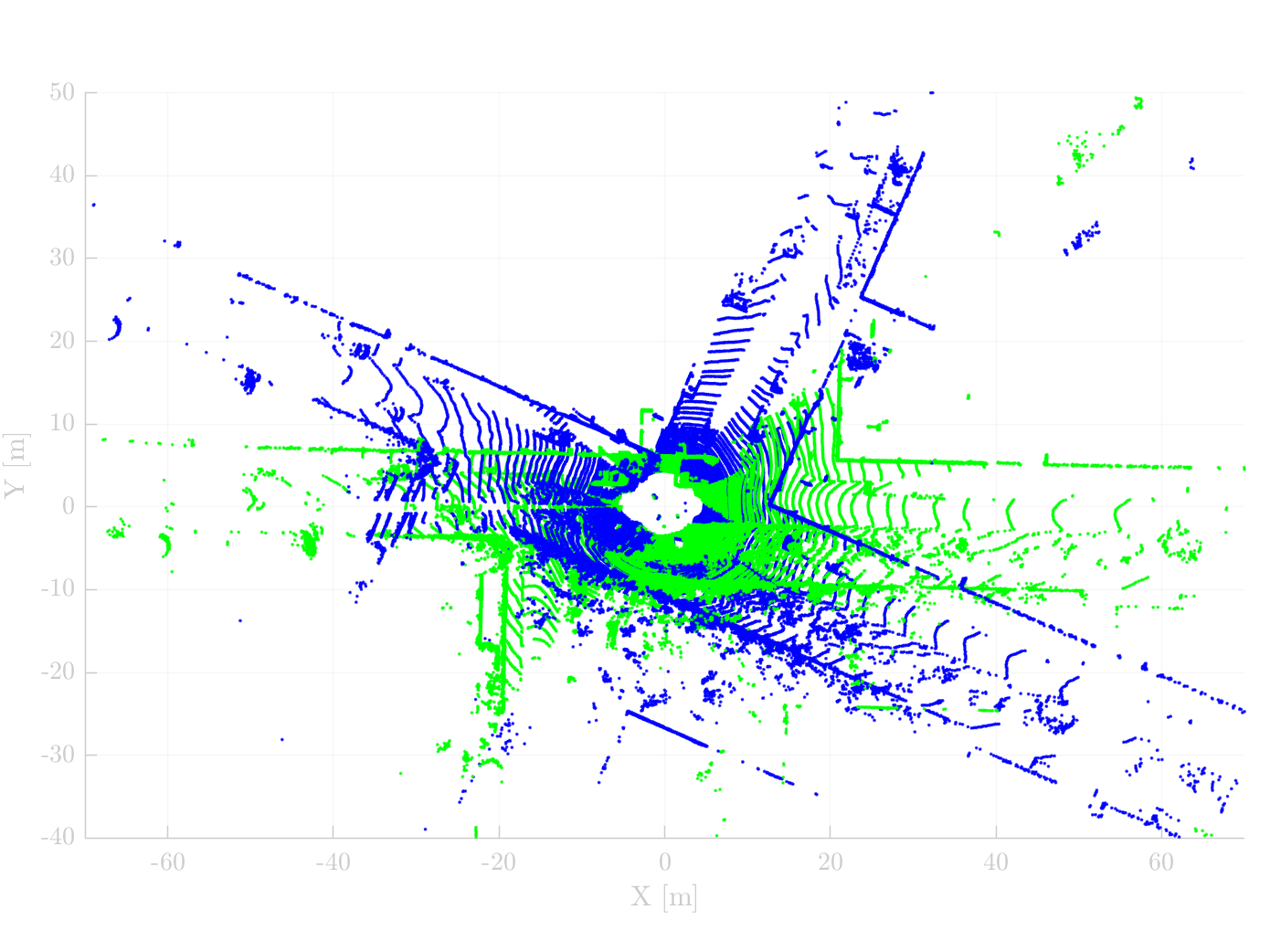}
		\includegraphics[width=1.0\textwidth]{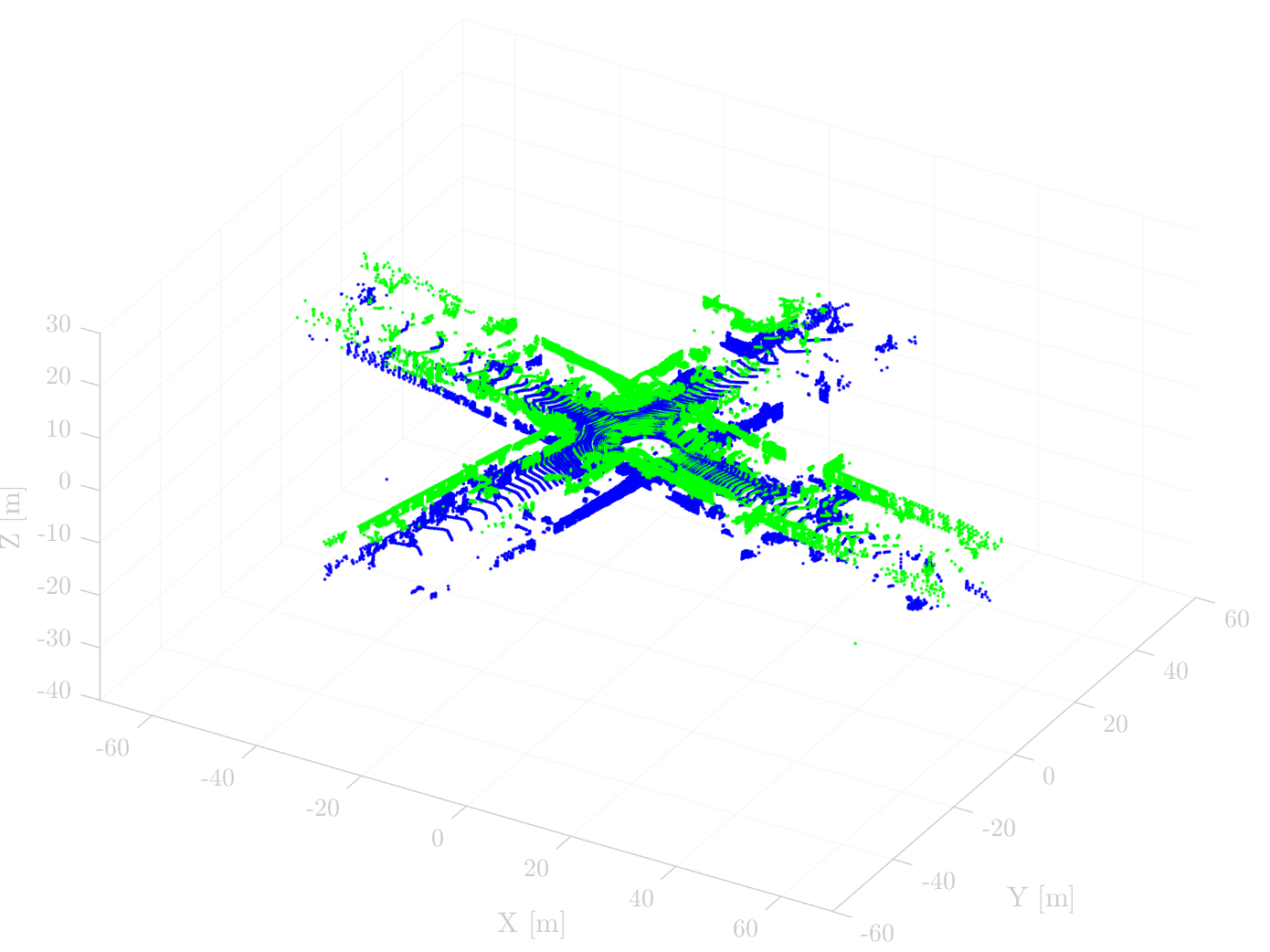}
		\includegraphics[width=1.0\textwidth]{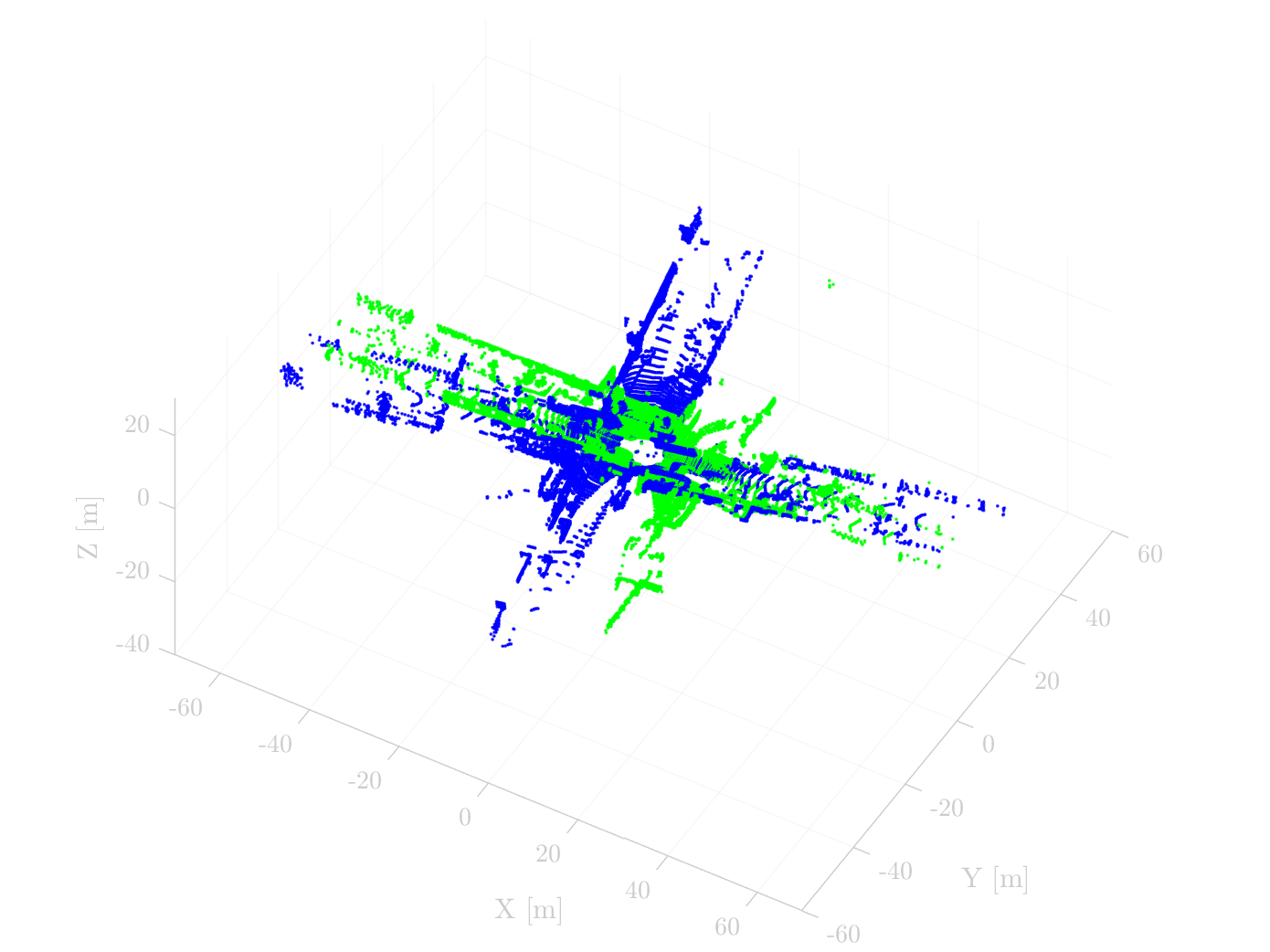}
		\includegraphics[width=1.0\textwidth]{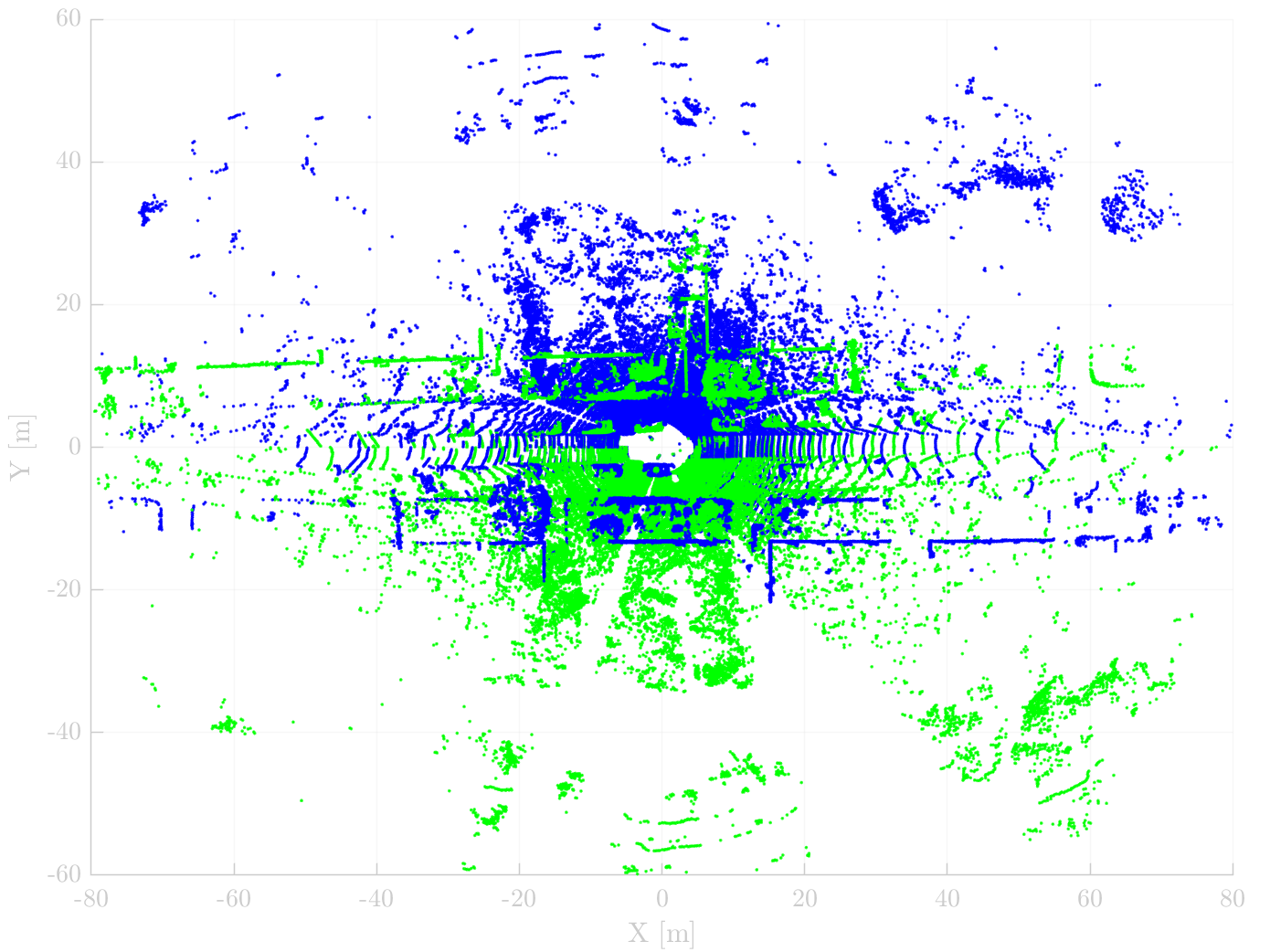}
		\caption{\centering RANSAC}
	\end{subfigure}
	\begin{subfigure}[b]{0.19\textwidth}
		\includegraphics[width=1.0\textwidth]{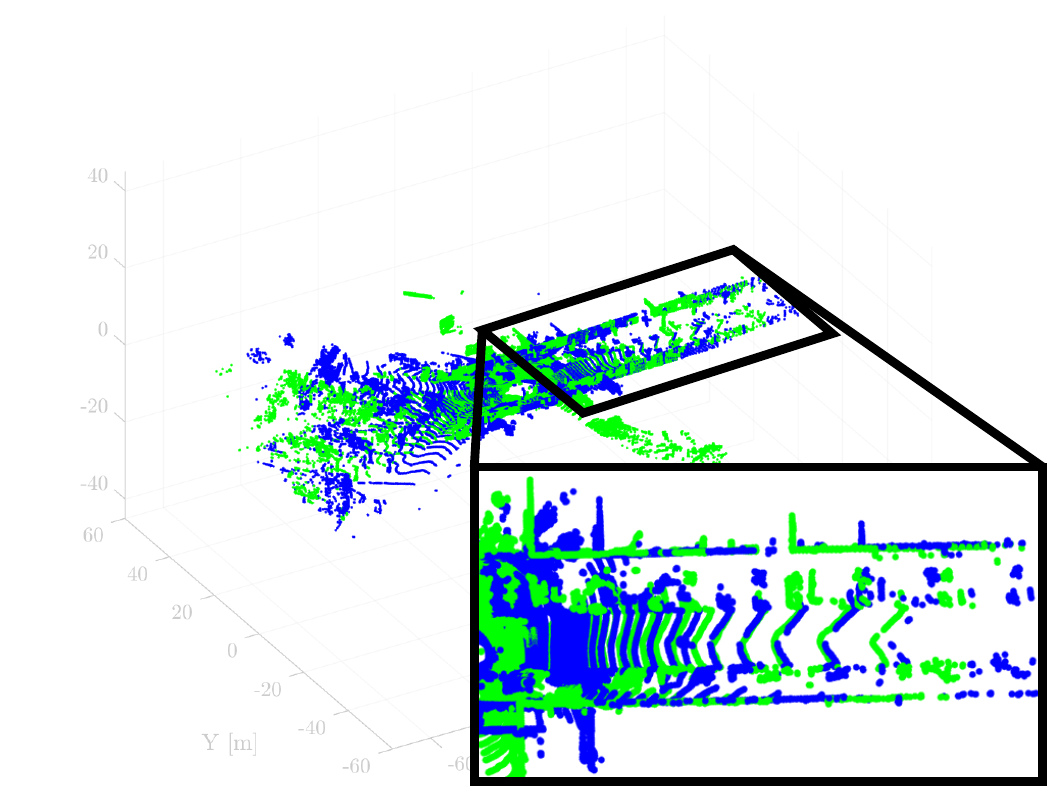}
		\includegraphics[width=1.0\textwidth]{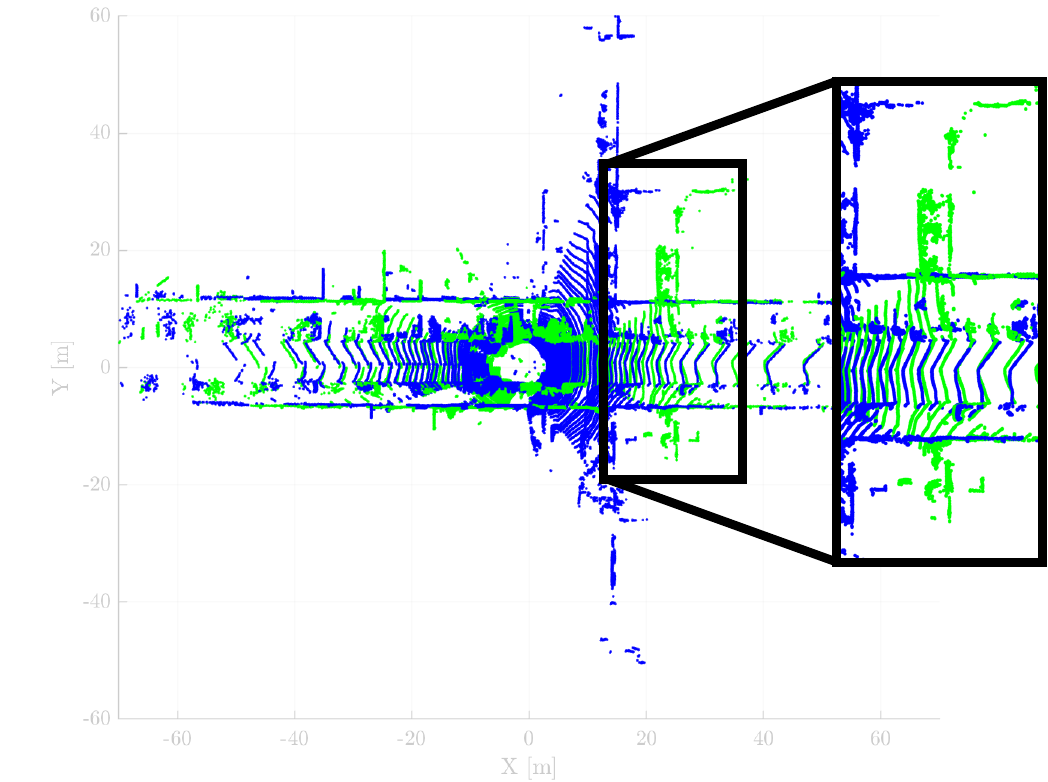}
		\includegraphics[width=1.0\textwidth]{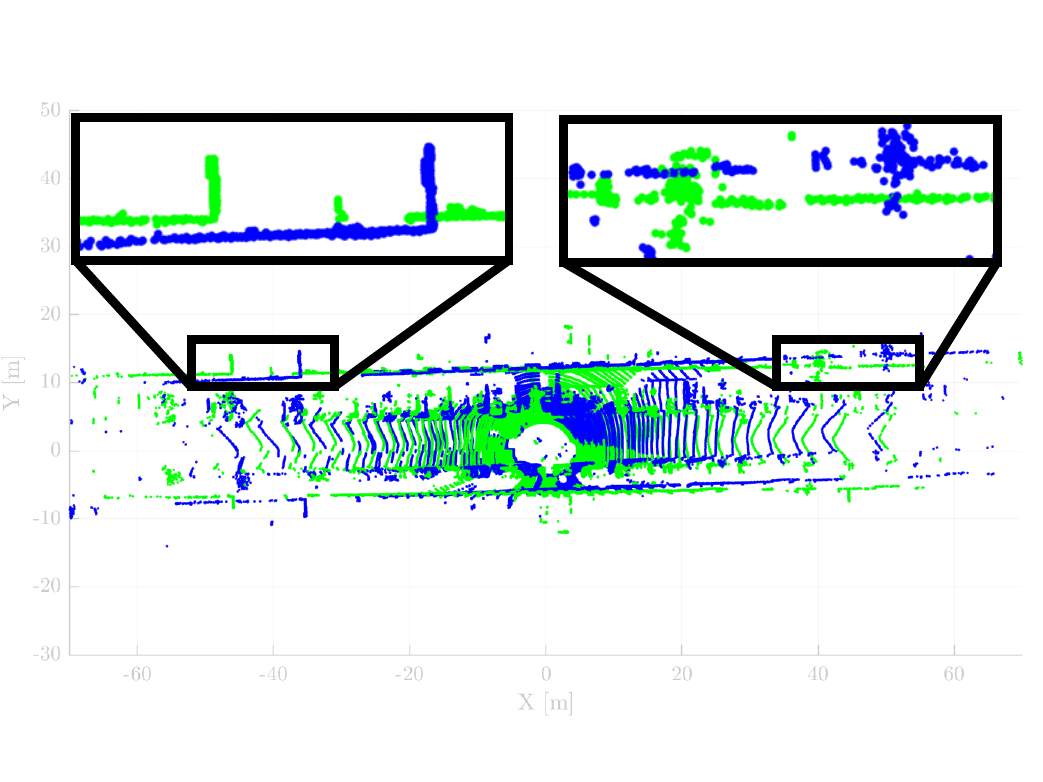}
		\includegraphics[width=1.0\textwidth]{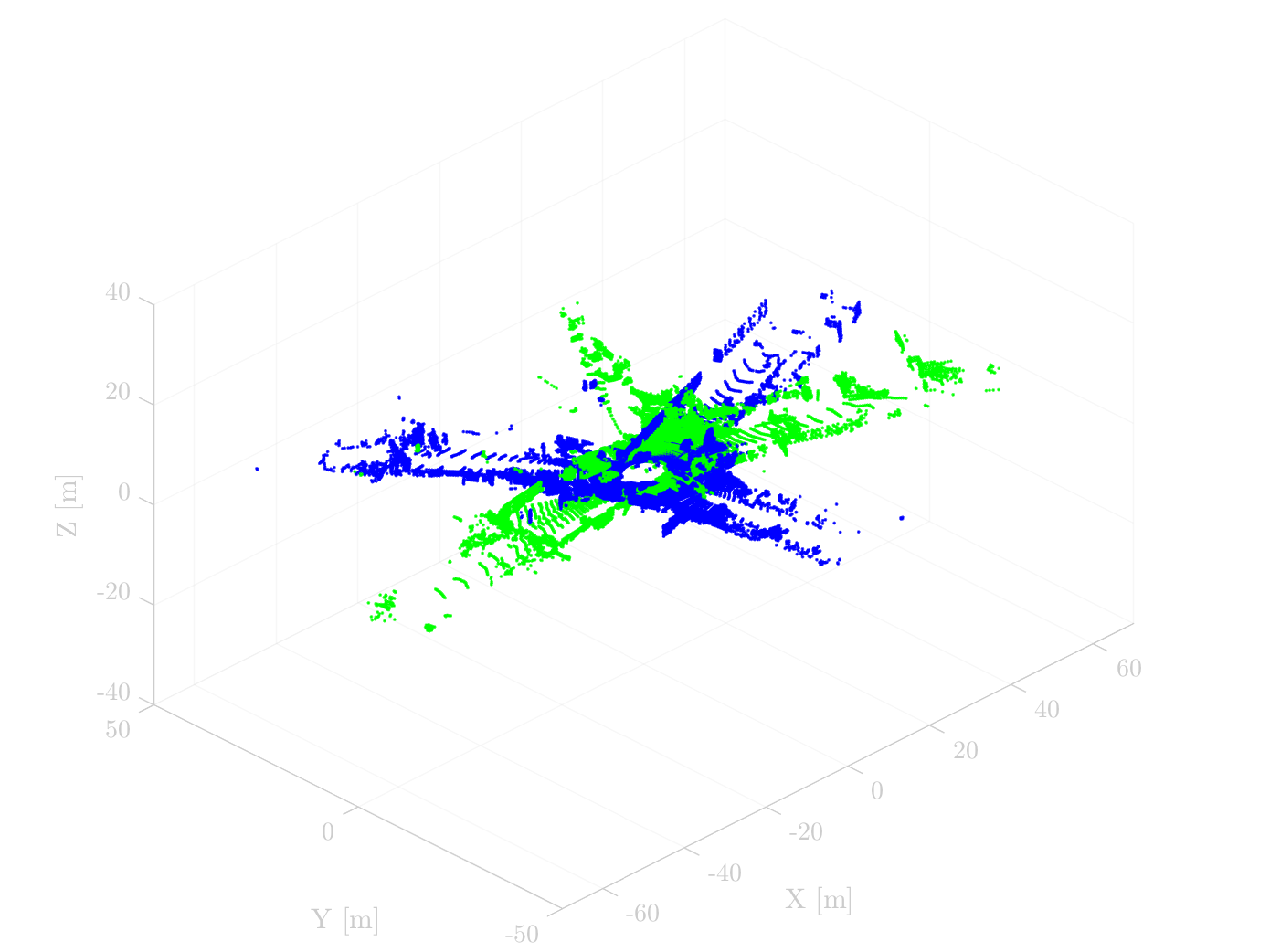}
		\includegraphics[width=1.0\textwidth]{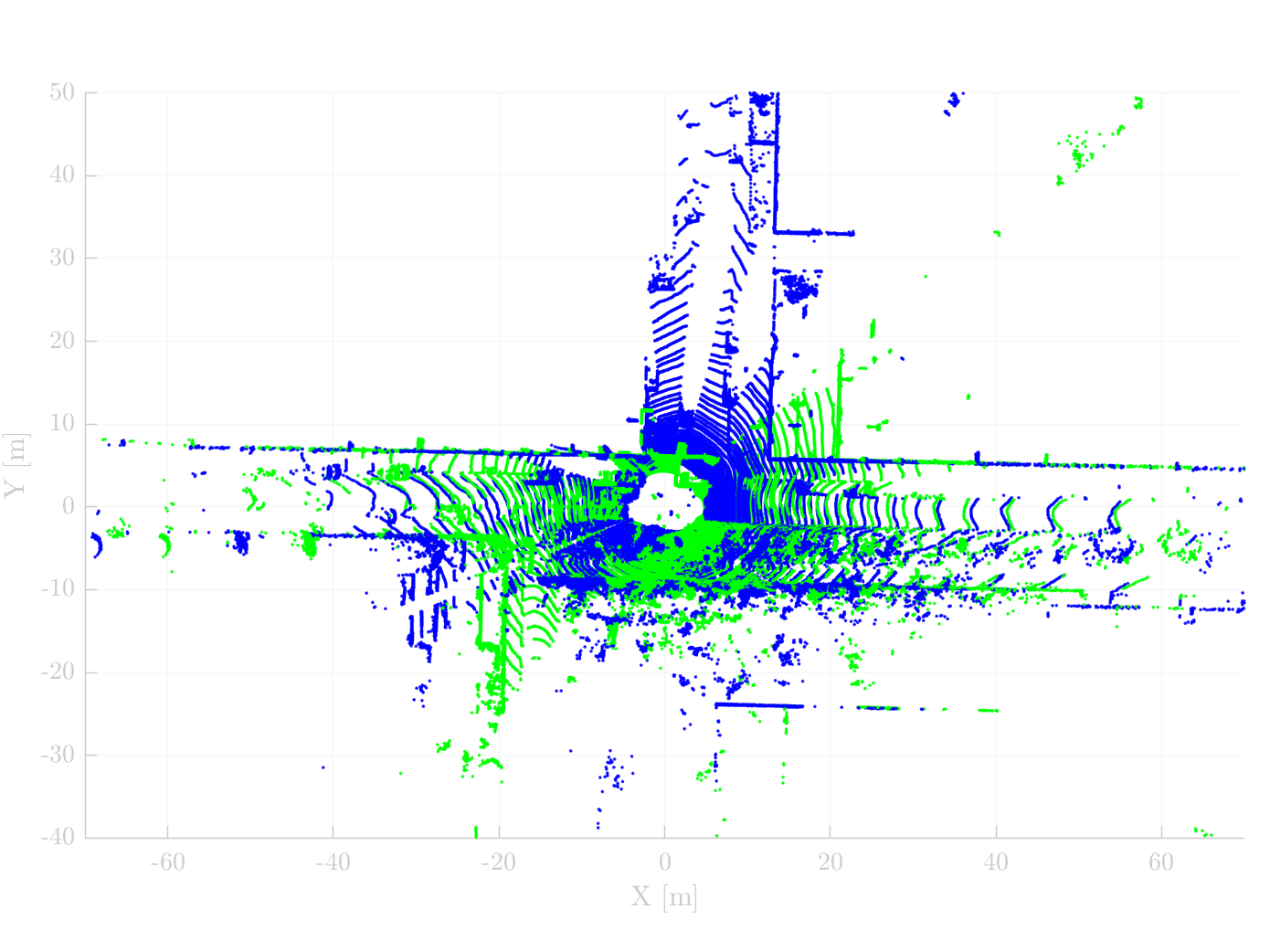}
		\includegraphics[width=1.0\textwidth]{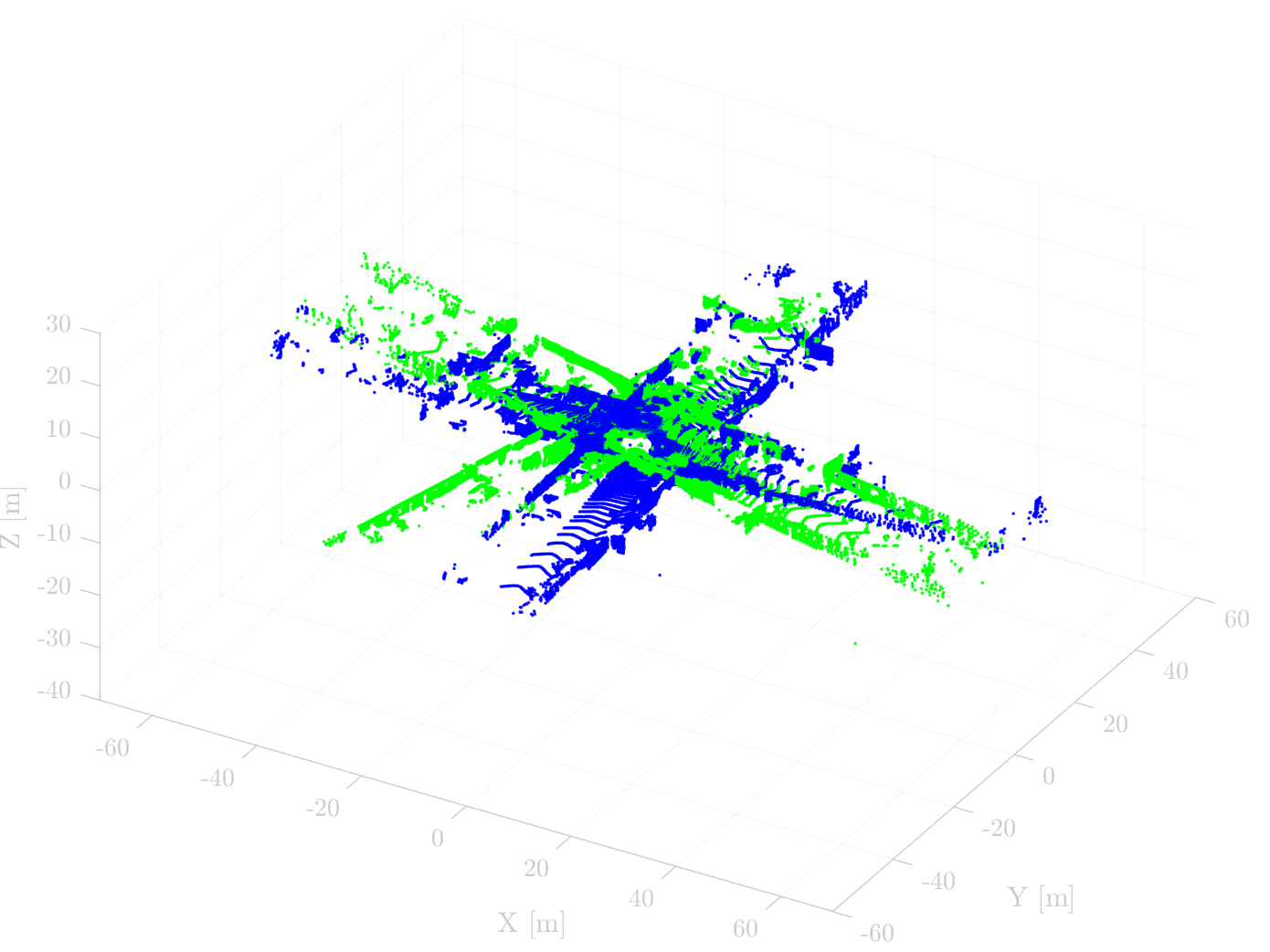}
		\includegraphics[width=1.0\textwidth]{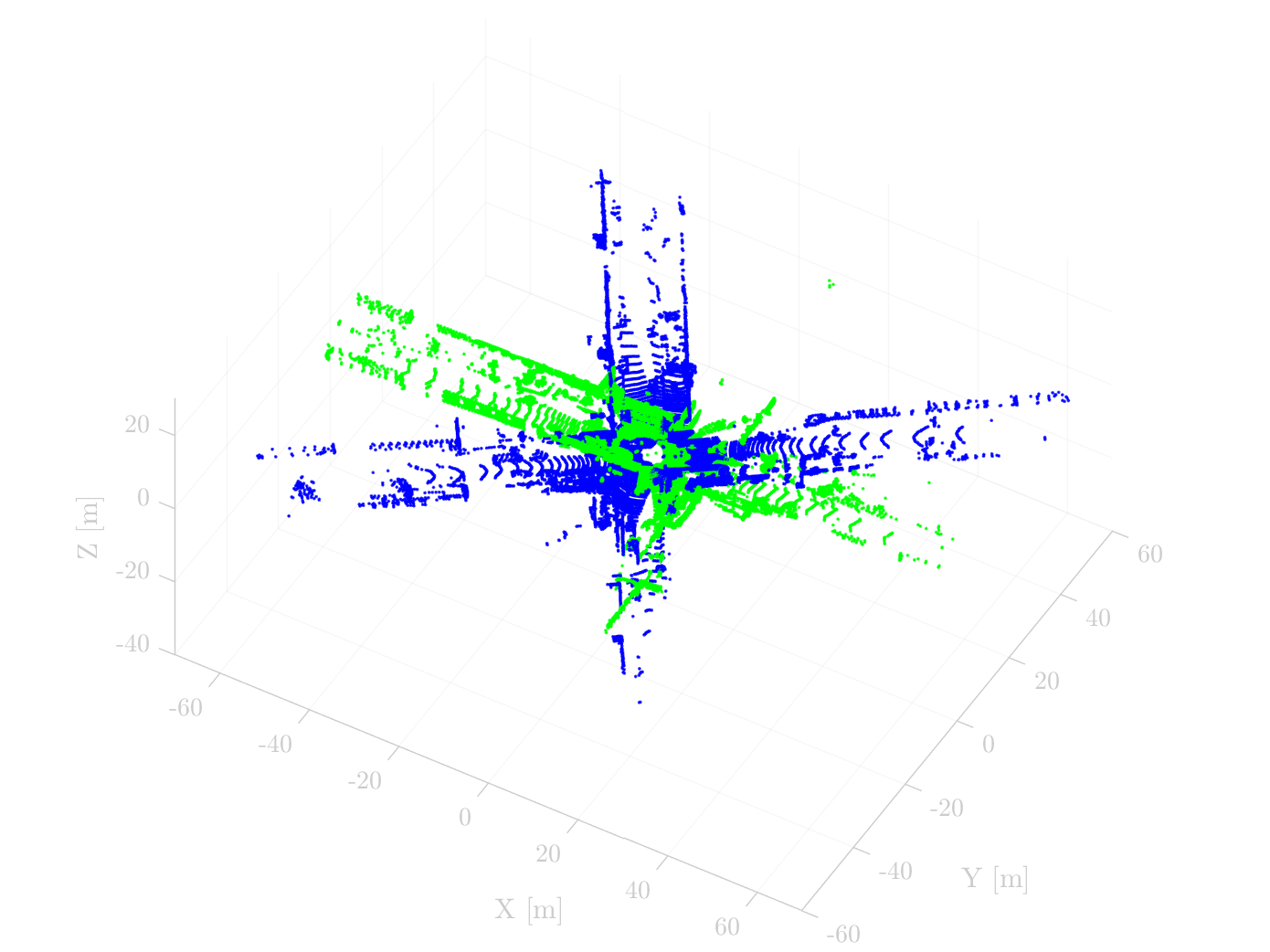}
		\includegraphics[width=1.0\textwidth]{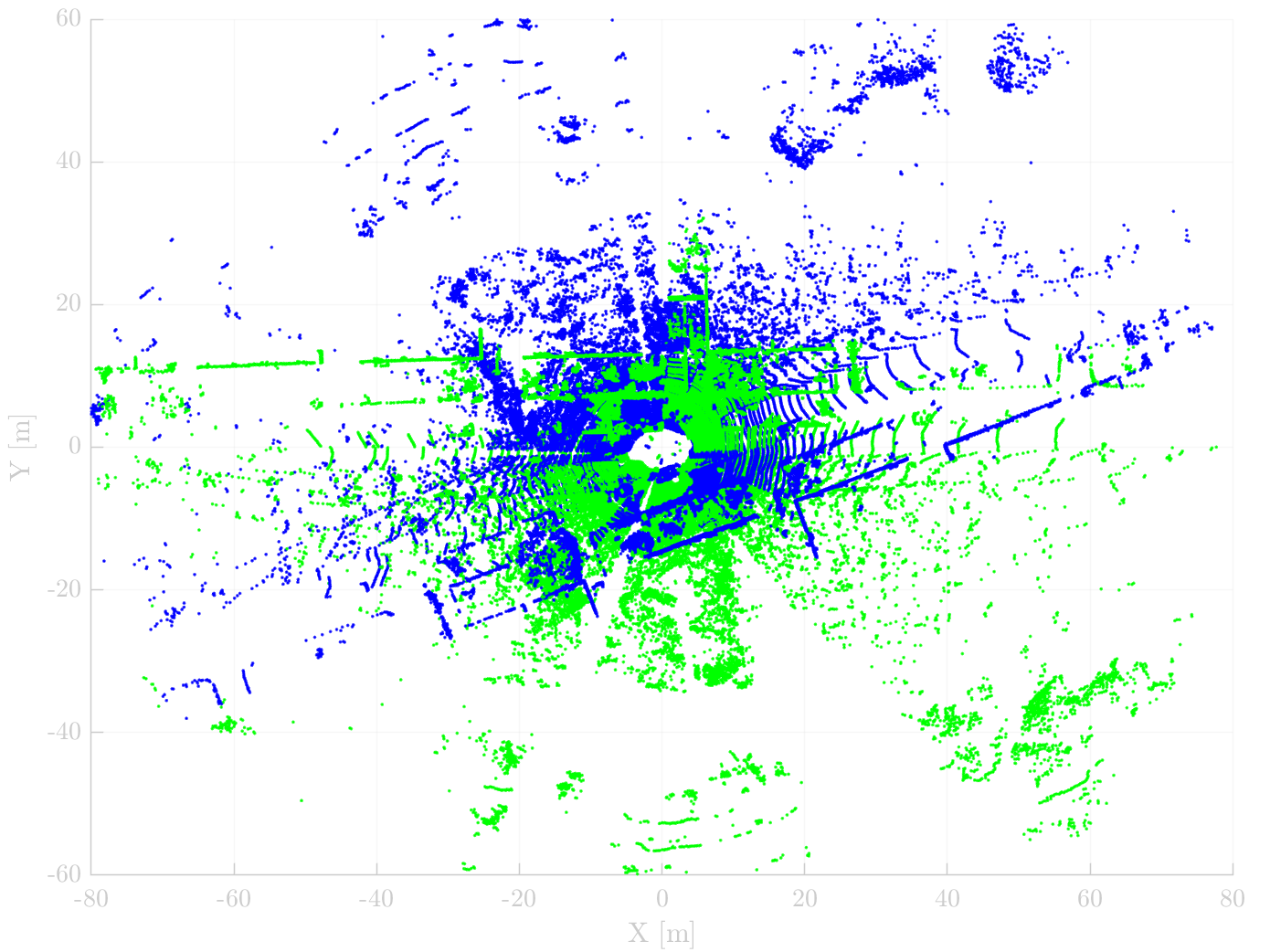}
		\caption{\centering FGR}
	\end{subfigure}
	\begin{subfigure}[b]{0.19\textwidth}
		\includegraphics[width=1.0\textwidth]{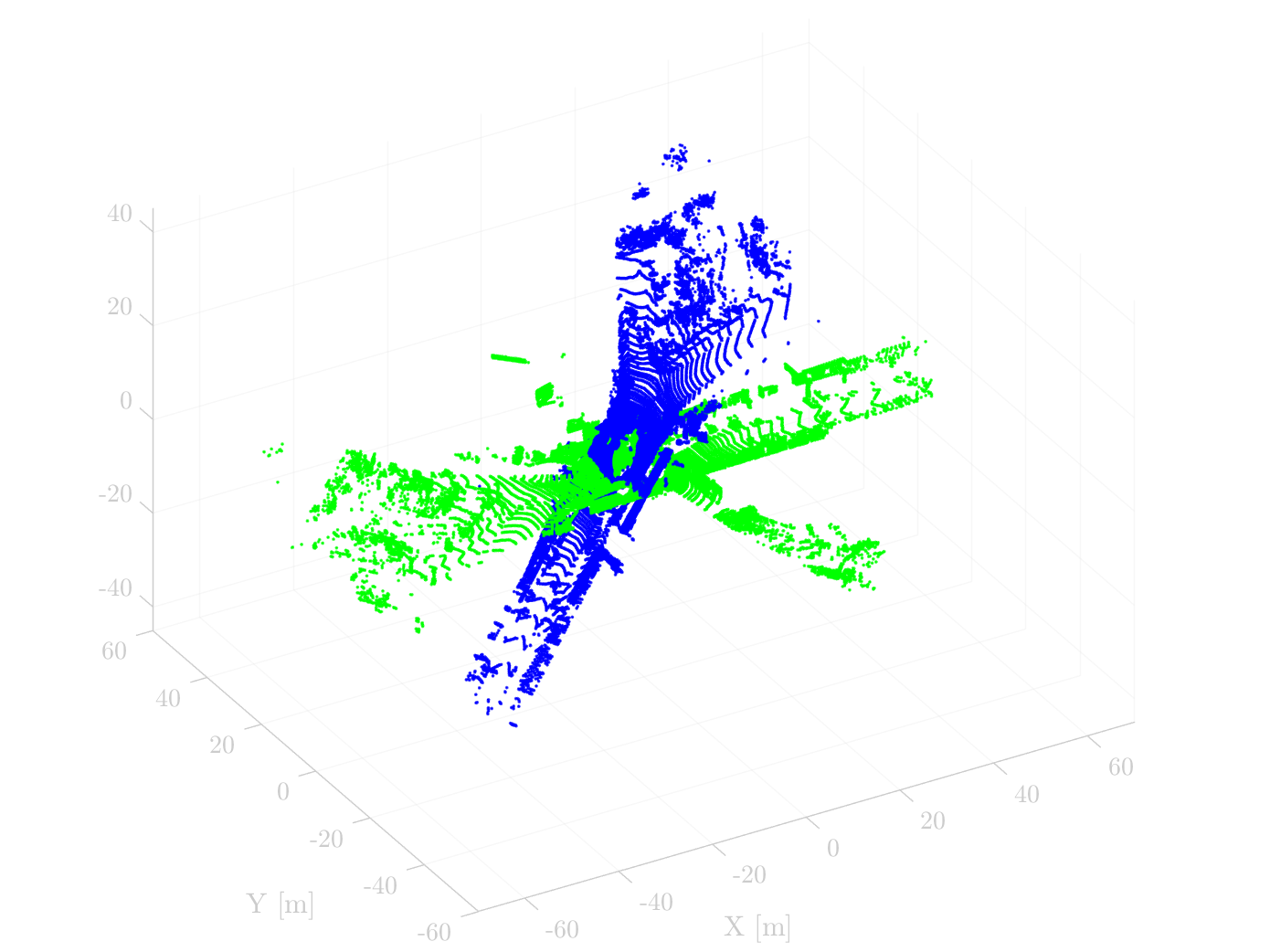}
		\includegraphics[width=1.0\textwidth]{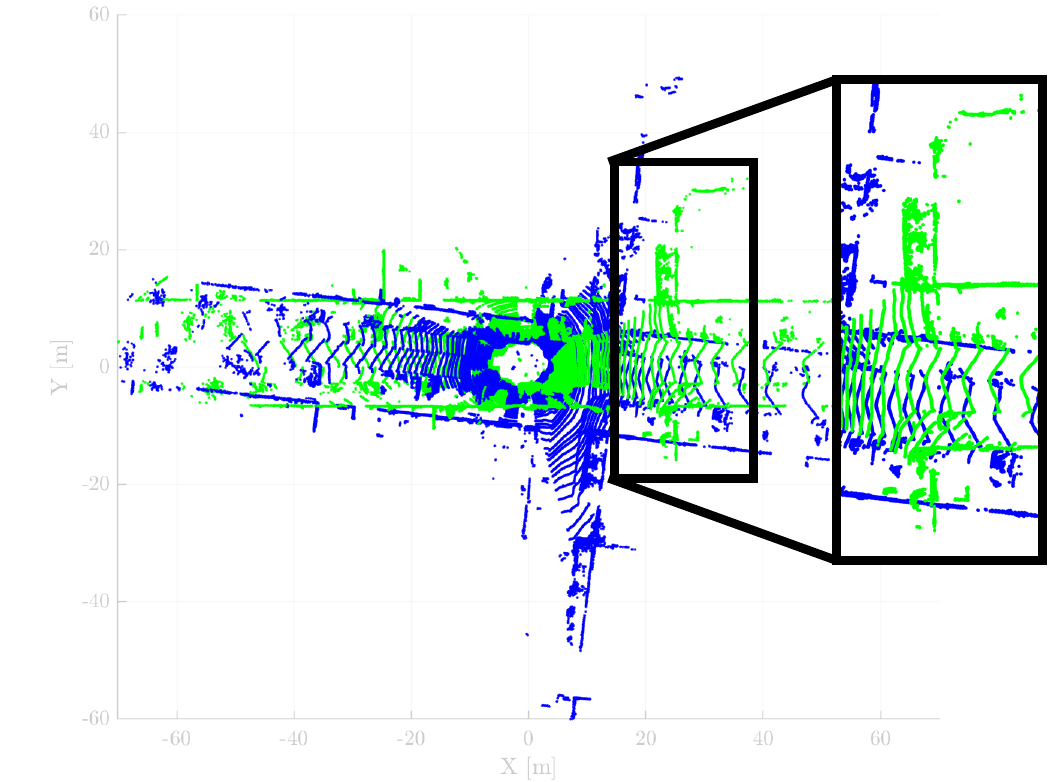}
		\includegraphics[width=1.0\textwidth]{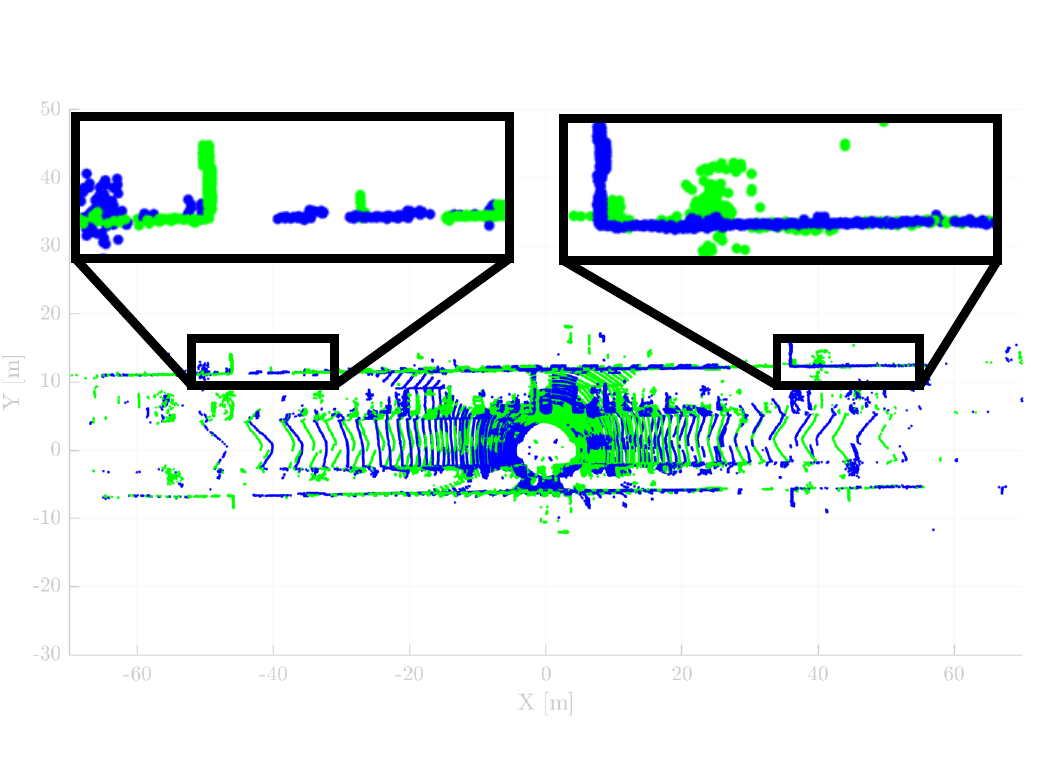}
		\includegraphics[width=1.0\textwidth]{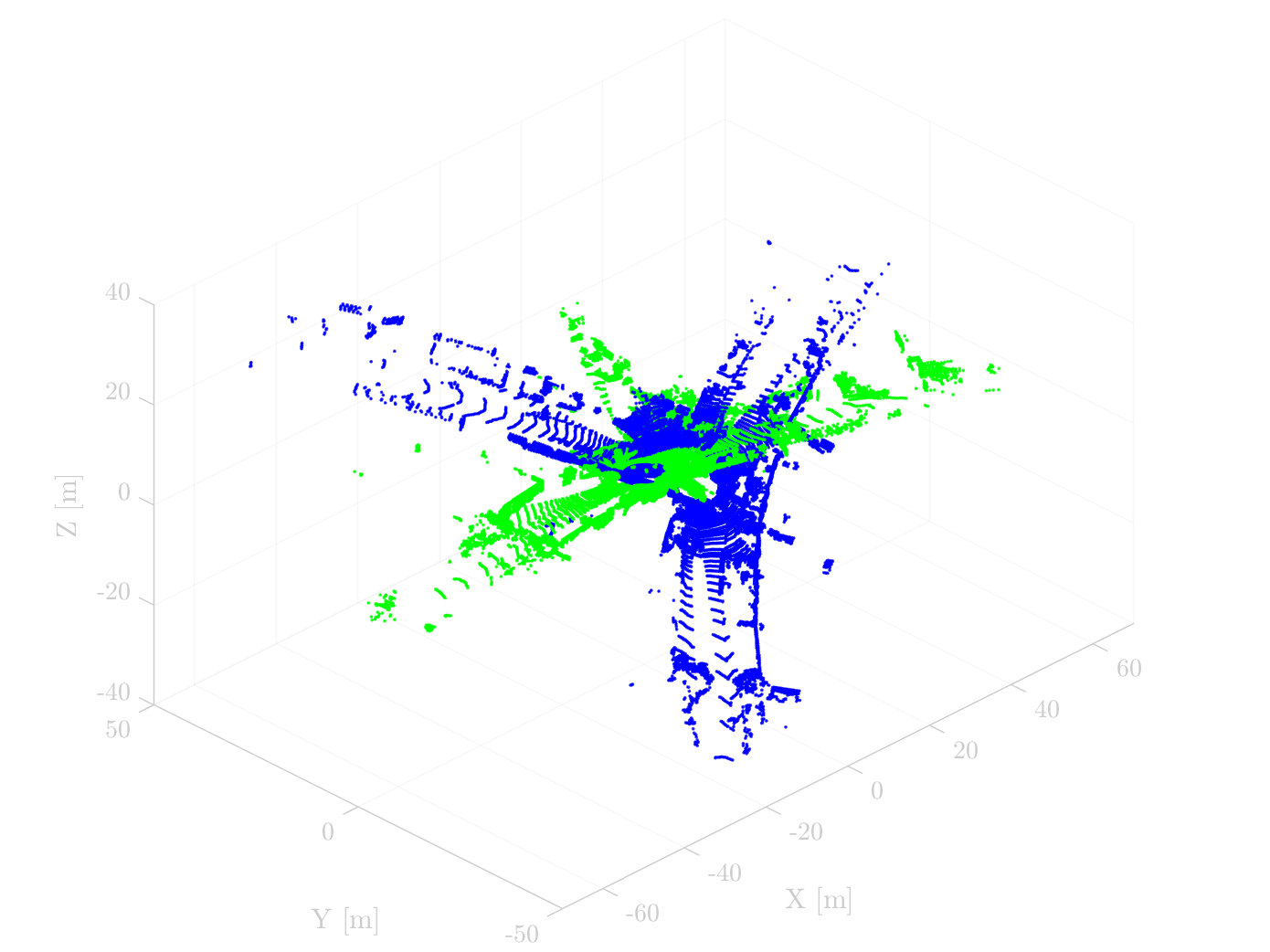}
		\includegraphics[width=1.0\textwidth]{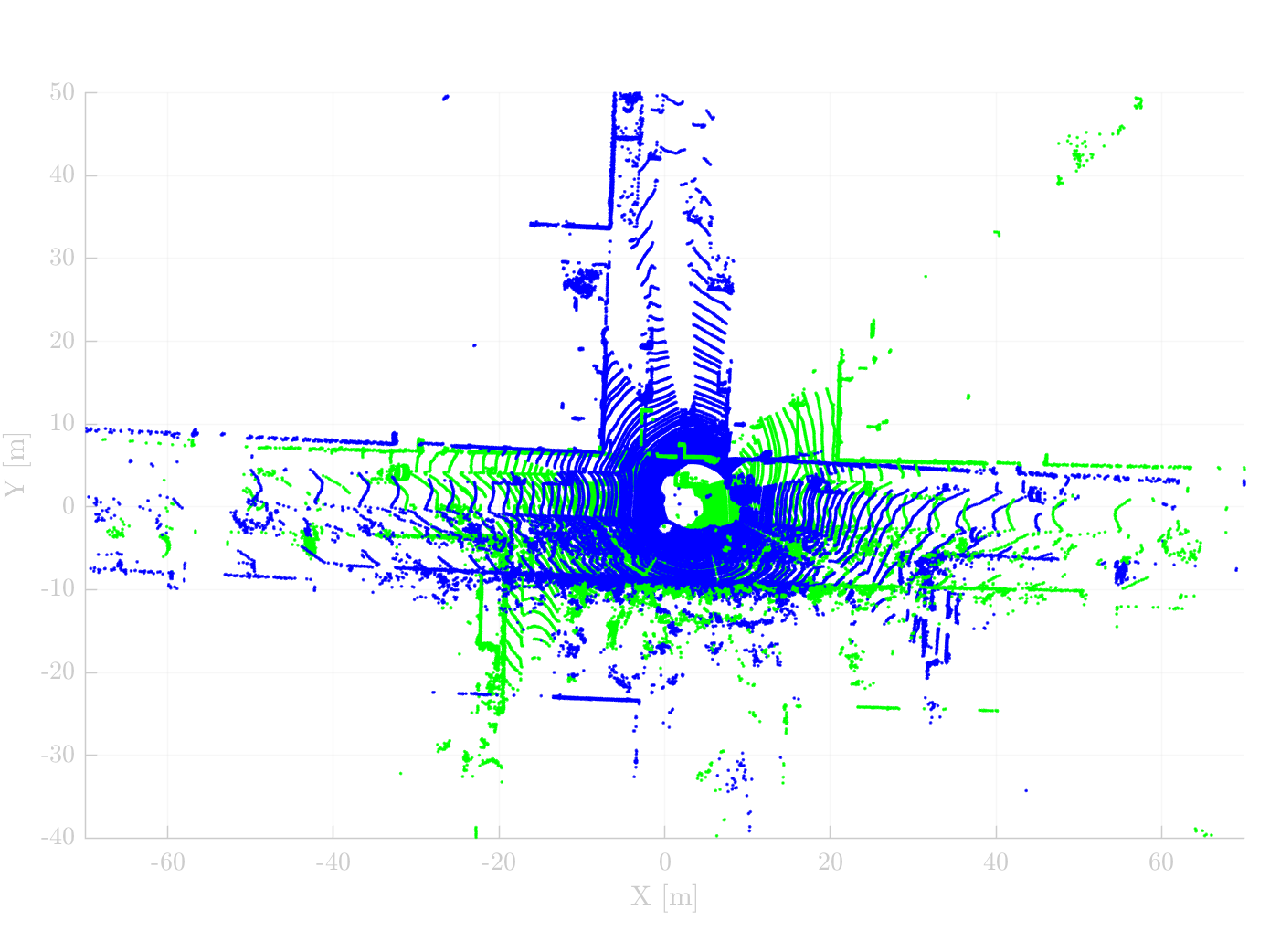}
		\includegraphics[width=1.0\textwidth]{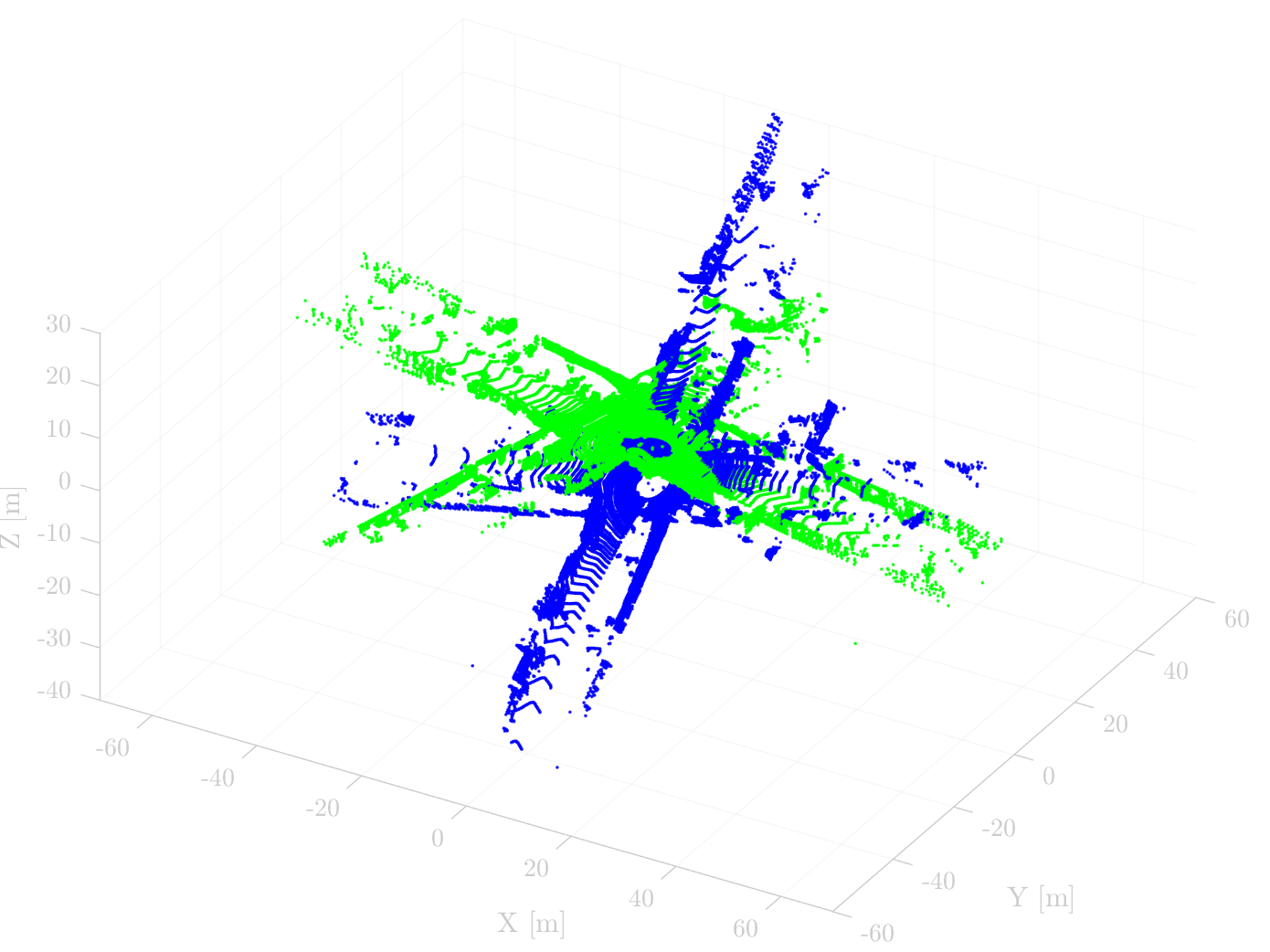}
		\includegraphics[width=1.0\textwidth]{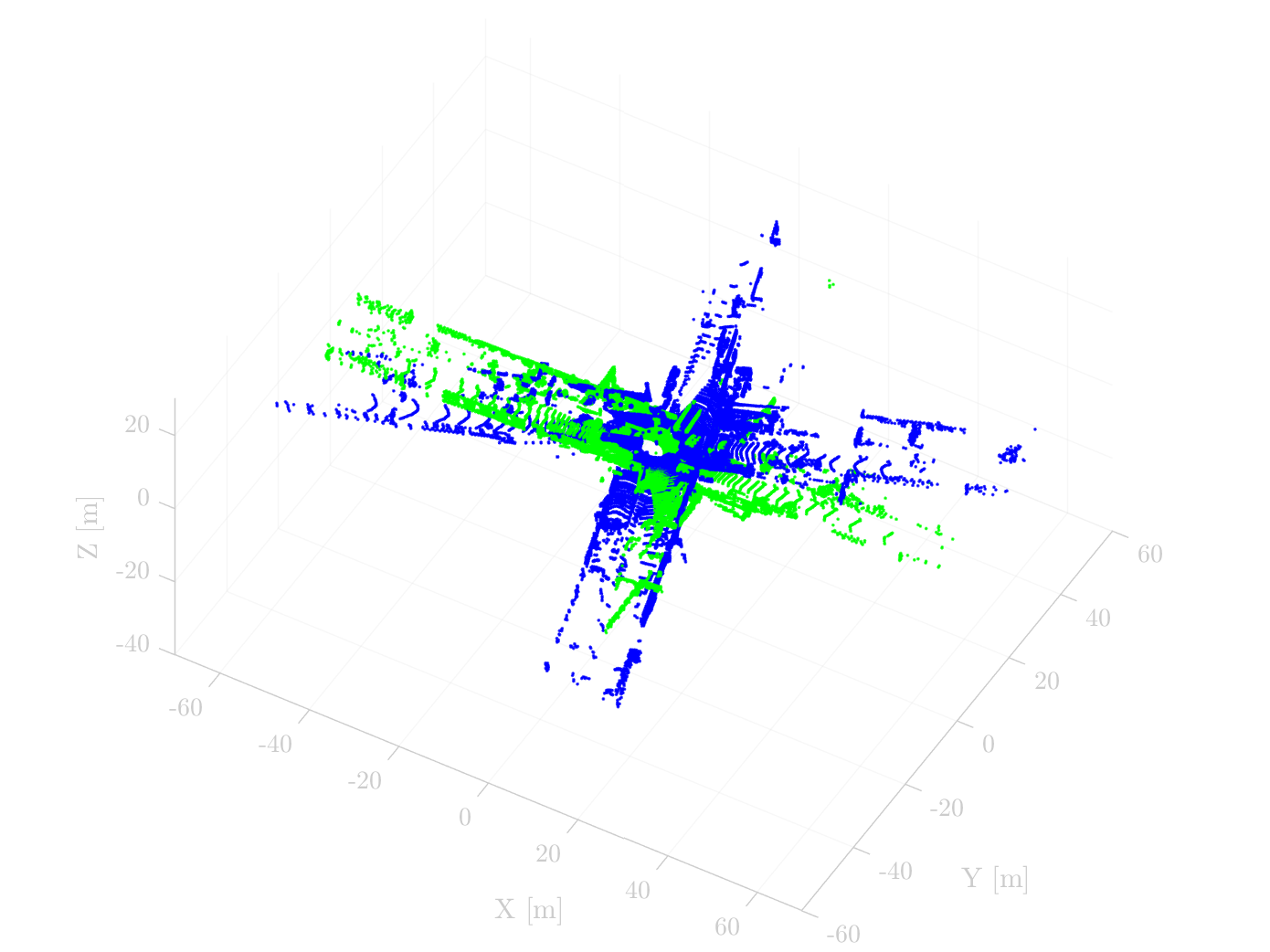}
		\includegraphics[width=1.0\textwidth]{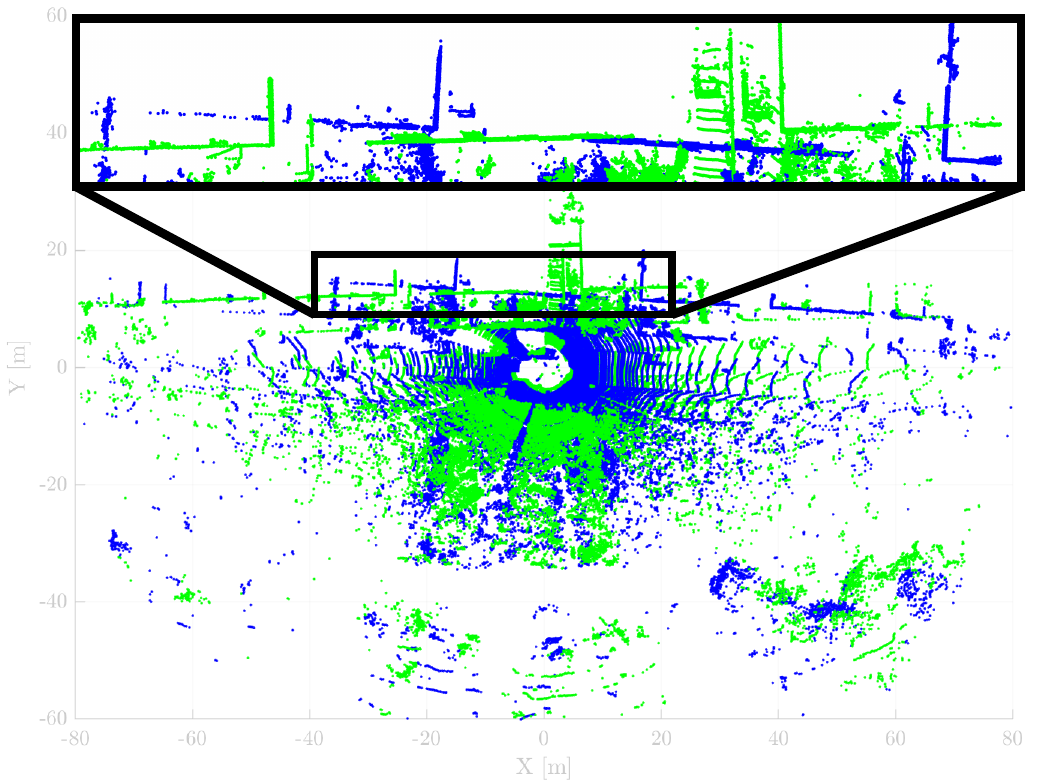}
		\caption{\centering TEASER++}
	\end{subfigure}
	\begin{subfigure}[b]{0.19\textwidth}
		\includegraphics[width=1.0\textwidth]{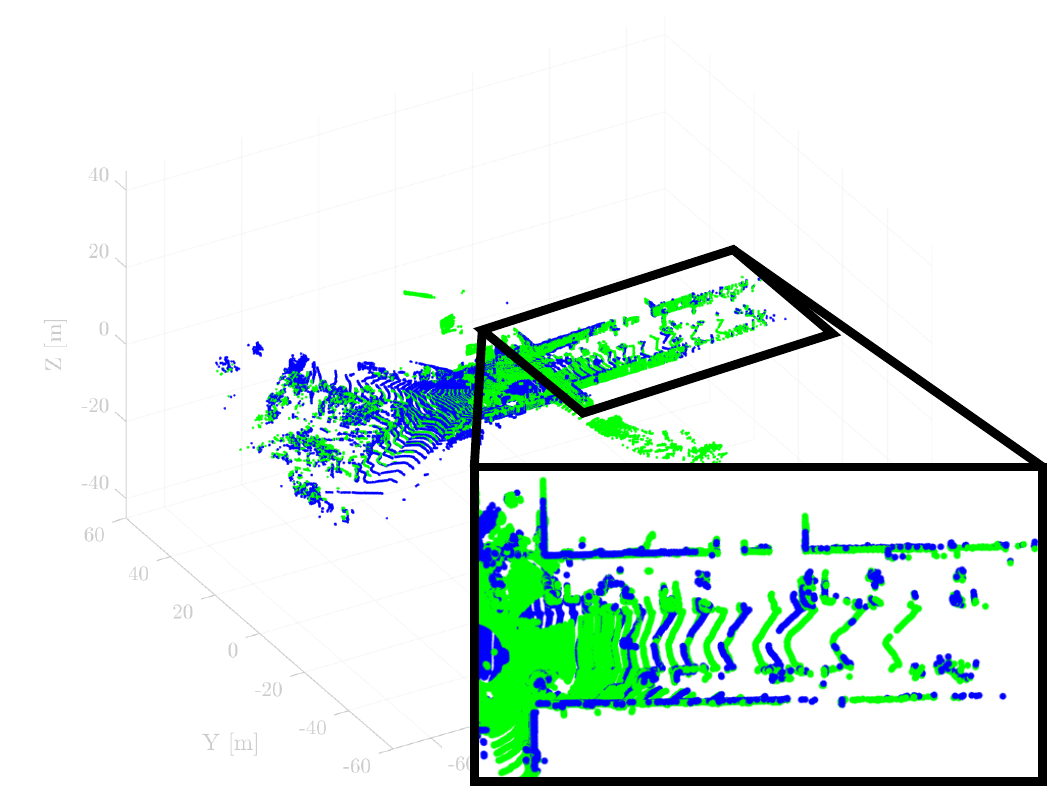}
		\includegraphics[width=1.0\textwidth]{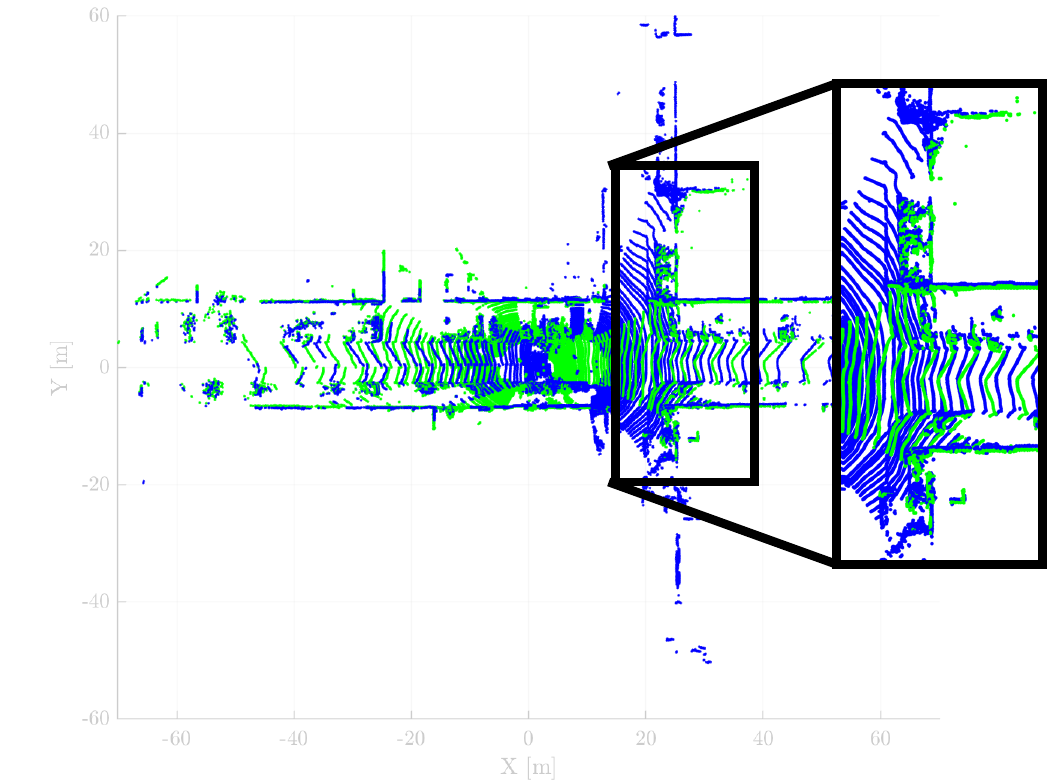}
		\includegraphics[width=1.0\textwidth]{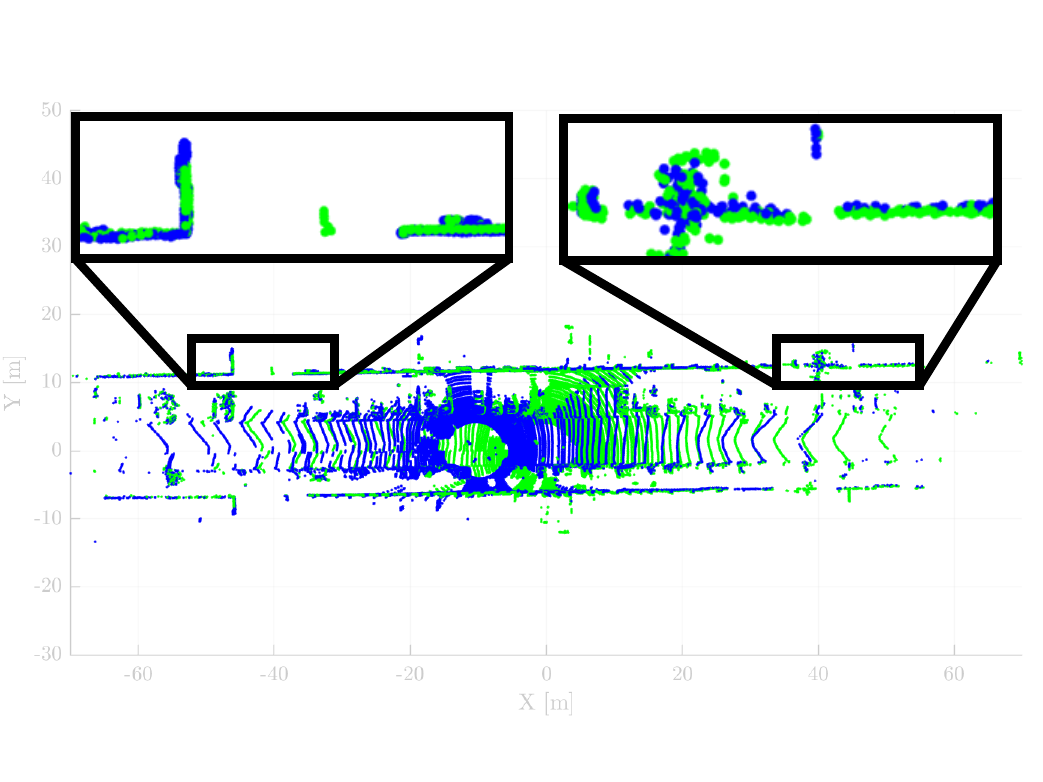}
		\includegraphics[width=1.0\textwidth]{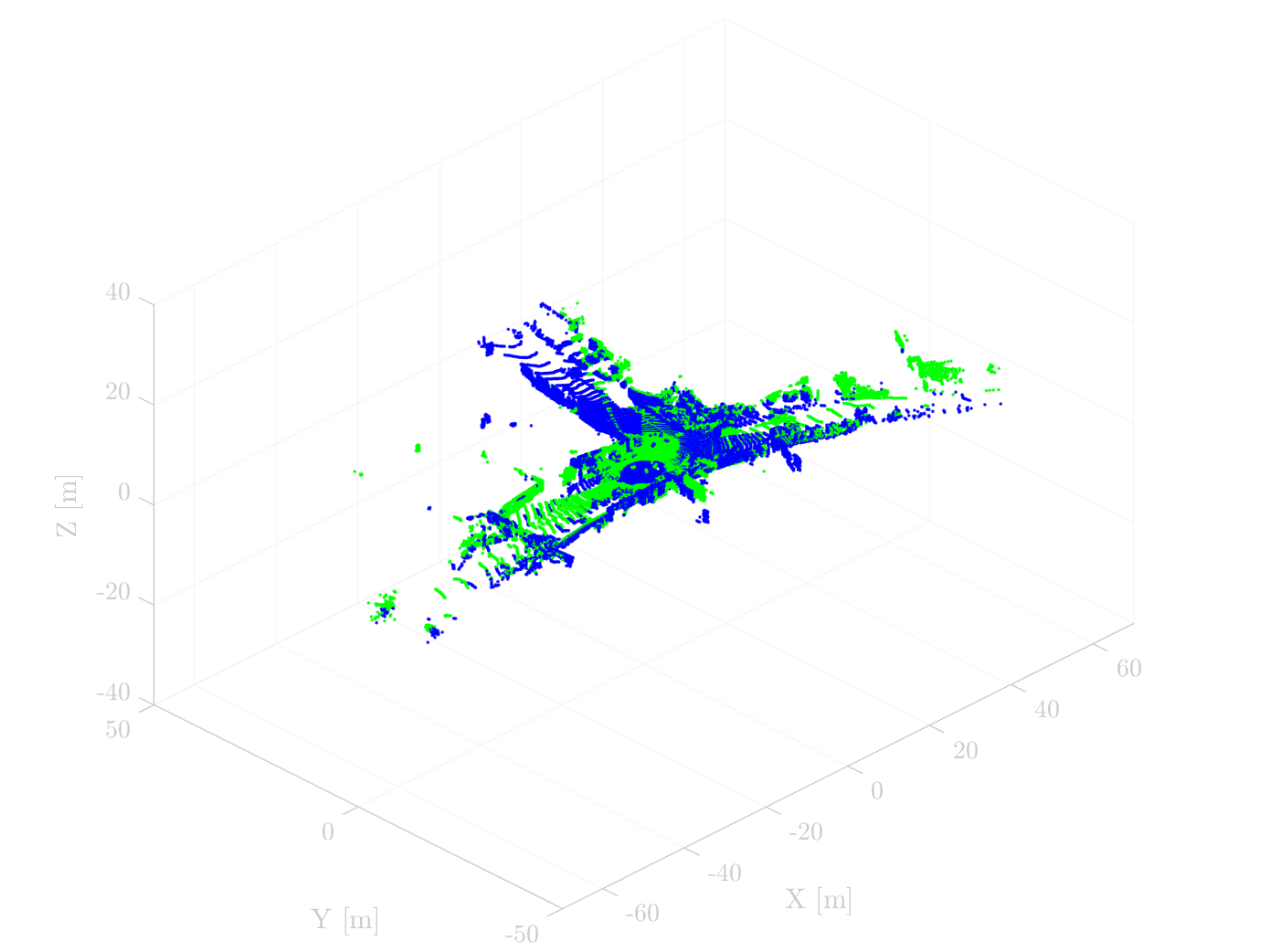}
		\includegraphics[width=1.0\textwidth]{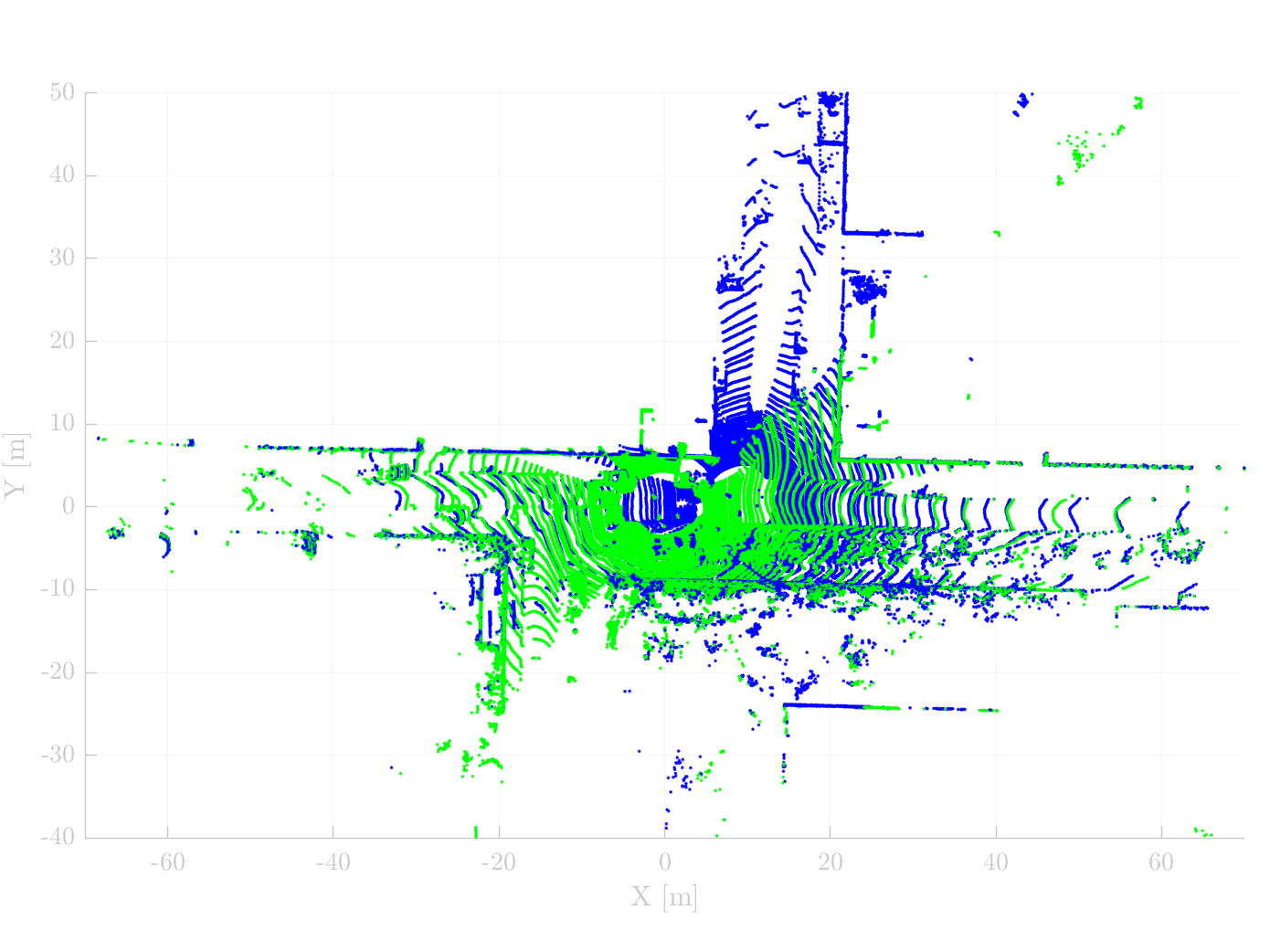}
		\includegraphics[width=1.0\textwidth]{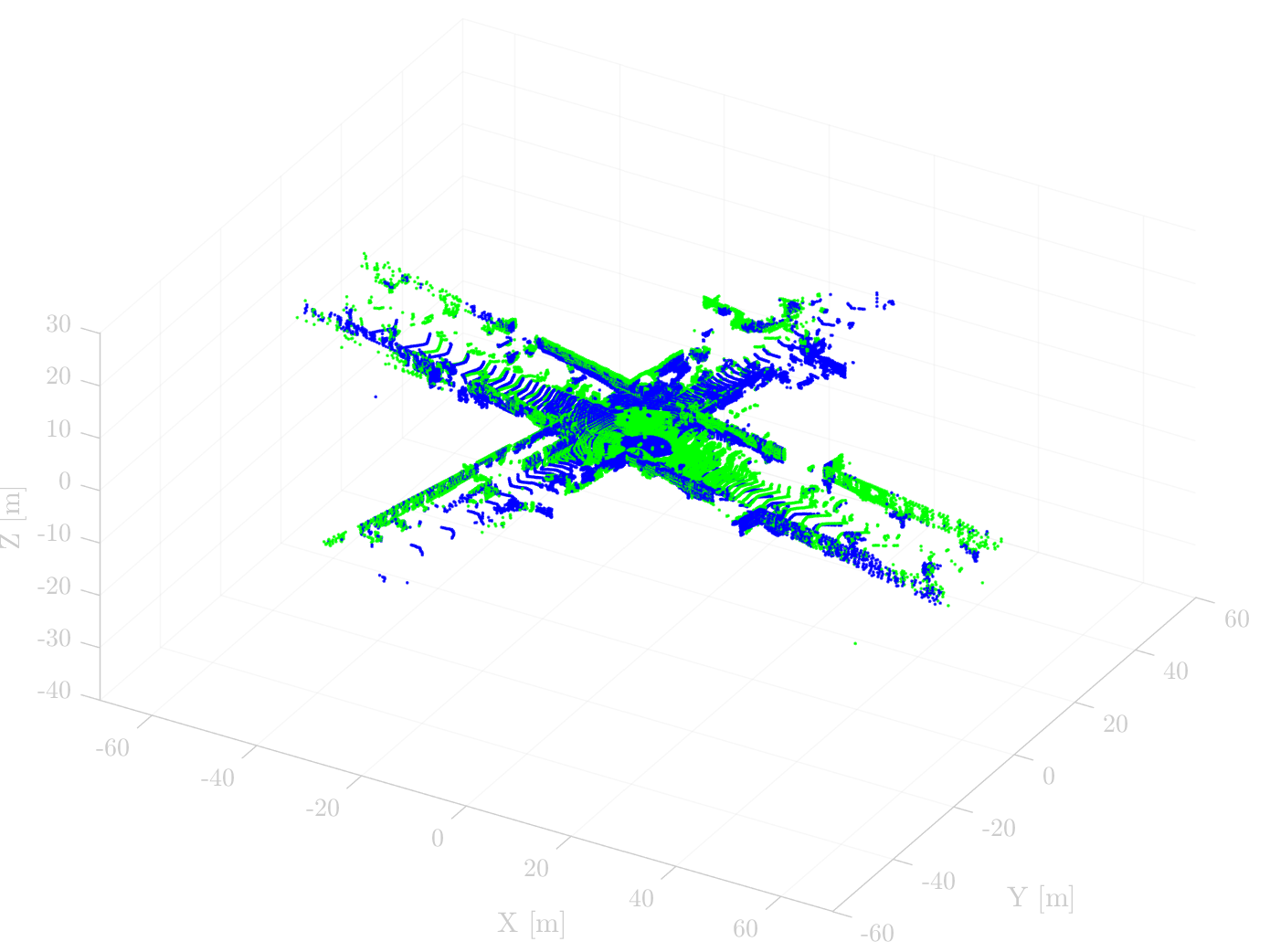}
		\includegraphics[width=1.0\textwidth]{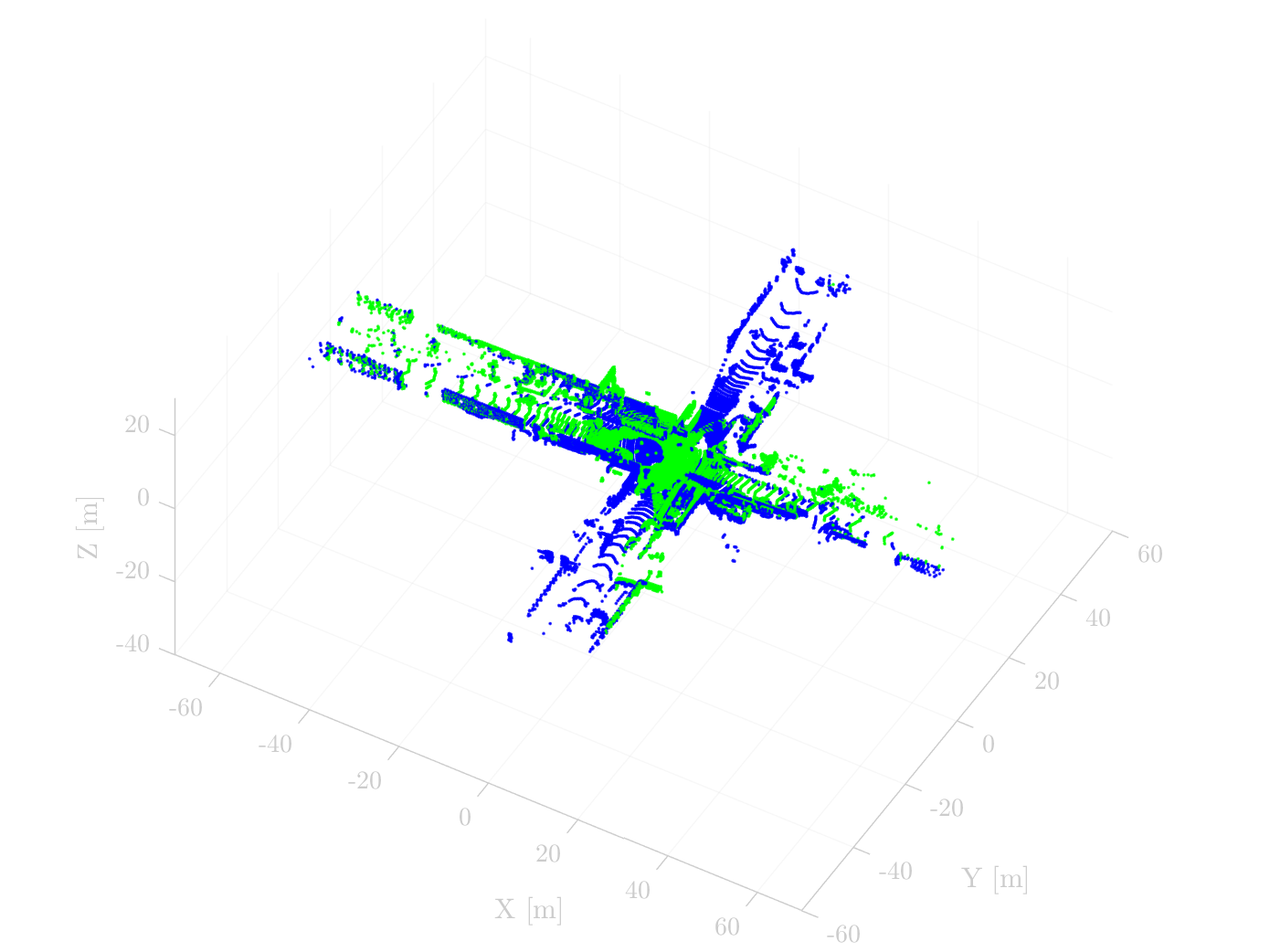}
		\includegraphics[width=1.0\textwidth]{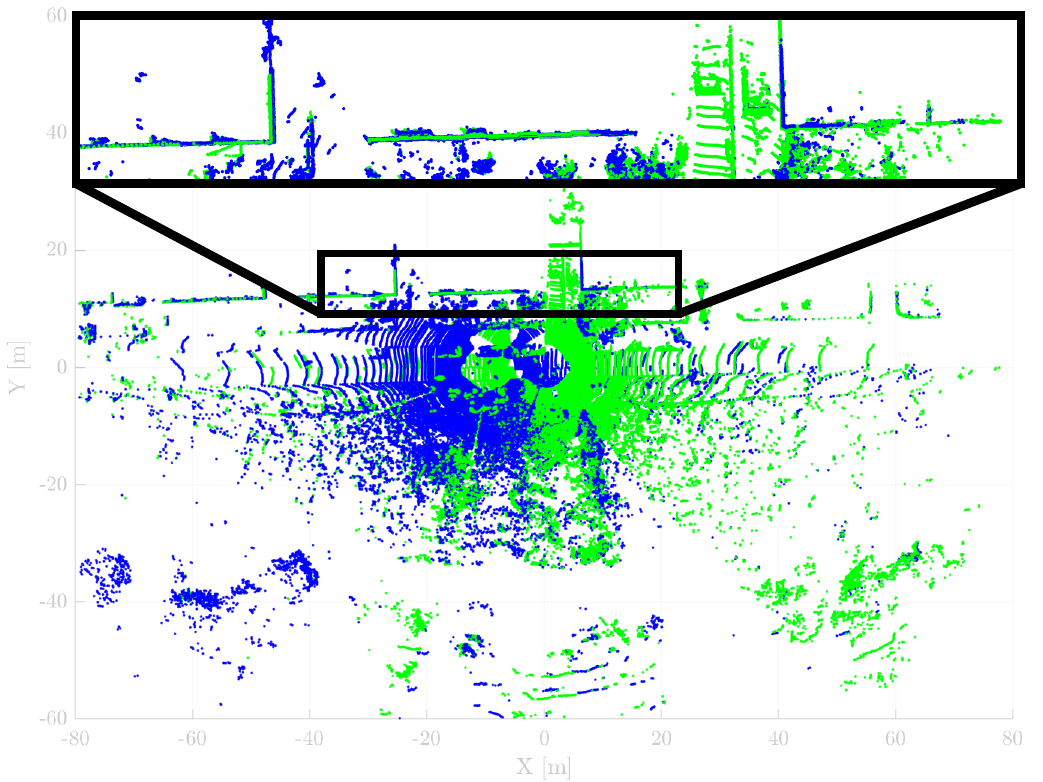}
		\caption{\centering Quatro++~(Ours)}
	\end{subfigure}
	\caption{Qualitative comparison in distant cases{\color{black}, where only our approach succeeded in performing registration of two partially-overlapped point clouds} in the KITTI dataset. Even though distant source~(red) and target~(green) clouds are given, our proposed method showed tightly aligned results, overcoming the sparsity and degeneracy issues. The closer the warped source cloud~(blue) and target cloud are, the better. The solid black boxes are zoomed views that highlight the misalignment of other state-of-the-art methods. The left-top texts represent the initial pose discrepancy in each scene. Note that the RANSAC returns the identity matrix, i.e. $\mathbf{I}_4$, in the failure cases, so the estimates of RASNAC are identical to the source clouds~(best viewed in color).}
	\label{fig:kitti_qualitative}
\end{figure*}

Furthermore, some cases where our Quatro++ only succeeded are presented in Fig.~\ref{fig:kitti_qualitative}, demonstrating the robustness of our proposed method against corridor-like environments~(the 1st$\sim$3rd rows,~Fig.~\ref{fig:kitti_qualitative}), three forked roads~(the 4th$\sim$5th rows,~Fig.~\ref{fig:kitti_qualitative}), and intersection or opposing viewpoint scenarios that have large angular discrepancies between the viewpoints of source and target clouds~(the 6th$\sim$8th rows,~Fig.~\ref{fig:kitti_qualitative}). 

The reasons why other state-of-the-art methods failed are presumed as follows. As mentioned earlier, as the pose discrepancy between two viewpoints becomes farther, the number of true inliers decreases and the ratio of outliers increases simultaneously. 
Under this circumstance, RANSAC is highly likely to fail to estimate correct relative pose because RANSAC is relatively weak to the outliers. 
In fact, it was reported that RANSAC often fails once the ratio of outliers is over 60\%~\citep{tzoumas2019outlier}.

FGR showed better performance compared with RANSAC because FGR was based on GNC. 
Unfortunately, FGR linearizes SE(3) during the optimization, yet this linearization is sometimes not valid once the pose discrepancy becomes larger, resulting in performance degradation~(8$\sim$ 10 m case, Table~\ref{table:kitti_IMU}).
In addition, FGR has no graph-based pruning module, i.e. MCIS, so that FGR cannot tolerate gross outliers.
 
On the other hand, TEASER++ showed a promising performance even in distant cases because TEASER++ also utilizes the graph-based outlier pruning method. 
However, the point is that MCIS also occasionally rejects too many correspondences by considering them as outliers, potentially leading to degeneracy in the GNC-based SO(3) rotation estimation step. As a result, the rotation estimation output flipped~(the 3rd, 5th, and 8th rows, Fig.~\ref{fig:kitti_qualitative}) or tilted results~(the other rows, Fig.~\ref{fig:kitti_qualitative}) owing to the degeneracy.
Note that TEASER++ is based on the decoupling of rotation and translation, so the failure of rotation estimation inevitably brings failure of translation.

\color{black}
Deep learning-based methods showed a substantial and inconsistent degradation in performance as the viewpoint distance between the source and target increased.
This is because these learning-based approaches were only trained to predict the relative pose in cases where the viewpoint difference between the source and target is sufficiently small and rarely trained in distant loop cases.
Furthermore, as presented in Table~\ref{table:success_rate_in_mulran}, the performance of deep learning-based methods was more significantly degraded than our approach when the surroundings or sensor configuration was different from the training data.
These results support our claim that our approach is more suitable for loop closing and applicable to various environments because our approach does not need any training procedure.
\color{black}

%%%%%%%%%%%%%%%%%%%%%%%%%
%%%%% Odometry test %%%%%
%%%%%%%%%%%%%%%%%%%%%%%%%
\noindent \textbf{In Odometry Test} Next, the registration methods were tested in the odometry test. Our proposed method is also compared with the conventional odometry methods and deep learning-based methods. 
Note that global registration basically does not aim to be used as an odometry, but we can check which method shows consistent performance by analyzing trajectory errors. 
In other words, better global registration shows better odometry performance once the frame interval $\Delta$ is larger than one, i.e. $\Delta = 3$ and $\Delta = 5$ in Table~\ref{table:kitti_odom}.

\color{black}
As baseline methods, we employ three local registration methods: ICP~\citep{besl1992method}, G-ICP~\citep{segal2009gicp}, and VGICP~\citep{koide2020vgicp}.
Furthermore, we compare our approach with learning-based odometry methods: LO-Net~\citep{li2019net}, DMLO~\citep{li2020dmlo}, and A-LOAM with StickyPillars~\citep{fischer2021stickypillars}.
Note that the codes of deep learning-based odometry approaches are not available, so we employ the results from the original papers.
We also employ conventional LiDAR odometry methods: SuMa~\citep{behley2018efficient} and A-LOAM~\citep{zhang2014loam}.

\color{black}

\begin{figure*}[t!]
	\centering
	\begin{subfigure}[b]{1\textwidth}
		\includegraphics[width=0.45\textwidth]{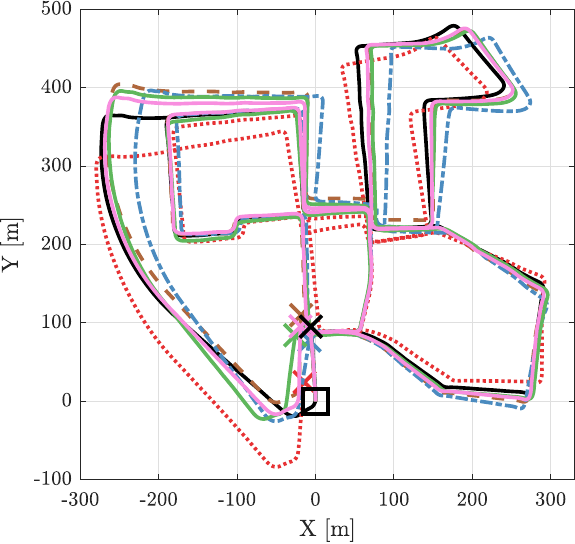}
		\includegraphics[width=0.45\textwidth]{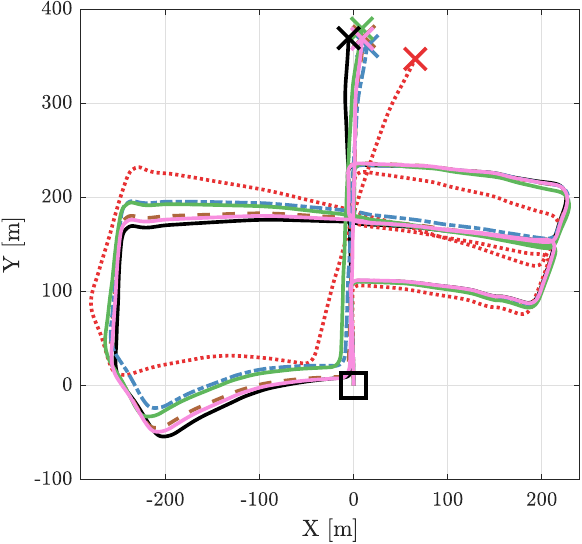}
		\includegraphics[width=0.45\textwidth]{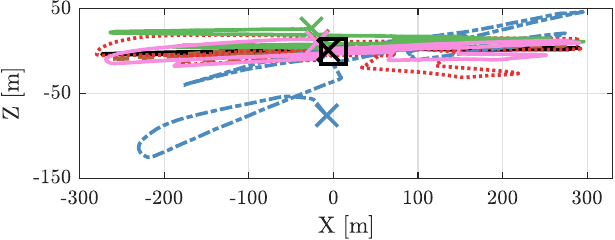}
		\includegraphics[width=0.45\textwidth]{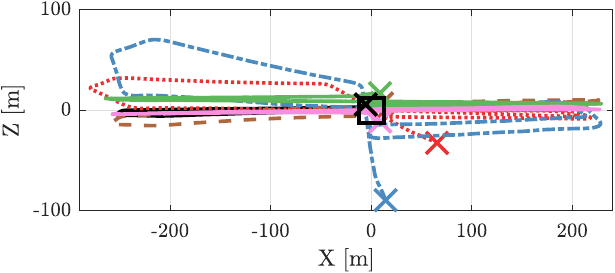}
		\centering
		\caption{\centering $\Delta =1$}
	\end{subfigure}
	\begin{subfigure}[b]{1\textwidth}
		\includegraphics[width=0.45\textwidth]{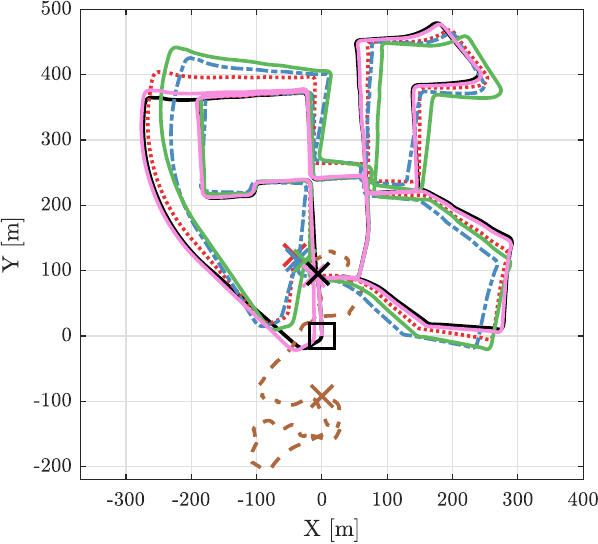}
		\includegraphics[width=0.45\textwidth]{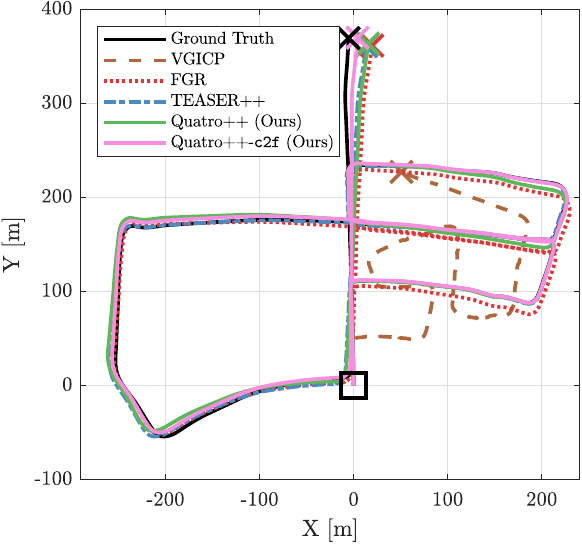}
		\includegraphics[width=0.45\textwidth]{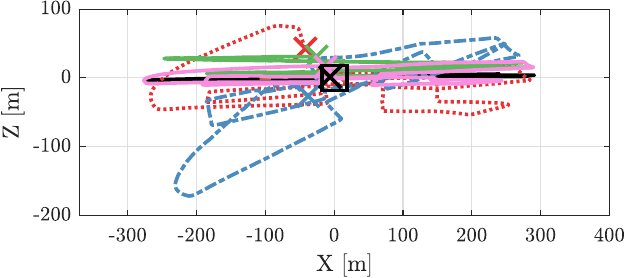}
		\includegraphics[width=0.45\textwidth]{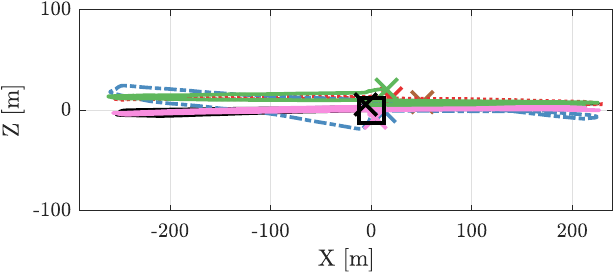}
		\centering
		\caption{\centering $\Delta = 5$}
	\end{subfigure}
	\caption{(L-R): Qualitative comparison of trajectories by the state-of-the-art registration methods in Seq.~\texttt{00} and Seq.~\texttt{05} of the KITTI dataset. $\Delta$ denotes the frame interval; for instance, $\Delta = \delta$ means that the $(i+\delta)$-th and $i$-th point clouds are taken as the source and target clouds, respectively. Our proposed methods showed more consistent results although the distant and partially overlapped two point clouds are given. In contrast, the trajectories of other global registration methods were likely to be twisted owing to the undesirable pitch and roll errors and some failure situations.}
	\label{fig:kitti_traj_00_05}
\end{figure*}

As shown in Table~\ref{table:kitti_odom}, local registration methods showed promising performance only if \color{black} the frame interval is not large, i.e. $\Delta = 1$. \color{black}
These results make sense because the pose discrepancy of two viewpoints is sufficiently within the coverage of the local registration, so their estimate can successfully converge into the global optimum. 
However, as the interval of frames \color{black} gets \color{black} larger, the performance of local registration is dramatically degraded because their strong assumption that the nearest point pairs between the source and target clouds are correlated, does not hold.
As a result, VGICP~\citep{koide2020vgicp}, which is one of the state-of-the-art local registration methods, showed catastrophic failure of odometry estimation, as shown in Fig.~\ref{fig:kitti_traj_00_05}(b). 

\begingroup
\begin{table}[t!]
	\captionsetup{font=footnotesize}
	\centering
	\caption{Comparison of odometry test with the state-of-the-art methods on Seq.~\texttt{00} of the KITTI dataset. $\Delta$ denotes the frame interval; for instance, $\Delta = \delta$ means that the $(i+\delta)$-th and $i$-th point clouds are taken as the source and target clouds, respectively. The suffix \texttt{c2f} means that the global registration is employed as an initial alignment followed by the local registration as a fine alignment. The results of deep learning-based methods are from \cite{li2020dmlo} and \cite{fischer2021stickypillars} (units for $t_\text{rel}$:~\%, $r_\text{rel}$:~$\deg$/100m).} 
	\setlength{\tabcolsep}{4pt}
	{\scriptsize
		\begin{tabular}{l|l|cccccc}
			\toprule \midrule
			&\multirow{2}[3]{*}{Method} & \multicolumn{2}{c}{$\Delta = 1$} & \multicolumn{2}{c}{$\Delta = 3$} & \multicolumn{2}{c}{$\Delta = 5$} \\  \cmidrule(lr){3-4} \cmidrule(lr){5-6} \cmidrule(lr){7-8} 
			&  & $t_{\text{rel}}$ & $r_{\text{rel}}$ & $t_{\text{rel}}$ & $r_{\text{rel}}$ & $t_{\text{rel}}$  & $r_{\text{rel}}$   \\ \midrule
			\parbox[t]{2mm}{\multirow{3}{*}{\rotatebox[origin=c]{90}{Local}}} &ICP  & 6.88 & 2.99 & 21.92 & 8.70 & 21.14 & 8.51 \\
			&G-ICP  & 1.26 & 0.45 & 5.50 & 1.45 & 14.20 & 3.32\\
			&VGICP  & \textbf{1.03} & \textbf{0.30} & 11.83 & 1.65 & 19.11 & 6.32  \\ \midrule
			\parbox[t]{2mm}{\multirow{4}{*}{\rotatebox[origin=c]{90}{Global}}}&%GH-ICP~\citep{dong2017gh-icp} & 23.27 & 9.99 & 22.54 & 10.32 & 21.11 & 10.16 \\
			FGR & 2.73 & 0.69 & 7.17 & 1.58 & 14.66 & 4.12  \\
			&TEASER++~  & 2.11 & 0.91 & 2.64 & 1.11  & 3.19 & 0.91 \\  
			&Quatro (Ours) & 1.45 & 0.41 & \textbf{1.38} & \textbf{0.24} & 1.94 & 0.46  \\
			&Quatro++ (Ours) & 1.90 & 0.53 & 1.45 & 0.32  & \textbf{0.99} & \textbf{0.28} \\
			\midrule \midrule
			\parbox[t]{2mm}{\multirow{5}{*}{\rotatebox[origin=c]{90}{Deep}}}&LO-Net & 1.47 & 0.72 & N/A & N/A & N/A & N/A  \\
			&LO-Net+\texttt{M} & 0.78 & 0.42 & N/A & N/A & N/A & N/A  \\
			&DMLO$^\dagger$  & 0.83 & 0.44 & N/A & N/A & N/A & N/A  \\
			&DMLO+\texttt{M}$^\dagger$  & 0.73 & 0.44 & N/A & N/A & N/A & N/A  \\
			&A-LOAM + StickyPillars$^{\dagger}$  & \textbf{0.65} & 0.26 & 0.79 & 0.31 & 1.29 & 0.48  \\\midrule
			\parbox[t]{2mm}{\multirow{4}{*}{\rotatebox[origin=c]{90}{Conv.}}}  &SuMa  & 0.68& 0.23 & 1.69 & 0.61 & 2.36 & 0.51 \\
			&A-LOAM  & 0.70 & 0.27 & 0.97 & 0.38 & 31.16 & 12.10  \\
			&Quatro-\texttt{c2f} (Ours) & \textbf{0.65} & \textbf{0.21} & 0.67 & \textbf{0.21} & 0.67 & 0.21  \\
			&Quatro++-\texttt{c2f} (Ours) & 0.68 & 0.23 & \textbf{0.61} & \textbf{0.21} & \textbf{0.50} & \textbf{0.20}  \\ \midrule\bottomrule
		\end{tabular}
	}
	\begin{flushleft}
	    \footnotesize{
		$\dagger$: Seq.~\texttt{00} is used for training of the network}
	\end{flushleft}
	\label{table:kitti_odom}
	\vspace{-0.3cm}
\end{table}
\endgroup

In terms of global registration, we stress two experimental observations. 
First, our Quatro and Quatro++ especially showed less decreased performance than the other global registration methods, which supports that forgoing the estimation of roll and pitch angles makes the algorithms more robust and consistent. 
That is, the trajectories of other global registration methods were likely to be twisted owing to undesirable pitch and roll errors, whereas our proposed method showed more consistent trajectories, as seen from the $xz$-viewpoint in Fig.~\ref{fig:kitti_traj_00_05}. 
Put differently, the trajectories from the other global registration methods showed the large discrepancy along the $z$-axis, whereas the result from our Quatro++ was closer to the ground truth. 

Second, using ground segmentation as a preprocessing step helps the global registration avoid local minima because Quatro++ showed a promising performance in the case where $\Delta=5$ in Table~\ref{table:kitti_odom}. 
Interestingly, when the intervals were small, Quatro showed lower errors than Quatro++. 
It can be interpreted that once the pose discrepancy between two viewpoints is not large, ground points rather help to estimate the relative pose because the ground points of source and target are mostly overlapped, leading to more true inliers. 
% However, Quatro++ shows more robust than Quatro when the frame interval is large in that Quatro++ provides much smaller errors when $\Delta5$. 
% This is because Quatro++ is less likely to fail when performing registration than Quatro, as explained in Sections~\ref{exp:quat_and_quatpp} and~\ref{exp:effect_of_ground_segmentation}. 

However, these errors do not directly affect the performance of fine alignment and thus are negligible for the local registration in $\Delta=1$ and $\Delta=3$ cases.
That is, there is an insignificant performance difference when Quatro or Quatro++ is utilized as an initial alignment when $\Delta = 1$ and $\Delta = 3$, showing similar level of errors in Table~\ref{table:kitti_odom}. 
Moreover, Quatro++ still robustly outputs the estimate to transform the viewpoint of the source cloud into the boundary within the narrow convergence region of the local registration.
Consequently, Quatro++-\texttt{c2f} showed a promising performance when $\Delta = 5$. 
Furthermore, Quatro++-\texttt{c2f} showed better performance compared with the state-of-the-art methods, even including conventional and deep learning-based approaches. It was remarkable that though some deep learning-based methods were trained using whole KITTI sequences including Seq.~\texttt{00}, our proposed method showed a competing performance without any prior knowledge.

Therefore, the suitability of our Quatro++ was demonstrated from the perspective of the coarse-to-fine alignment, which is our ultimate goal to achieve successful loop closing, helping local registration algorithms perform the fine alignment.

%%%%%%%%%$$$$$$%%%%%%%%%%%%%%%
%%%%% Augmented Rotation %%%%%
%%%%%%%%%%%%%%%%$$$$$%%%%%%%%%
\noindent \textbf{In Augmented Rotation Situations} Next, the performance was tested in augmented rotation situations in NAVER LABS localization dataset. The augmented rotation situations mean that the source cloud is additionally rotated along the yaw direction. By doing so, the feasibility of the algorithms on the opposing viewpoint or intersection cases could be checked.

\begin{figure}[t!]
	\captionsetup{font=footnotesize}
	\centering 
	\begin{subfigure}[b]{0.46\textwidth}
		\includegraphics[width=1.0\textwidth]{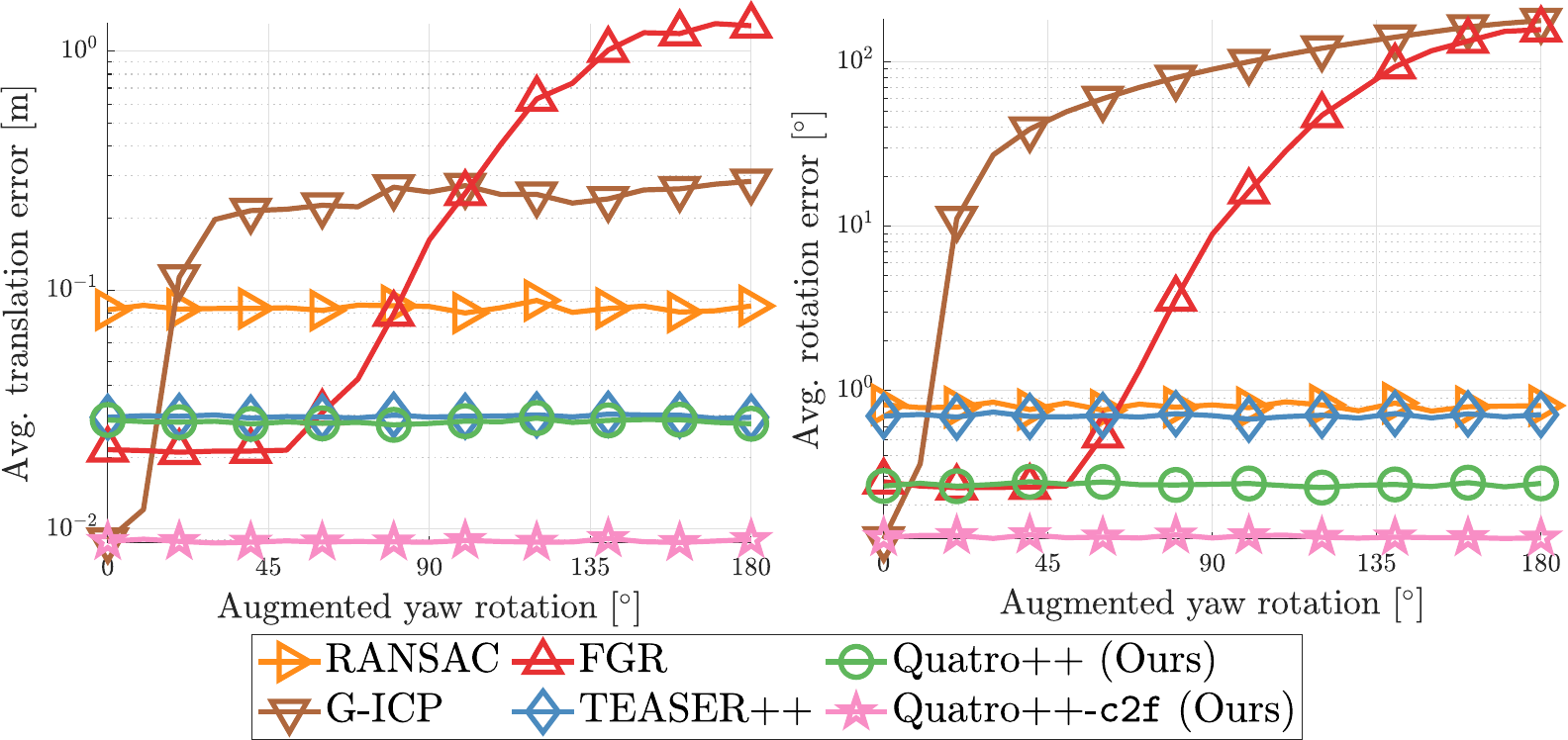}
	\end{subfigure}
	\caption{Performance changes with respect to the augmented yaw rotation when the frame interval is set to 5 (i.e. $\Delta = 5$) in \texttt{HD\_B1} of the \textcolor{qw}{NAVER LABS localization} dataset~(best viewed in color).} % Note that \textcolor{qw}{NAVER LABS localization} dataset was captured in indoor environments, so the ground is mostly flat; thus, the performance of Quatro and Quatro++ have no big difference~(best viewed in color).}
	\label{fig:ablation_wrt_range_diff}
\end{figure}

As shown in Fig.~\ref{fig:ablation_wrt_range_diff}, our proposed method showed lower errors and maintained the level of errors even though the augmented angle reached 180$^\circ$. 
FGR showed better performance if the augmented angle was not large; however, as the augmented angle became larger, the performance of FGR was degraded because the linearization assumption did not hold. 
%Note that NAVER LABS localization dataset was captured in indoor environments where the ground is mostly flat and not dominant than the KITTI dataset. Thus, the performance of Quatro and Quatro++ have no big difference. 

\begin{figure}[t!]
	\captionsetup{font=footnotesize}
	\centering 
	\begin{subfigure}[b]{0.23\textwidth}
		\includegraphics[width=1.0\textwidth]{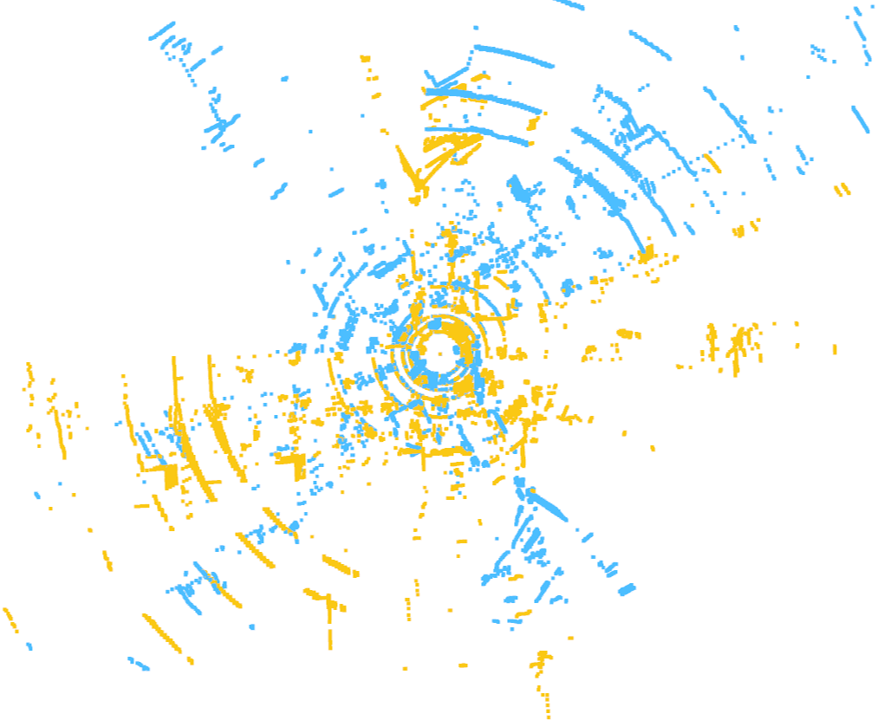}
		\caption{\centering Source and target}
	\end{subfigure}
	\begin{subfigure}[b]{0.23\textwidth}
		\includegraphics[width=1.0\textwidth]{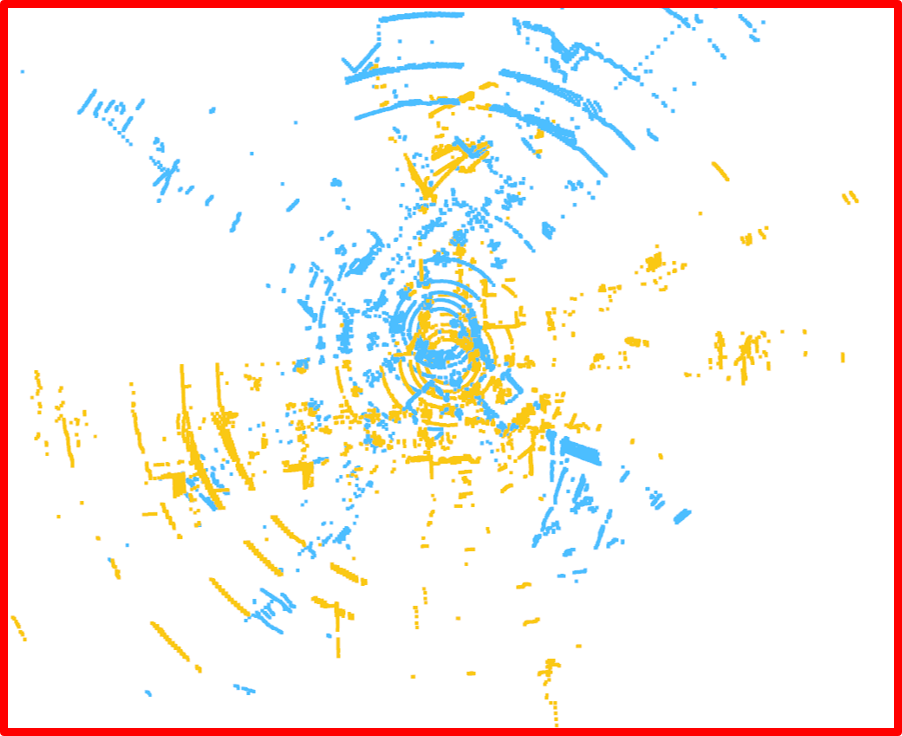}
		\caption{\centering FGR}
	\end{subfigure}
	\begin{subfigure}[b]{0.23\textwidth}
		\includegraphics[width=1.0\textwidth]{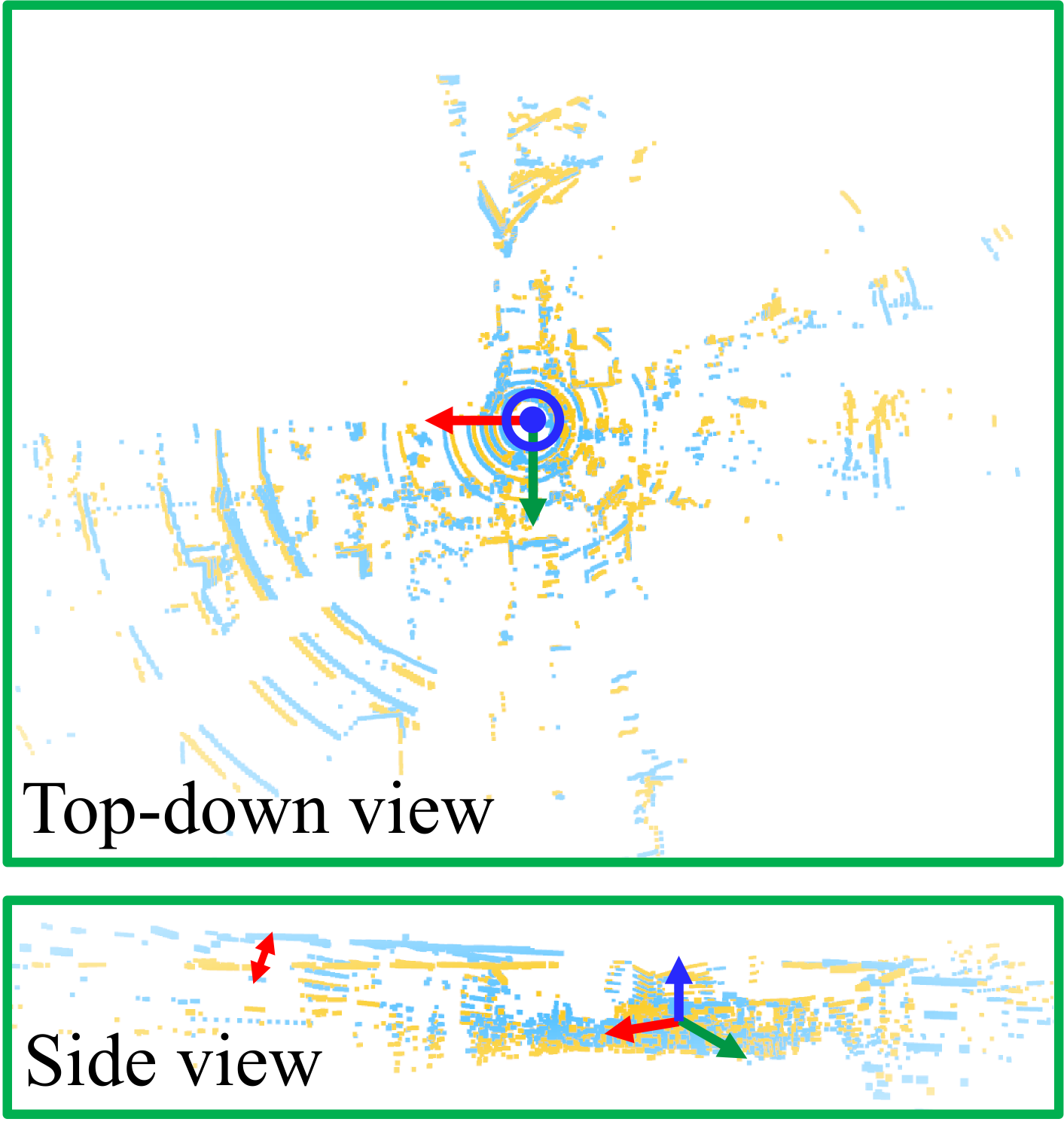}
		\caption{\centering TEASER++}
	\end{subfigure}
	\begin{subfigure}[b]{0.23\textwidth}
		\includegraphics[width=1.0\textwidth]{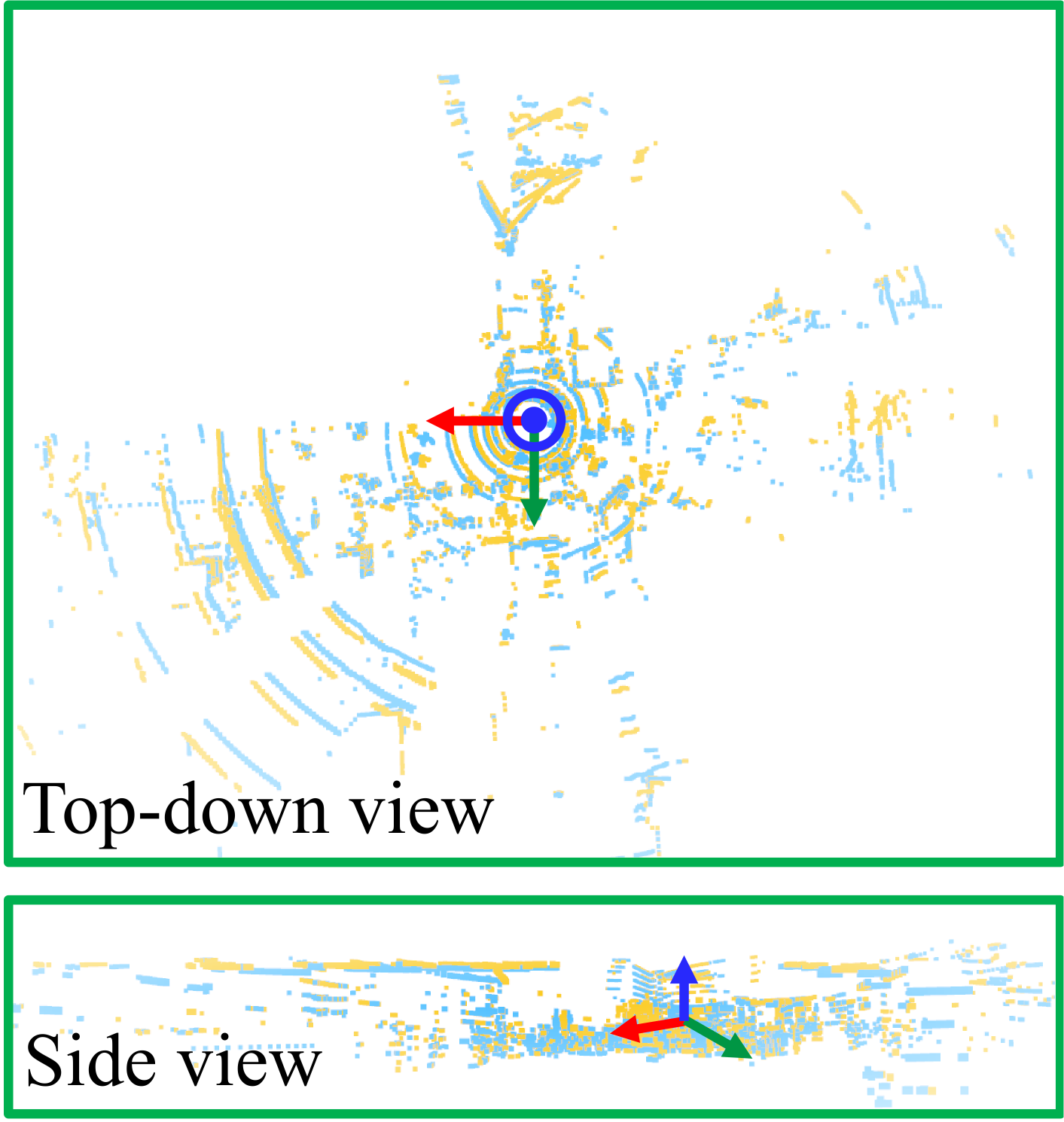}
		\caption{\centering Quatro++~(Ours)}
	\end{subfigure}
	\caption{Example of registration when large rotation discrepancy exists in \texttt{HD\_B1} of the NAVER LABS localization dataset. (c)~TEASER++ showed successful and robust registration, yet it yielded the undesirable angular error, highlighted as the red double arrow in the side view. The solid red and green boxes indicate the algorithms failed and succeeded, respectively (best viewed in color).}
	\label{fig:lab_agument_rot}
\end{figure}

TEASER++ showed successful and robust registration results, yet it yielded undesirable rotation errors, showing the offset between our proposed method and TEASER++ in Fig.~\ref{fig:ablation_wrt_range_diff}. 
In other words, the estimate by TEASER++ includes unintended roll and pitch errors. 
In contrast, our Quatro++ showed rather precise performance because our assumption that $\angle (\mathbf{R}_y \cdot \mathbf{R}_x)$ can be approximated as the identity matrix is met in most indoor environments.
In doing so, our quasi-SO(3) estimation prevents these roll and pitch errors, as shown in Fig.~\ref{fig:lab_agument_rot}.

Therefore, it was proven that our proposed method is also advantageous in indoor situations where the ground is mostly flat and the non-ground objects are mostly orthogonal to the ground.

\subsection{Computational Cost of Modules in Quatro++}

\begin{figure}[t!]
	\captionsetup{font=footnotesize}
	\centering 
	\begin{subfigure}[b]{0.5\textwidth}
		\includegraphics[width=1.0\textwidth]{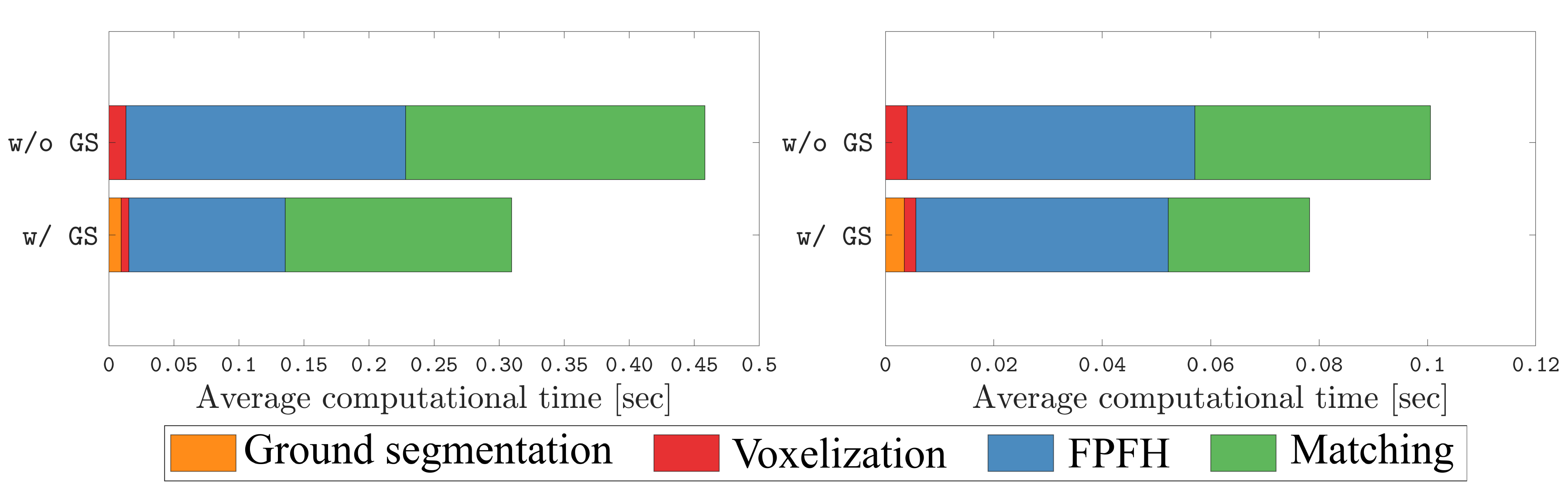}
	\end{subfigure}
	\caption{(L-R): The average time for each module in the preprocessing and correspondence estimation on Intel(R) Core(TM) i7-7700K and Intel(R) Core(TM) i9-13900. Note that the application of ground segmentation effectively reduces the time taken to set correspondences.}
	\label{fig:optim_time_feature}
\end{figure}

\begin{figure}[t!]
	\captionsetup{font=footnotesize}
	\centering 
	\begin{subfigure}[b]{0.50\textwidth}
		\includegraphics[width=1.0\textwidth]{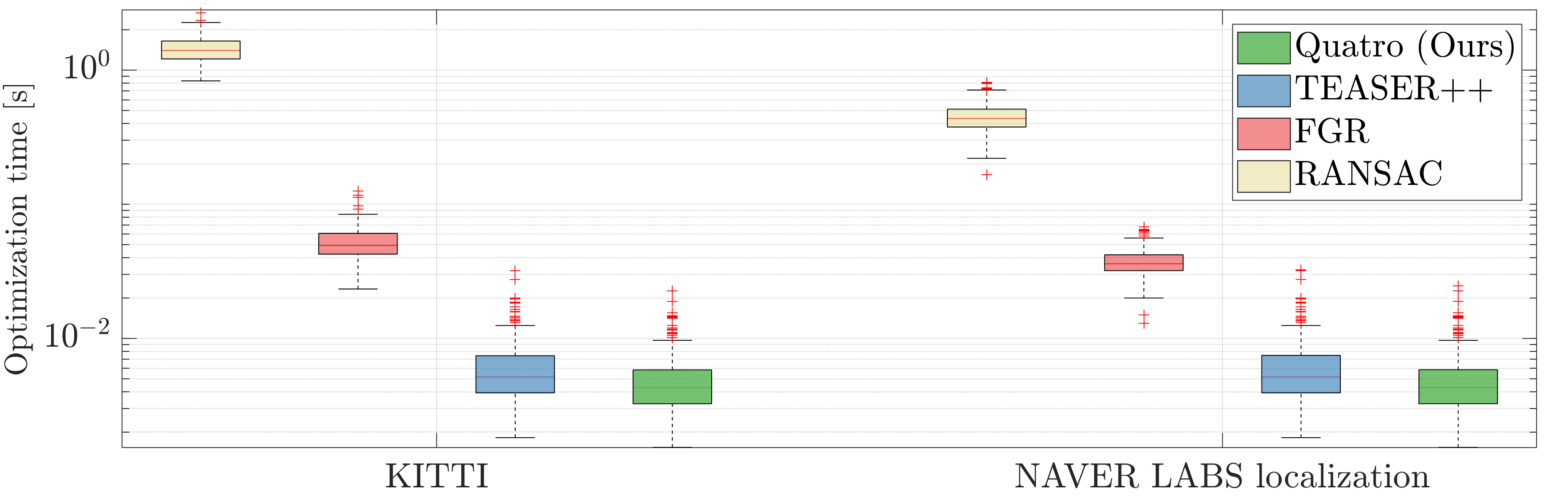}
	\end{subfigure}
	\caption{Average optimization time in the KITTI dataset and the \textcolor{qw}{NAVER LABS localization} dataset on Intel(R) Core(TM) i9-9900KF (Avg. of our Quatro: 5.0 and 6.4 ms, respectively).} % which sfollows Atlanta world assumption more strictly.}
	\label{fig:optim_time_backend}
\end{figure}

Here, the computational costs with respect to the feature extraction/matching and optimization of our global registration were analyzed. 
As shown in Fig.~\ref{fig:optim_time_feature}, the application of ground segmentation effectively reduced the time taken to set correspondences.
In addition, as shown in Fig.~\ref{fig:optim_time_backend}, the optimization part in our proposed method showed fastest speed compared with other methods.  

Therefore, the total time of Quatro++, which is the summation of the time taken for preprocessing, correspondence estimation, and Quatro, takes less than one second, so \color{black} we confirm \color{black} that our proposed method can be used in the loop closing module of SLAM frameworks, which requires over 1 Hz.

% correspondences의 크기에 따라 걸리는 시간
% 전체 시간
% Furthermore, our proposed method shows the fastest optimization time by virtue of MCIS-\texttt{heuristic}, as represented in Fig.~\ref{fig:optim_time}. On average, our method only takes 5~msec per optimization, which is sufficient for using our proposed method as a coarse alignment in real-time. 

%%%%%%%%%%%%%%%%%%%%%%%%%%%%%%%%%%%%%%%%%%%%%%%%%%%%%%%%%%%%%%%%%%%%
%                          Application
%%%%%%%%%%%%%%%%%%%%%%%%%%%%%%%%%%%%%%%%%%%%%%%%%%%%%%%%%%%%%%%%%%%%
\subsection{Application~\rom{1}: Leveraging an INS in Non-Flat Regions}\label{sec:exp_ins}

Our Quatro++ (or Quatro) sacrifices roll and pitch angles estimation to be more robust. However, Quatro++ and Quatro can achieve more precise initial alignment results by utilizing the INS measurements to compensate roll and pitch angles.
This compensation means that the approximated $\mathbf{R}_y \cdot \mathbf{R}_x$ is replaced by the rotation from the roll and pitch angles measured by INS, which is explained in Section~\ref{sec:ins}.
As presented in Table~\ref{table:kitti_IMU}, even though the raw measurements by the INS system were used, the pitch and roll compensation using these measurements significantly reduced the rotation errors, which is followed by the improvement of translation estimation. 
This is because our proposed method is based on the decoupling method, which estimates the relative rotation followed by translation estimation; 
thus, the quality of rotation estimation directly affects the quality of translation estimation.

\begin{figure}[t!]
	\captionsetup{font=footnotesize}
	\centering 
	\begin{subfigure}[b]{0.23\textwidth}
		\includegraphics[width=1.0\textwidth]{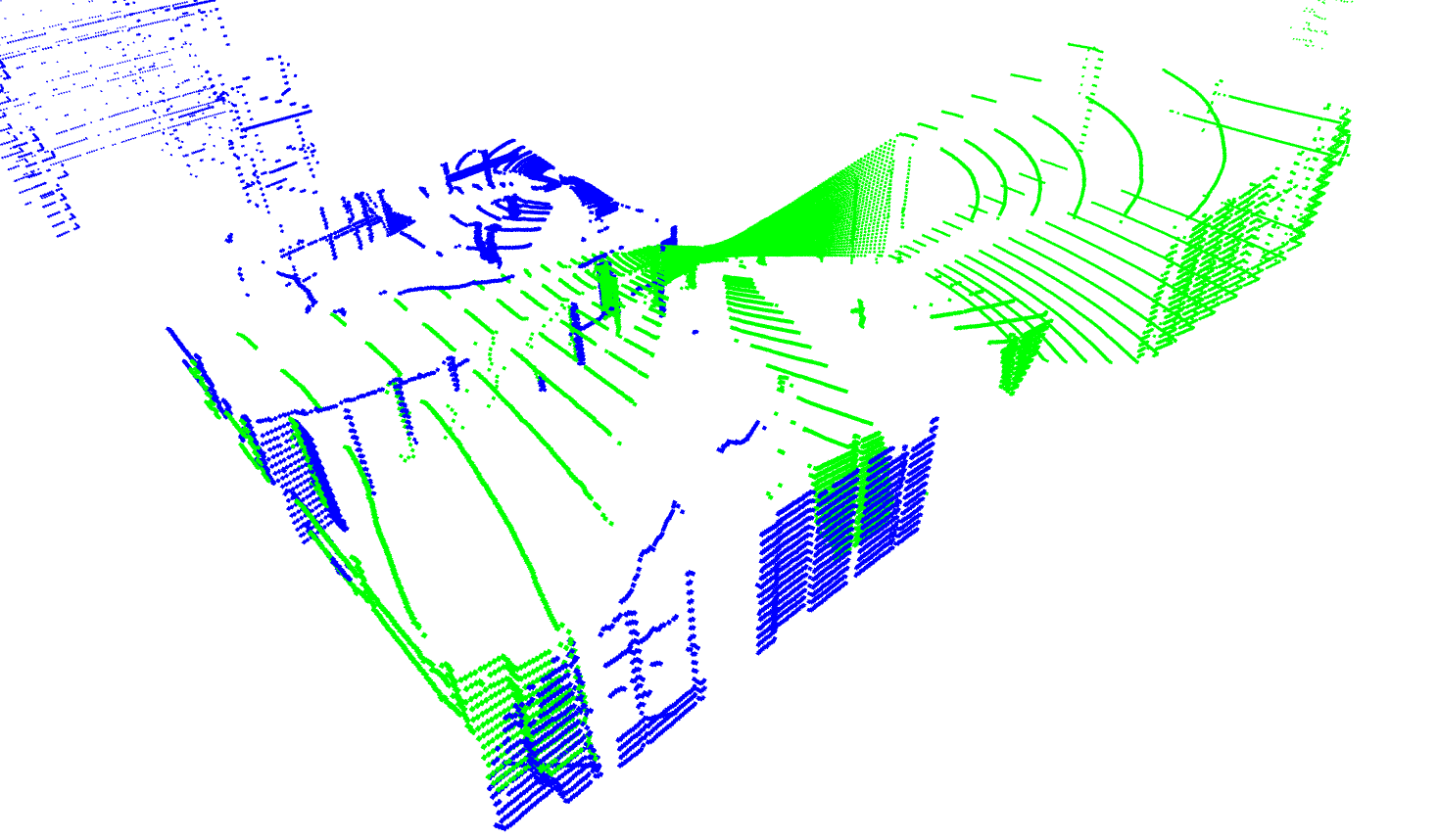}
	\end{subfigure}
	\begin{subfigure}[b]{0.23\textwidth}
		\includegraphics[width=1.0\textwidth]{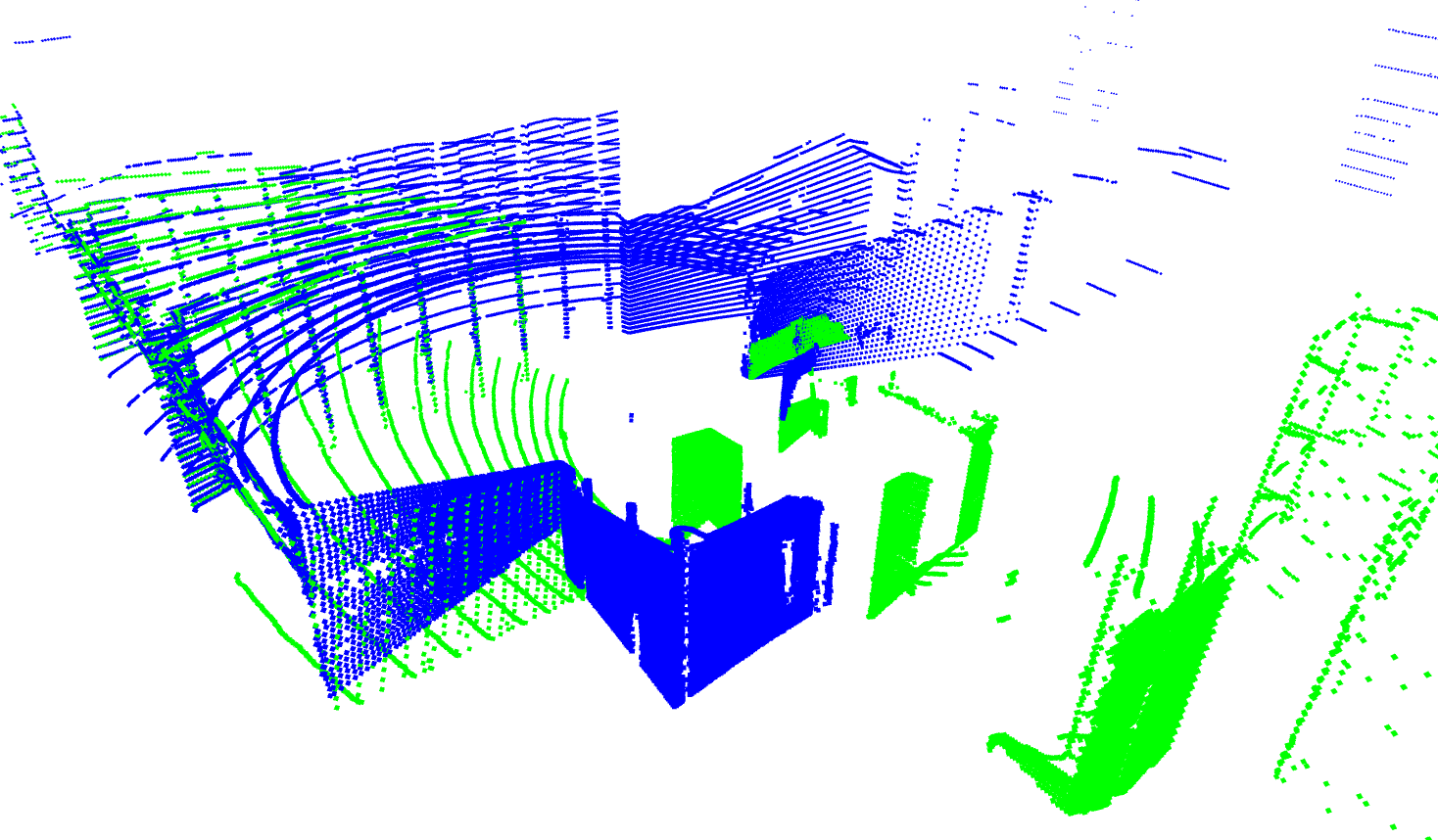}
	\end{subfigure}
	\caption{Registration results in a hand-held system in Oxford-Hilti dataset. Once the yaw and pitch angles are compensated by an INS system, our proposed method is also effective on a hand-held system. The blue and green points denote the warped source and target clouds, respectively (best viewed in color).}
	\label{fig:hilti}
\end{figure}

\begin{figure*}[t!]
	\captionsetup{font=footnotesize}
	\centering 
	\begin{subfigure}[b]{0.42\textwidth}
		\includegraphics[width=1.0\textwidth]{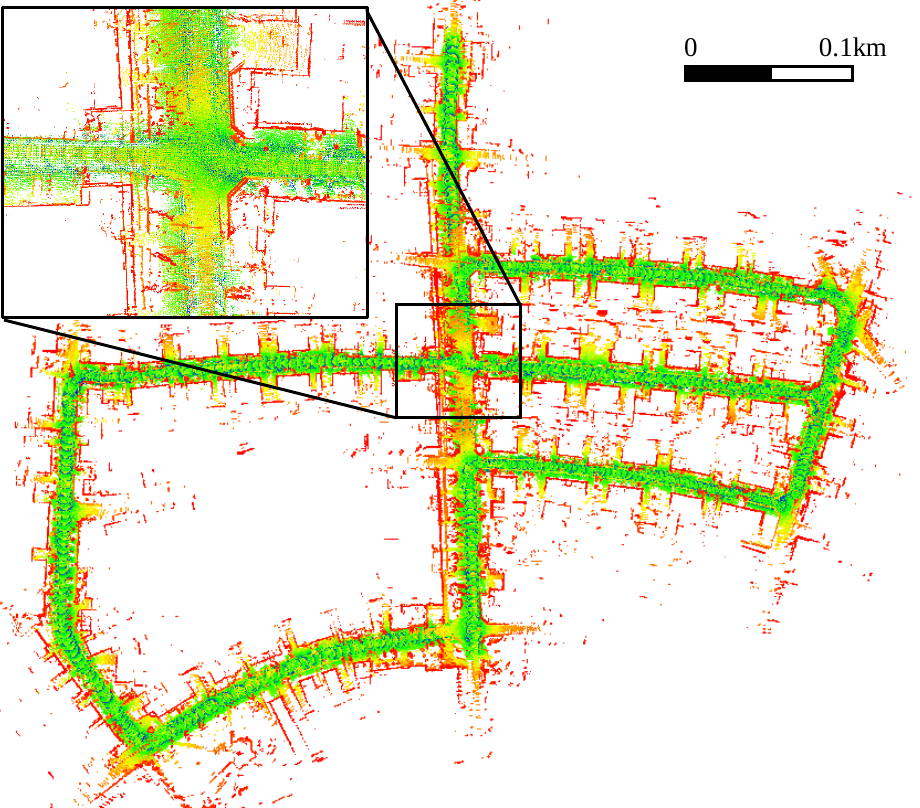}
		\caption{\centering}
	\end{subfigure}
	\begin{subfigure}[b]{0.42\textwidth}
		\includegraphics[width=1.0\textwidth]{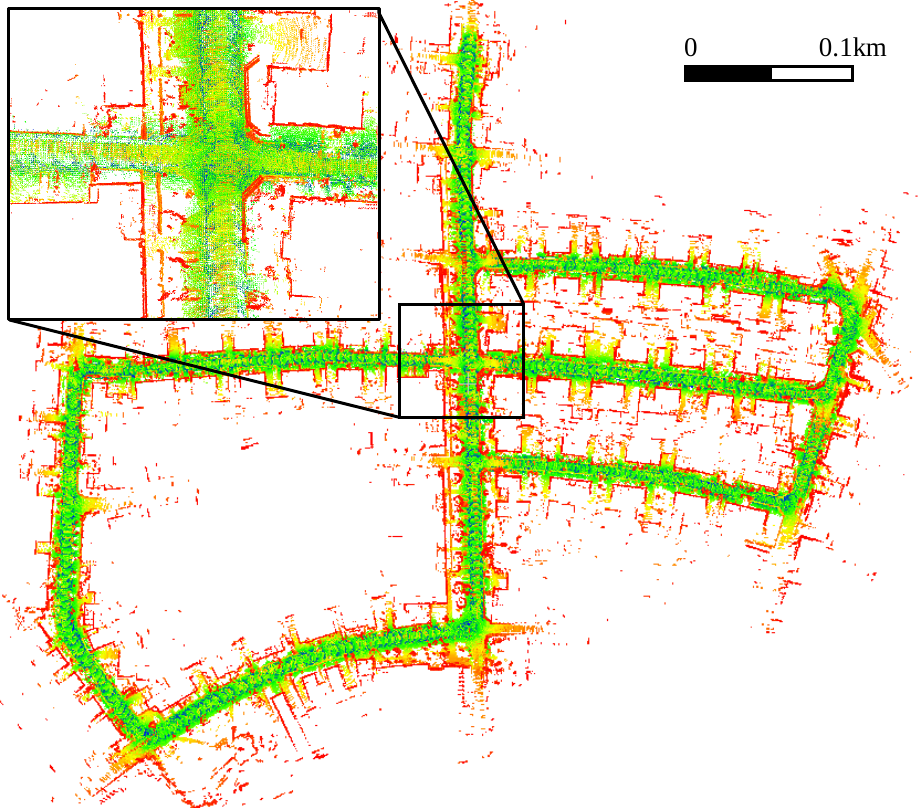}
		\caption{\centering}
	\end{subfigure}
	\caption{(a)-(b) Mapping results of LeGO-LOAM without and with the application of our proposed method as a loop closing in Seq.~\texttt{05} of the KITTI dataset. Note that all the parameters of LeGO-LOAM are set to be the same. }
	\label{fig:LeGO_Q_LeGO}
\end{figure*}

Not only for the improvement in performance, but we also stress that our proposed method can easily utilize these INS measurements without any additional fusion technologies, such as weighted average; thus, our proposed method is more INS-friendly than other methods.
One may argue that other methods can also improve performance by using INS measurements. However, other methods require an additional manner to appropriately average the estimated rotation and the rotation measured by the INS. 
Even though the weighted average is performed, this sensor fusion technique does not let the failed estimates successfully go into the narrow convergence region, still showing large pose errors. 
In contrast, as mentioned in Section~\ref{sec:ins}, the INS measurements can be easily exploitable for our proposed method by substituting the approximated roll and pitch rotations with those from INS measurements. 
By doing so, our proposed method showed lower errors while preserving higher success rates.

In addition, we conducted a feasibility study on a dataset made by a hand-held system. 
As shown in Fig.~\ref{fig:hilti}, even though the axes were rotated, estimation of $\hat{\mathbf{R}}_\text{INS}$, which was explained in Section~\ref{sec:ins}, followed by our proposed method showed successful registration results.
Therefore, it was demonstrated that our proposed method is also effective on a hand-held system once roll and pitch angles are compensated as an initial guess of our global registration. 
% sThat is, once the roll and pitch angles are compensated by the estimate of an INS system beforehand, our proposed method is still available.

 Therefore, these results demonstrated the feasibility of our proposed method in INS systems where the attitude of $xy$-plane of the sensor frame is not likely fixed\footnote{Consequently, we achieved the 4th and 1st place by utilizing our proposed method as a loop closing module in the 2022 and 2023 Hilti SLAM challenges, respectively. Please refer to the following site: \\ \href{https://hilti-challenge.com/leader-board-2023.html}{\texttt{https://hilti-challenge.com/leader-board-2023.html}}}.
 
 %%% Daebeom %%%%%%%%%%%%%%%%%%%%%%%%%%%%%%%%%%%%
 
 \begin{figure*}[t!]
 	\captionsetup{font=footnotesize}
 	\centering 
 	\begin{subfigure}[b]{0.83\textwidth}
 		\includegraphics[width=1.0\textwidth]{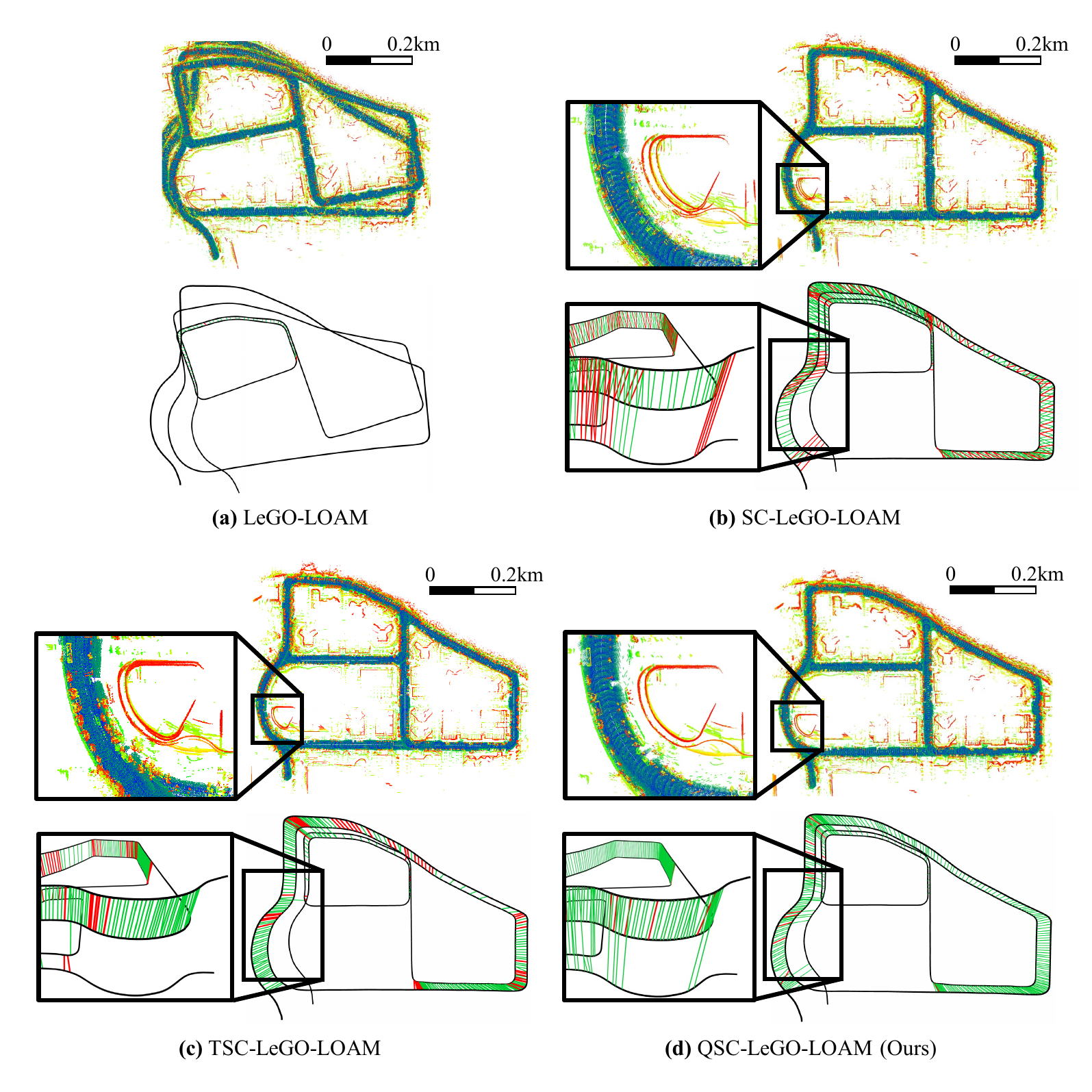}
 	\end{subfigure}
 	\caption{(T-B): Point cloud mapping results and the corresponding loop constraints in \texttt{DCC01} of the MulRan dataset. (a)~The original LeGO-LOAM. (b)~LeGO-LOAM with ScanContext as a loop detection. \color{black}(c) LeGO-LOAM with both TEASER++ and ScanContext, named as TSC-LeGO-LOAM. \color{black} (d)~LeGO-LOAM with both Quatro++ and ScanContext, \color{black} named as QSC-LeGO-LOAM, showing a tightly aligned mapping result and fewer false loops. The green lines denote the true positive loop constraints and red lines denote the false positive loop constraints \color{black} (best viewed in color).}
 	\label{fig:dcc_map}
 \end{figure*}
 
 \begin{figure*}[t!]
 	\captionsetup{font=footnotesize}
 	\centering 
 	\begin{subfigure}[b]{0.85\textwidth}
 		\includegraphics[width=1.0\textwidth]{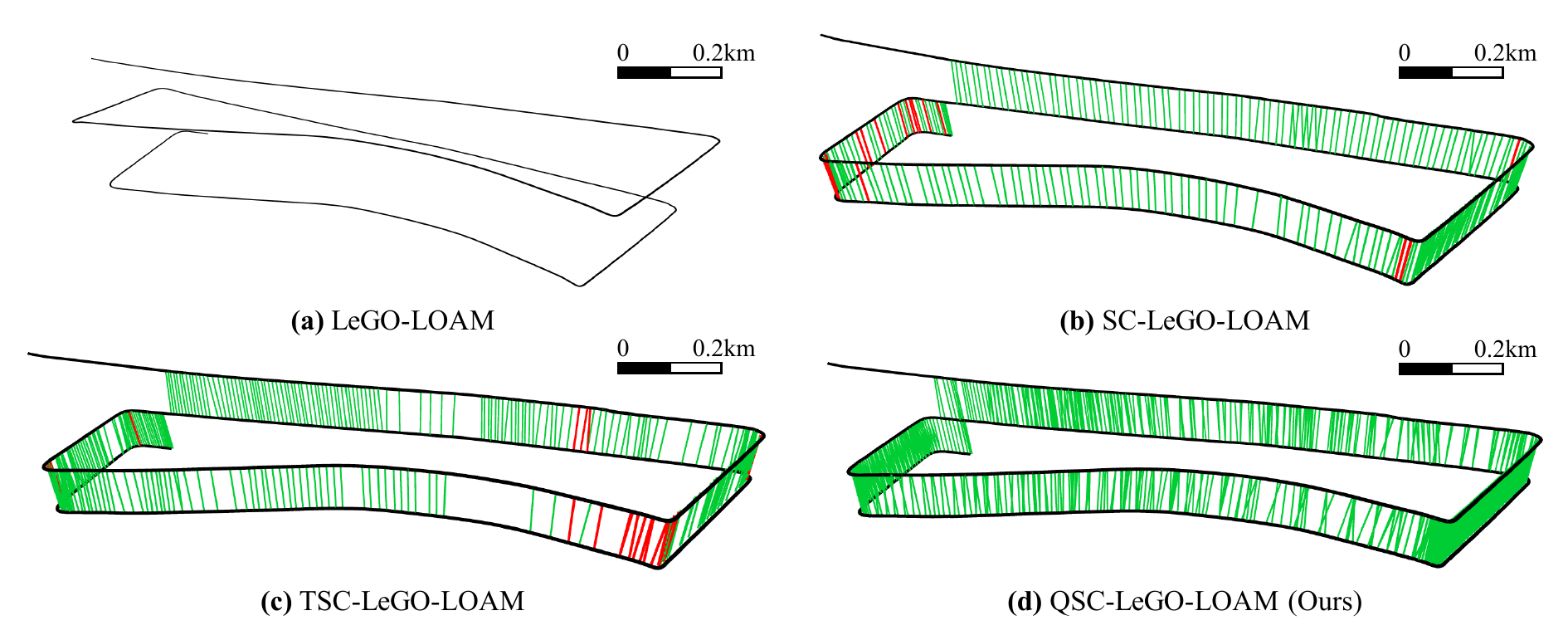}
 	\end{subfigure}
 	\caption{Visualized loop constraints of \color{black} (a) The original LeGO-LOAM, (b) LeGO-LOAM with ScanContext as a loop detection and (c) TSC-LeGO-LOAM, and (d) QSC-LeGO-LOAM in \texttt{Riverside01} of the MulRan dataset. \color{black} Note that QSC-LeGO-LOAM reduced false positive loops while increasing true positive loops simultaneously. \color{black}The green lines denote the true positive loop constraints and red lines denote the false positive loop constraints \color{black} (best viewed in color).}
 	\label{fig:riverside_loops}
 \end{figure*}

 \begin{figure}[t!]
 	\captionsetup{font=footnotesize}
 	\centering 
 	\begin{subfigure}[b]{0.49\textwidth}
 		\includegraphics[width=1.0\textwidth]{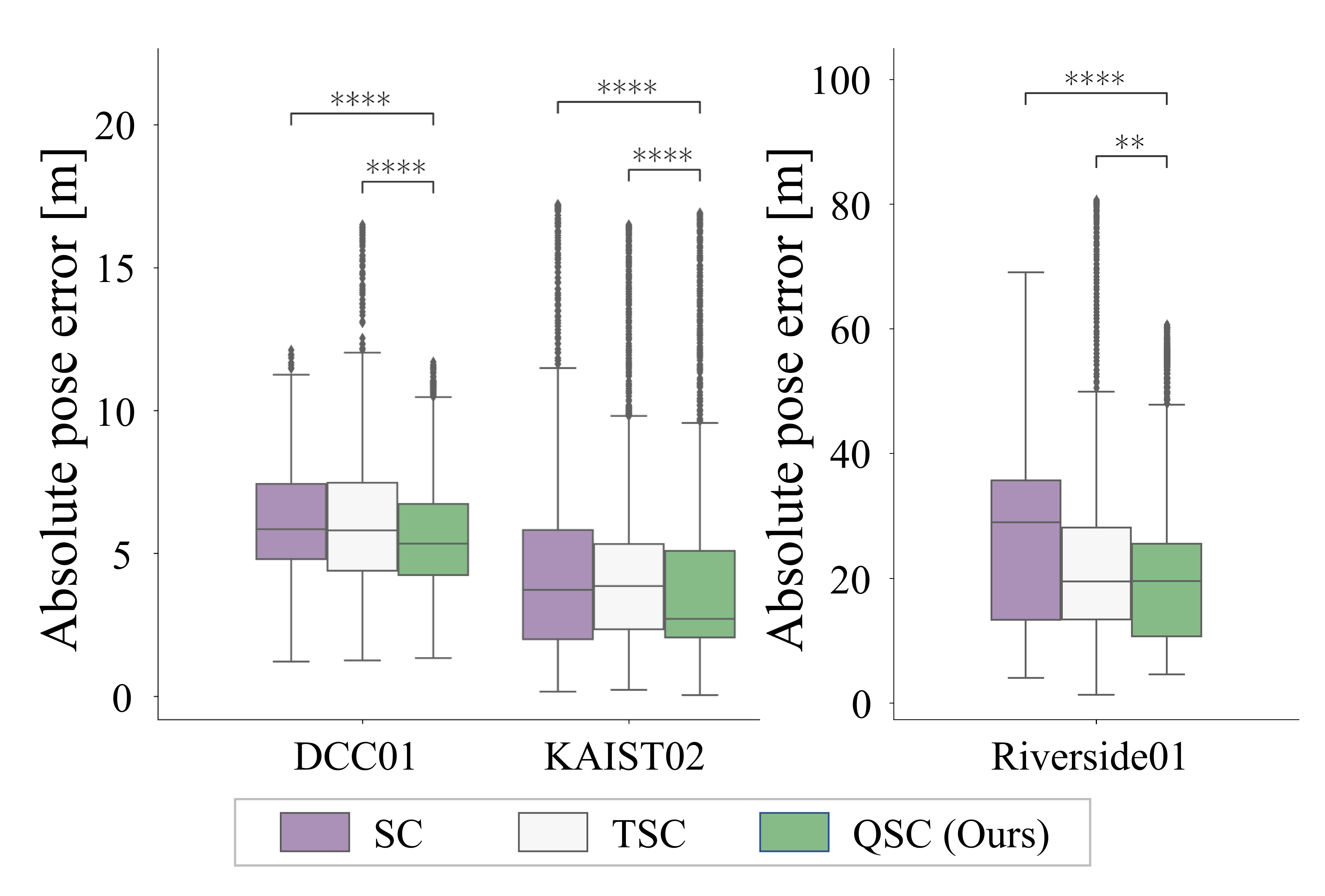}
 	\end{subfigure}
 	\caption{Box plot of absolute pose errors after the application of \color{black} ScanContext~(SC) as a loop detection, both TEASER++ and ScanContext~(TSC), and both our Quatro++ and ScanContext~(QSC). \color{black} The $\ast \! \ast \! \ast \ast$ annotations indicate measurements with $p\text{-value} < 10^{-4}$ after a paired \textit{t}-Test.
 	}
 	\label{fig:Barplot_ape_rpe}
 \end{figure}

 %%%%%%%%%%%%%%%%%%%%%%%%%%%%%%%%%%%%%%%
 
\subsection{Application~\rom{2}: Quatro++ in LiDAR SLAM Frameworks}\label{sec:exp_slam}

So far, we investigated performance of a single registration given a pair of point clouds. 
Here, the impact of Quatro++ to the LiDAR SLAM was analyzed. 
Note that our Quatro++ can be employed in keyframe-based pose-graph SLAM systems.

% false negative
First, we compared the performance before and after the application of our global registration into LeGO-LOAM~\citep{shan2018lego}. 
LeGO-LOAM utilizes the radius search to find loop candidates by using the current position as a query, which is followed by ICP. 
However, as mentioned in Section~\ref{sec:limit_loop_closing}, ICP usually fails to estimate the correct relative pose in distant cases, rejecting many loop candidates.
That is, even though the radius search finds the actual loop, the loop candidate whose pose discrepancy is large is rejected owing to the narrow convergence region of the local registration methods, resulting in a high MSE. 
Consequently, these loops are not included in the graph structure, leaving trajectory errors~(Fig.~\ref{fig:LeGO_Q_LeGO}(a)).

In contrast, LeGO-LOAM with our proposed method showed a more precise mapping result~(Fig.~\ref{fig:LeGO_Q_LeGO}(b)).
It was observed that the number of false positive rejections can be significantly reduced and the quality of loop constraints is improved when our Quatro++ provides an initial alignment to the local registration.
As a result, even if a large drift is accumulated through the large loop, the global trajectory errors can be successfully minimized by leveraging more accurate and abundant loop constraints.
 
% 이러한 현상은 loop detection 모듈과 함께 썼을 때에도 사용 가능! 
The positive impact on the quality of mapping becomes more distinguishable and significant in large-scale environments.
In this experiment, three types of LeGO-LOAM were compared: the original LeGO-LOAM, {SC-LeGO-LOAM} that exploits ScanContext~\citep{kim2018scancontext} as a loop detection, \color{black} {TSC-LeGO-LOAM} that employs both TEASER++ and ScanContext, \color{black} and {QSC-LeGO-LOAM} that utilizes both our proposed method (Quatro++) and ScanContext.

As shown in Fig.~\ref{fig:dcc_map}, \color{black} LeGO-LOAM without any loop detection and closing approaches \color{black} showed imprecise mapping results.
In large-scale environments, the radius search could not find potential loop candidates owing to the large pose drift. 
For this reason, loop closing was performed only locally, so its global trajectory errors were not reduced by PGO~(Fig.~\ref{fig:dcc_map}(a)).

The application of the loop detection module, i.e. SC-LeGO-LOAM, showed a more precise map by finding additional loop candidates that could not be searched solely by the radius search.
However, some trajectory errors still remained due to the undesirable false positive loops.
This is because the loop detection sometimes wrongly estimates two geometrically similar scenes as a loop candidate.
For instance, the scenes where the buildings are densely located on the lateral side of the robot have ambiguous geometrical characteristics.
Accordingly, sometimes distant two point clouds measured in the corridor-like environments or opposing viewpoint cases are considered as the same place. 
The problem is that the MSEs of these scenes are likely to have small values, so some false loops are not rejected via the MSE-based thresholding.
Therefore, these wrong loops are formed on the graph structure, resulting in the misaligned map~(Fig.~\ref{fig:dcc_map}(b)).

\begin{table}[t!]
	\captionsetup{font=footnotesize}
	\centering
	\caption{\color{black}Absolute pose errors of SLAM results in the MulRan dataset. For all the metrics, the lower, the better.}
	\setlength{\tabcolsep}{5pt}
	{\scriptsize \color{black}
		\begin{tabular}{l|l|cccccc}
			\toprule \midrule
			& Algorithm & Mean & Median & RMSE & Stdev.  \\ \midrule
			% DCC01
			\parbox[t]{2mm}{\multirow{4}{*}{\rotatebox[origin=c]{90}{{DCC01}}}}
			& LeGO-LOAM & 28.89 & 29.21 & 32.37 & 14.60   \\
			& SC-LeGO-LOAM & 6.13 & 5.59 & 6.46 & \textbf{1.89}   \\
			& TSC-LeGO-LOAM  & 6.10 & 5.49 & \textbf{5.56} & 2.58 \\
			& QSC-LeGO-LOAM~(Ours) & \textbf{5.65} & \textbf{5.33} & 5.97 & 1.92  \\ \midrule
			% KAIST02
			\parbox[t]{2mm}{\multirow{4}{*}{\rotatebox[origin=c]{90}{{KAIST02}}}}
			& LeGO-LOAM & 24.21 & 18.48 & 30.25 & 18.14   \\
			& SC-LeGO-LOAM & 4.28 & 3.72 & 5.33 & 3.17  \\
			& TSC-LeGO-LOAM  & 4.27 & 3.85 & 5.12 & \textbf{2.82} \\
			& QSC-LeGO-LOAM~(Ours) & \textbf{3.88} & \textbf{2.71} & \textbf{4.86} & 2.93  \\ \midrule
			% Riverside01
			\parbox[t]{2mm}{\multirow{4}{*}{\rotatebox[origin=c]{90}{{Riverside01}}}}
			& LeGO-LOAM & 96.60 & 62.57 & 124.94 & 79.27   \\
			& SC-LeGO-LOAM  & 26.87 & 28.99 & 30.59 & 26.87   \\
			& TSC-LeGO-LOAM  & 21.23 & 21.23 & 32.89 & \textbf{19.50} \\
			& QSC-LeGO-LOAM~(Ours) & \textbf{20.19} & \textbf{19.57} & \textbf{23.01} & 19.57  \\ \midrule \bottomrule
		\end{tabular}
	}
	\label{table:ape}
\end{table}

 \begin{figure*}[t!]
	\captionsetup{font=footnotesize}
	\centering
 	\begin{subfigure}[b]{0.82\textwidth}
 		\includegraphics[width=1.0\textwidth]{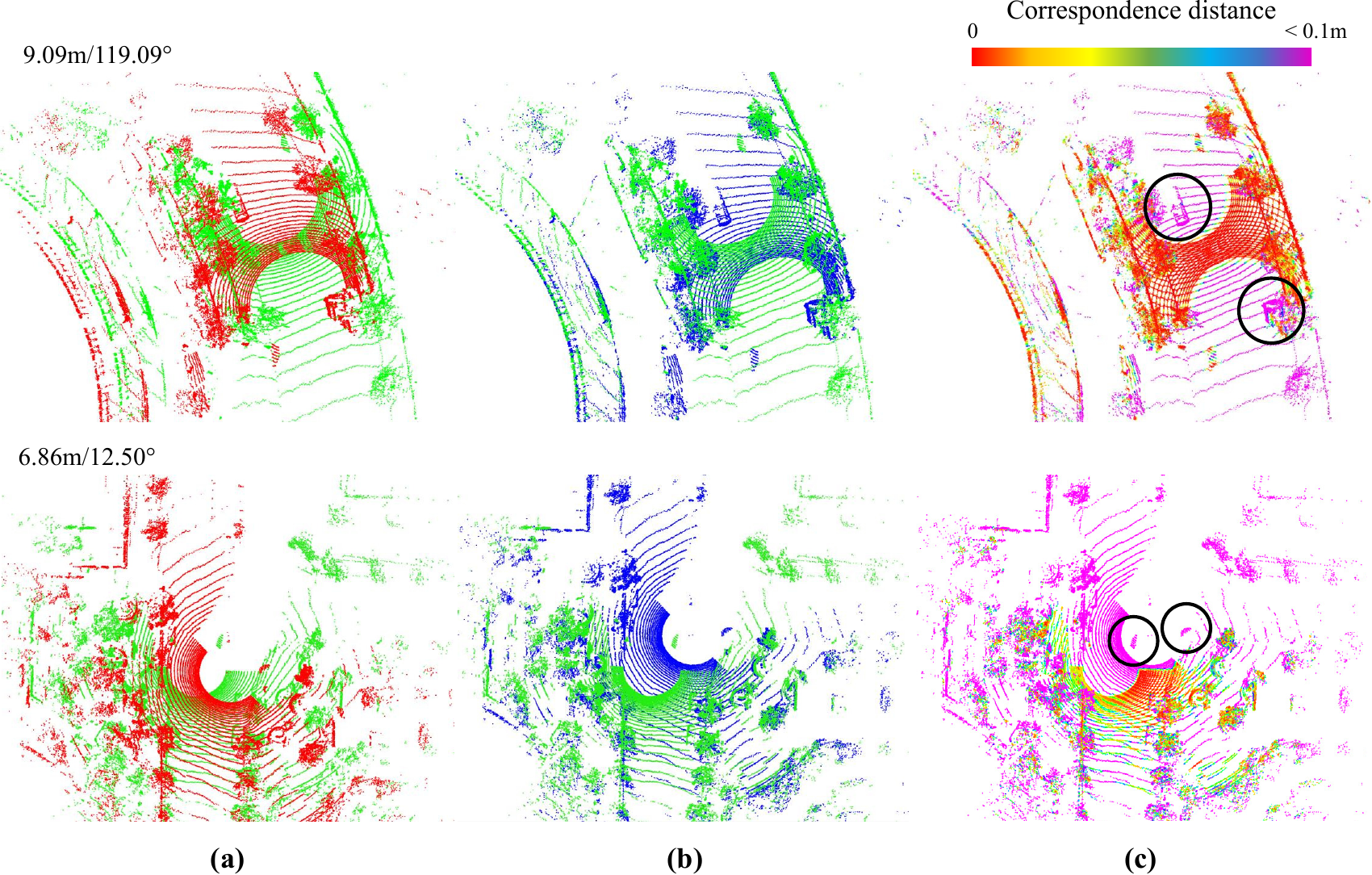}
 	\end{subfigure}
 	\caption{\color{black}Registration results of Quatro++ in distant cases~(T-B): Registration results between 467 and 4,175; 691 and 4,887 frames in \texttt{DCC01} of the MulRan dataset. (a) Even though the partially overlapped source~(red) and target~(green) clouds, which contain moving objects, are given, (b)~our proposed method showed tightly aligned results. The closer the warped source cloud~(blue) and target cloud are, the better. (c)~The solid black circles highlight the existence of moving objects and rainbow colors indicate the closest distances between the points of source and target clouds. The top-left texts represent the initial pose discrepancy in each scene~(best viewed in color). \color{black}}
 	\label{fig:correspondence_distance}
 \end{figure*}

% riverside - wide-open view에서도 그러한 비슷한 효과가 나타남
In contrast, the application of our proposed method in the loop closing showed a tightly aligned mapping result and fewer false loops, as shown in \color{black} Figs.~\ref{fig:dcc_map}(d) and~\ref{fig:riverside_loops}(d). \color{black}
As explained earlier, our proposed method showed more robust performance against these ambiguous scenes and reverse cases by leveraging ground segmentation.
Accordingly, even though distant or reversed loop pairs are given by the loop detection module, our proposed method can successfully estimate the relative pose in a coarse manner.
Consequently, the local registration method, which follows the global registration, can estimate the correct relative pose as a fine alignment.
In particular, we also highlight that our proposed method increases the number of true positive loops in widely opened environments, showing more dense green lines in Fig.~\ref{fig:riverside_loops}(d) \color{black} compared with Fig.~\ref{fig:riverside_loops}(b). \color{black}

\color{black}
Certainly, TSC-LeGO-LOAM also increases the number of loop pairs~(Fig.~\ref{fig:riverside_loops}(c)), yet TEASER++ triggers some false positive loops.
This is because TEASER++ showed the lower success rates in the riverside scenes, as presented in Table~\ref{table:success_rate_in_mulran}, so TEASER++ occasionally let the fine alignment by local registration converge to the local minimum by providing a wrong initial guess as a coarse alignment.
Compared with TSC-LeGO-LOAM, our QSC-LeGO-LOAM only led to few false positive loop pairs due to the robustness of Quatro++ against the outliers and sparsity issues, showing better SLAM performance (Fig.~\ref{fig:Barplot_ape_rpe} and Table~\ref{table:ape}).
In addition, by leveraging its robustness, Quatro++ can perform registration as a loop closing even when moving objects, which may lead to erroneous feature matching, are included~(Fig.~\ref{fig:correspondence_distance}).

\color{black}

% Note that these circumstances that our proposed method helps the convergence of the local registration were observed repeatedly, as shown in Fig.~\ref{fig:riverside_loops}.

Through these experiments, \color{black} we demonstrate \color{black} that our proposed method has positive effect on the LiDAR SLAM: our proposed method increases the quality of loop constraints by reducing both false positive and false negative loops, reducing the pose errors, as presented in Fig.~\ref{fig:Barplot_ape_rpe} and \color{black} Table~\ref{table:ape}. \color{black}

\section{Conclusion}

In this study, a robust global registration method, \textit{Quatro++}, has been proposed. 
Our proposed method was proven to be more robust against the sparsity and degeneracy issues than the state-of-the-art global registration methods by utilizing ground segmentation.
Furthermore, we applied our proposed method to the loop closing module in LiDAR SLAM and confirmed that the quality of the loop constraints was improved, showing more precise mapping results.

Although we demonstrated that our proposed method outperformed the state-of-the-art methods in terms of global registration and loop closing in LiDAR SLAM, there are some limitations.
That is, we placed more emphasis on the back-end part of the correspondence-based global registration, so the improvement of feature extraction and matching was beyond our scope.
In future works, we plan to improve the front-end of the correspondence-based global registration.
Accordingly, the way to increase the quality of the estimated correspondences should be addressed to tackle the sparsity and degeneracy issues in terms of the front-end part as well.

\section{Funding}

This work was financially supported in part by Korea Evaluation Institute of Industrial Technology (KEIT) funded by the Korea Government (MOTIE) under Grant No.20018216, Development of mobile intelligence SW for autonomous navigation of legged robots in dynamic and atypical environments for real application,
and the Unmanned Swarm CPS Research Laboratory Program of Defense Acquisition Program Administration and Agency for Defense Development~(UD220005VD).
The researchers are supported by BK21 FOUR.

\bibliographystyle{SageH}
\bibliography{main}

\end{document}